\documentclass[a4paper]{article}

\usepackage{arxiv}

\usepackage{lipsum}
\usepackage[mathlines]{lineno}
\usepackage{siunitx}
\usepackage{booktabs}
\usepackage{multirow}
\usepackage{graphicx,xcolor}
\usepackage{url,hyperref}
\usepackage{algorithm,algorithmicx,algpseudocode}
\usepackage{listings}
\usepackage[hang,small,bf]{caption}
\usepackage[subrefformat=parens]{subcaption}
\usepackage{amsmath,amssymb,amsfonts,amsthm}
\usepackage{bm}
\usepackage{mathrsfs,mathtools}
\usepackage{latexsym}
\usepackage{fix-cm}
\usepackage{lmodern}
\usepackage[title]{appendix}
\usepackage{textcomp}
\usepackage{manyfoot}
\usepackage{algorithm,algorithmicx,algpseudocode}
\usepackage{listings}
\usepackage{url,hyperref}
\usepackage{orcidlink}
\usepackage{cite}

\hypersetup{
    colorlinks=true,
    linkcolor=blue,
    filecolor=magenta,
    urlcolor=cyan,
    linkbordercolor=red,
}

\title{
    Reliable and efficient inverse analysis
    using physics-informed neural networks
    with normalized distance functions
    and adaptive weight tuning
}

\author{
    Shota DEGUCHI\thanks{Corresponding author.}\orcidlink{0000-0002-9538-8663} \\
    Institute of Systems and Information Engineering \\
    University of Tsukuba \\
    Ibaraki, Japan \\
    \texttt{deguchi.shota.gm@u.tsukuba.ac.jp}
    \And
    Mitsuteru ASAI\orcidlink{0000-0002-1124-2895} \\
    Department of Civil Engineering \\
    Kyushu University \\
    Fukuoka, Japan \\
    \texttt{asai@doc.kyushu-u.ac.jp}
}

\date{\today}

\begin{document}
\maketitle


\begin{abstract}
    Physics-informed neural networks
    have attracted significant attention in scientific machine learning
    for their capability to solve forward and inverse problems
    governed by partial differential equations.
    However, the accuracy of PINN solutions is often limited by the treatment of boundary conditions.
    Conventional penalty-based methods,
    which incorporate boundary conditions as penalty terms in the loss function,
    cannot guarantee exact satisfaction of the given boundary conditions
    and are highly sensitive to the choice of penalty parameters.
    This paper demonstrates that distance functions, specifically R-functions,
    can be leveraged to enforce boundary conditions, overcoming these limitations.
    R-functions provide normalized distance fields,
    enabling flexible representation of boundary geometries,
    including non-convex domains,
    and facilitating various types of boundary conditions.
    Nevertheless, distance functions alone are insufficient for accurate inverse analysis in PINNs.
    To address this, we propose an integrated framework that combines
    the normalized distance field with bias-corrected adaptive weight tuning
    to improve both accuracy and efficiency.
    Numerical results show that the proposed method
    provides more accurate and efficient solutions to various inverse problems
    than penalty-based approaches,
    even in the presence of non-convex geometries with complex boundary conditions.
    This approach offers a reliable and efficient framework for inverse analysis using PINNs,
    with potential applications across a wide range of engineering problems.
\end{abstract}

\keywords{
    scientific machine learning,
    physics-informed neural network,
    normalized distance function,
    adaptive weight tuning,
    inverse analysis,
    incompressible flow
}


\subsubsection*{Article Highlights}
The main contributions of this study are as follows:
\begin{itemize}
    \item Providing a systematic classification of potential hard imposition approaches in PINNs and clarifying their differences in terms of geometric representation and boundary condition treatment.
    \item Employing R-function-based normalized distance fields to impose a wide range of boundary conditions in PINNs over convex and non-convex domains.
    \item Integrating normalized distance fields with bias-corrected adaptive weight tuning to achieve faster convergence and accurate inverse analysis in PINNs.
\end{itemize}


\section{Introduction}   \label{sec:introduction}
Partial differential equations (PDEs) are of fundamental importance in science and engineering,
as they describe the physical laws and behaviors of many real-world phenomena.
Obtaining analytical solutions to PDEs is often challenging.
Consequently, numerical methods such as the finite difference method,
the finite volume method
and the finite element method
have been studied and developed to solve them numerically.
While these methods have successfully addressed many problems,
they have limitations, such as the need for grid or mesh generation prior to simulation.
Moreover, these methods are primarily focused on forward problems,
where the solutions are sought for given parameters and boundary conditions (BCs).
In contrast, these methods are not directly applicable to inverse problems,
which involve estimating physical parameters or unobserved quantities from scarce and noisy data.

In recent years, neural networks have emerged as an alternative approach for solving PDEs,
with the pioneering studies appearing in the 1990s~\cite{Dissanayake1994,Lagaris1998,Leephakpreeda2002,McFall2009}.
With recent advancements in computational hardware and the spread of machine learning libraries,
neural network-based methods have further gained significant attention across diverse disciplines,
not only in computer vision and natural language processing,
but also in science and engineering.
However, a major challenge associated with neural network-based methods is
their lack of explainability and trustworthiness~\cite{Arrieta2020,Roscher2020}.
Such concerns have led to a shift towards the development of models that incorporate prior knowledge,
thereby improving their interpretability and reliability
compared to purely data-driven methods~\cite{vonRueden2023}.
This is also the case when applying ML to physical problems,
which has led to the emergence of a new research field known as scientific machine learning (SciML)~\cite{Rackauckas2021SciML}.
Several approaches have been proposed to integrate known physical laws
into the training process of neural networks.
Notable examples include the Physics-Informed Neural Network (PINN)~\cite{Raissi2019PINN},
in which the governing PDEs are introduced into the loss function;
the Deep Ritz Method~\cite{E2018DRM,Liao2021DNM,Samaniego2020Energy},
where the corresponding energy functional is minimized;
and the Variational PINN~\cite{Kharazmi2019VPINN,Kharazmi2021hpVPINN},
the Weakly Adversarial Network~\cite{Zang2020WAN},
and the deep mixed residual method~\cite{Lyu2022MIM},
which are similar to PINN but relax the regularity of trial solutions
by using weak formulations or auxiliary variables.
One of the key advantages of PINN and its variants is their flexibility in handling
both forward and inverse problems in a unified framework with minimal code modifications.
In addition, embedded physical laws serve to effectively constrain the search space of parameters,
leading to more accurate solutions and robustness to noisy data~\cite{Deguchi2023Doboku,Wong2022Robustness,Eivazi2024}.
PINNs and variants also reduce the need for large amounts of data,
making them suitable for problems with limited data availability~\cite{Tartakovsky2020SubsurfaceFlow,Kissas2020MRI,Arzani2021}.
In this regard, PINN can be viewed as a class of Sobolev training networks~\cite{Czarnecki2017Sobolev,Maddu2022InvDirichlet},
while one can explicitly introduce an additional regularization~\cite{Yu2022GradEnhc,Son2021Sobolev,Vlassis2021Sobolev}.
This prior knowledge-based regularization has been effectively utilized in various fields,
as demonstrated by Kissas et al.~\cite{Kissas2020MRI} for pressure reconstruction from scarce and noisy medical data,
and Sun et al.~\cite{Sun2023ShapeOptimization} for airfoil design optimization, to name a few.
There are also studies to improve the robustness and approximation capabilities of PINN by various means.
These include architecture modifications
such as employing gate or residual structures~\cite{Sirignano2018DGM,Wang2021GradPath,Wang2023ExpertGuide};
basis transformations
with Fourier~\cite{Tancik2020FourierFeature,Wang2021FourierFeature} or polynomial series~\cite{Tang2023PolyPINN};
and domain decompositions,
where distinct neural network approximations are applied to individual subdomains,
with predetermined non-overlapping~\cite{Jagtap2020XPINN},
overlapping~\cite{Li2023D3M,Moseley2023FBPINN},
or trainable overlapping subdomains~\cite{Hu2023APINN}.

Despite the considerable potential and advancements,
the accuracy of PINN solutions is still limited by the treatment of boundary conditions.
In practice, many studies only consider the BCs via the loss function,
which we refer to as `soft imposition' of boundary conditions.
Although this penalty-based method is relatively easy to implement and has been widely adopted,
it can lead to poor approximations if not applied carefully~\cite{Sun2020Surrogate}.
Potential countermeasures have been proposed, including:
\begin{itemize}
    \item[(1)] Increasing the weight of the boundary condition loss to enhance soft imposition effect~\cite{E2018DRM,Chen2020DGMvsDRM,Wight2021}
    \item[(2)] Utilizing multiple networks, each learning the prescribed BCs and the governing PDEs separately~\cite{Berg2018,Rao2020arXiv,Sheng2021PFNN}
    \item[(3)] Enforcing boundary conditions using distance functions to the boundary~\cite{Lagaris1998,Sun2020Surrogate,Lu2021HardDesign,Sukumar2022MixedBC}
\end{itemize}
For methods (1) and (2), however, the accuracy of the solution is
heavily dependent on the choice of the weight (penalty parameter) and the optimization method.
In particular, for method (1), several studies have proposed adaptive weight tunings
based on statistics or norm of backpropagated gradients~\cite{Maddu2022InvDirichlet,Wang2021GradPath,Deguchi2023DynNorm},
or Neural Tangent Kernel theory~\cite{Jacot2018NTK,Wang2022NTK}.
Nonetheless, including method (2), these approaches only provide approximations,
and therefore the resulting solutions are not guaranteed to satisfy the given BCs.
Method (3), by contrast, is independent of the optimizer or approximator in terms of boundary condition enforcement,
and is therefore referred to as `hard imposition'.
In this method, the approximate solution is obtained by projecting the output of the neural network
onto the space that exactly satisfies the boundary conditions.
This projection is realized using distance functions,
which can be constructed in several ways.
Here, we note that the term `hard imposition' has been used across various contexts,
resulting in a broad range of interpretations without clear distinction.
In particular, studies employing distance functions, whether normalized or not (as discussed below),
or even those approximated by another neural network, which should fall under method (2) above,
are sometimes called hard imposition.
While we do not attempt to standardize the terminology in this study,
it remains of great importance to distinguish the different approaches presented in the literature,
so that it is clear which strategy is applicable to a given problem setting.
Here, we classify method (3), distance function-based strategy,
into the following three distinct classes from two perspectives:
flexibility in geometric representation and versatility in boundary condition treatment (see Table~\ref{tab:taxonomy_hard_imposition} for a summary):
\begin{itemize}
    \item[(3-i)] Na\"ive approach:
        represents the domain as the intersection of half-spaces whose zero level sets are located to coincide with boundary components;
        applicable only to convex geometries and typically restricted to homogeneous Dirichlet conditions
        (or at most up to an additive constant for a global translation; presented in, e.g.,~\cite{Lagaris1998,Lu2021HardDesign}).
    \item[(3-ii)] Non-normalized approach:
        represents the domain boundary as the union of distance functions to boundary components, which are based on R-functions, but joined without normalization;
        applicable to both convex and non-convex geometries yet still limited to homogeneous Dirichlet conditions
        (or up to an additive constant)
    \item[(3-iii)] Normalized approach:
        represents the domain boundary as the union of distance functions to boundary components, based on R-functions, and joined with normalization to preserve essential properties near the boundary;
        applicable to both convex and non-convex geometries, and capable of handling various boundary conditions including spatially varying inhomogeneous Dirichlet and mixed boundary conditions (e.g.,~\cite{Sukumar2022MixedBC}).
\end{itemize}
Among them, this study focuses on approach (3-iii),
which builds distance functions based on the theory of R-functions~\cite{Sheiko1982,Rvachev2001,Shapiro2007},
a class of functions to encode logical operations into real-valued functions
widely utilized in geometric modeling~\cite{Shapiro2007,Shapiro1994,Shapiro1999DiffProperties}.
This choice is motivated by the advantages of R-functions in geometric representation and boundary condition enforcement.
In comparison to (3-iii),
approach (3-i) has been frequently adopted in the literature due to its simplicity.
However, it exhibits severe limitations in representing complex geometries and imposing various BCs,
while these shortcomings are often overlooked and left undiscussed.
Approach (3-ii) also employs R-functions but lacks normalization in the joining process.
As a result, its BC enforcement capability is limited,
while this class of formulation is readily extendable to (3-iii).
The distinctions among these approaches,
including normalization properties and associated limitations,
are detailed with illustrative examples in Section~\ref{sec:r_function}.

While the application of R-function-based normalized distance fields (approach (3-iii))
to numerical simulations~\cite{Shapiro2002SAGE,Tsukanov2003MeshfreeFluid,Millan2015}
and machine learning methods~\cite{Sukumar2022MixedBC,Berrone2023BCEnforcement,Gladstone2025FOPINN}
has been explored for forward problems,
their use in inverse problems with PINNs has not been fully investigated.
This study addresses the aforementioned limitations in accuracy,
attributed to the inadequate treatment of boundary conditions,
by employing R-function-based normalized distance fields
to enforce various BCs over convex and non-convex domains~\cite{Sukumar2022MixedBC,Rvachev2001,Rvachev1995,Shapiro2007}.
However, we found that distance functions alone are insufficient to achieve accurate inverse analysis in PINNs.
To overcome this limitation, we propose an integrated framework that combines
the normalized distance field with a bias-corrected adaptive weight tuning~\cite{Deguchi2023DynNorm}
to balance contributions from multiple loss components.
The proposed integrated approach provides two key benefits.
First, normalized distance functions enable precise representation of complex geometries
and various types of conditions, thereby improving the reliability of PINN solutions.
Second, the adaptive weight tuning provides appropriate weights to multiple objectives
from the early stages of optimization, with the help of bias-correction,
offering faster convergence and higher accuracy.
To the best of our knowledge,
this is the first study to extend R-function-based normalized distance functions to inverse analysis using PINNs,
further enhanced by integration with bias-corrected adaptive weights.
We demonstrate the effectiveness of this integrated approach through numerical experiments,
showing improved accuracy and efficiency in solving inverse problems with complex geometries and boundary conditions.

The remainder of this paper is organized as follows.
In Section~\ref{sec:methods},
we first provide an overview of physics-informed neural networks (PINNs),
taking the Poisson equation as a model problem.
We then summarize the fundamental properties of exact and approximate distance functions,
which are tightly related to the normalization properties of distance fields.
Next, we review the construction of distance fields in approaches (3-i), (3-ii), and (3-iii)
and illustrate their differences with examples.
We also describe how boundary conditions are incorporated into trial functions.
Following this, we introduce a gradient norm-bsaed adaptive weight tuning method,
which is necessary for optimizing multiple objectives that arise in inverse problems.
Furthermore, we show that adaptive weights can be biased towards the initial values under certain conditions
and present a correction strategy to mitigate this issue,
along with a technique to further enhance the reliability of inverse analysis.
In Section~\ref{sec:results}, we present a series of numerical experiments
to evaluate the performance of the proposed integrated method in comparison with the widely adopted penalty-based approach.
Finally, Section~\ref{sec:conclusion} concludes the paper by highlighting the significance of our work.

\begin{table}[tpb]
    \centering
    \caption{
        \emph{Taxonomy of distance function-based hard imposition approaches in the context of PINNs.}
        Classified according to geometric representation flexibility and boundary condition enforcement versatility.
        See Section~\ref{sec:r_function} for details.
    }
    \label{tab:taxonomy_hard_imposition}
    \begin{tabular}{l|ccc}
        \toprule
            & Na\"ive (Eq.~\eqref{eq:naive_adf}) & Non-normalized (Eq.~\eqref{eq:non_normalized_adf}) & Normalized (Eq.~\eqref{eq:normalized_adf}) \\
        \midrule
        Geom. repr. & Convex & Convex/non-convex & Convex/non-convex \\
        Bound. cond. enf. & Hmg. Dirichlet & Hmg. Dirichlet & Hmg./inhmg. Dirichlet/mixed \\
        Refs.  & \cite{Lagaris1998,Lu2021HardDesign,Li2024HardAdvcDiff} & -- & \cite{Sukumar2022MixedBC,Berrone2023BCEnforcement} \\
        \bottomrule
    \end{tabular}
\end{table}


\section{Methods}   \label{sec:methods}
\subsection{Background: physics-informed neural networks}   \label{sec:pinn}
For the sake of clarity,
we first provide a brief overview of physics-informed neural networks (PINNs),
taking the Poisson equation as a model problem.
The discussion can be readily extended to other types of equations or to unsteady problems
by employing a space-time strategy,
which is a common practice in the literature (see, e.g.,~\cite{Sun2020Surrogate,Kharazmi2021hpVPINN,Yu2022GradEnhc}).

Consider the following boundary value problem of the Poisson equation:
\begin{alignat}{2}
    -\nabla^2 u
    &= f
    && \quad \text{in} \quad \Omega,
    \label{eq:poisson}
    \\
    u
    &= g_D
    && \quad \text{on} \quad \Gamma_D,
    \label{eq:poisson_Dirichlet}
    \\
    \bm{n} \cdot \nabla u
    &= g_N
    && \quad \text{on} \quad \Gamma_N,
    \label{eq:poisson_Neumann}
\end{alignat}
where $\Omega \subset \mathbb{R}^d$ is an open, bounded domain,
and $\Gamma = \partial \Omega$ denotes its boundary
($\Gamma = \bar{\Gamma}_D \cup \bar{\Gamma}_N$,
$\Gamma_D \cap \Gamma_N = \emptyset$,
$\Gamma_D \neq \emptyset$).
Here, $\bm{n}$ represents the outward unit normal vector on $\Gamma$,
and $f: \Omega \to \mathbb{R}$, $g_D: \Gamma_D \to \mathbb{R}$, and $g_N: \Gamma_N \to \mathbb{R}$ are given functions.
The objective is to find the solution $u: \Omega \to \mathbb{R}$ that satisfies Equations~\eqref{eq:poisson}--\eqref{eq:poisson_Neumann}.

Following~\cite{Raissi2019PINN}, we proceed by approximating the solution $u (\bm{x})$
with a multi-layer perceptron (MLP) $\hat{u} (\bm{x}; \bm{\theta})$~\cite{Hornik1989,Leshno1993},
where $\bm{\theta}$ denotes the set of trainable parameters.
Given an input $\mathbf{x} \in \mathbb{R}^{q^{(0)}}$
and an output $\hat{\mathbf{y}} \in \mathbb{R}^{q^{(L)}}$,
the forward pass $\mathbf{z}^{(l)} \in \mathbb{R}^{q^{(l)}} (l = 1, \ldots, L)$ is given by:
\begin{equation}
    \mathbf{z}^{(l)}
    = \sigma^{(l)} \left(
        \mathbf{W}^{(l)} \mathbf{z}^{(l-1)} + \mathbf{b}^{(l)}
    \right),
    \label{eq:mlp_forward}
\end{equation}
where $\mathbf{z}^{(0)} = \mathbf{x}$ and $\mathbf{z}^{(L)} = \hat{\mathbf{y}}$.
$\sigma^{(l)} (\cdot)$ is the element-wise nonlinearity,
and $\mathbf{W}^{(l)} \in \mathbb{R}^{q^{(l)} \times q^{(l-1)}}$ and $\mathbf{b}^{(l)} \in \mathbb{R}^{q^{(l)}}$
represent the learnable weights and biases, respectively,
with $\bm{\theta} = \lbrace \mathbf{W}^{(l)}, \mathbf{b}^{(l)} \rbrace_{l=1}^{L}$.
In this context, $\mathbf{x}$ corresponds to the spatial coordinates $\bm{x}$,
and $\hat{\mathbf{y}}$ represents the approximate solution $\hat{u}$.
PINN is trained by minimizing the following loss function:
\begin{equation}
    \mathcal{L} (\bm{\theta})
    = \mathcal{L}_{\text{PDE}} (\bm{\theta})
    + \lambda_{\text{BC}} \mathcal{L}_{\text{BC}} (\bm{\theta})
    + \lambda_{\text{Data}} \mathcal{L}_{\text{Data}} (\bm{\theta}),
    \label{eq:loss}
\end{equation}
where each of the terms is evaluated via (quasi-)Monte Carlo approximation (e.g.,~\cite{Matsubara2023GLT}):
\begin{align}
    \begin{split}
        \mathcal{L}_{\text{PDE}}
        &= \int_{\Omega}
            \left|
                -\nabla^2 \hat{u} (\bm{x}; \bm{\theta}) - f (\bm{x})
            \right|^2 \, d\bm{x} \\
        &\simeq \frac{1}{N_{\text{PDE}}} \sum_{i=1}^{N_{\text{PDE}}}
            \left|
                -\nabla^2 \hat{u} (\bm{x}_i; \bm{\theta}) - f (\bm{x}_i)
            \right|^2,
    \end{split}
    \label{eq:loss_pde}
    \\
    \begin{split}
        \mathcal{L}_{\text{BC}}
        &= \int_{\Gamma_D}
            \left|
                \hat{u} (\bm{x}; \bm{\theta}) - g_D (\bm{x})
            \right|^2 \, d\bm{x}
        + \int_{\Gamma_N}
            \left|
                \bm{n} \cdot \nabla \hat{u} (\bm{x}; \bm{\theta}) - g_N (\bm{x})
            \right|^2 \, d\bm{x} \\
        &\simeq \frac{1}{N_{\text{DBC}}} \sum_{i=1}^{N_{\text{DBC}}}
            \left|
                \hat{u} (\bm{x}_i; \bm{\theta}) - g_D (\bm{x}_i)
            \right|^2
        + \frac{1}{N_{\text{NBC}}} \sum_{i=1}^{N_{\text{NBC}}}
            \left|
                \bm{n} \cdot \nabla \hat{u} (\bm{x}_i; \bm{\theta}) - g_N (\bm{x}_i)
            \right|^2,
    \end{split}
    \label{eq:loss_bc}
    \\
    \begin{split}
        \mathcal{L}_{\text{Data}}
        &= \int_{\Omega}
            \left|
                \hat{u} (\bm{x}; \bm{\theta}) - u_{\text{Data}} (\bm{x})
            \right|^2 \, d\bm{x} \\
        &\simeq \frac{1}{N_{\text{Data}}} \sum_{i=1}^{N_{\text{Data}}}
            \left|
                \hat{u} (\bm{x}_i; \bm{\theta}) - u_{\text{Data}} (\bm{x}_i)
            \right|^2.
    \end{split}
    \label{eq:loss_data}
\end{align}
Here, $\lambda_{\text{BC}}, \lambda_{\text{Data}} \in \mathbb{R}_{\ge 0}$ are weights (penalty parameters)
to control the contributions of boundary conditions and observed data to the loss function.
The terms $\mathcal{L}_{\text{PDE}}$, $\mathcal{L}_{\text{BC}}$, and $\mathcal{L}_{\text{Data}}$ represent the loss functions
associated with the PDE residual, BC, and observed data, respectively.
For forward problems, $\lambda_{\text{Data}}$ can be set to zero,
whereas for inverse problems, it is assigned a positive value to reflect the observed data.
The penalty parameter $\lambda_{\text{BC}}$ is often set to a large value
to better propagate the effects of boundary conditions~\cite{E2018DRM,Wight2021,Rohrhofer2023IEEE}.
These weights may be predefined as fixed constants or adaptively adjusted during the training process.
The latter includes methods based on
gradient statistics~\cite{Maddu2022InvDirichlet,Wang2021GradPath},
norms~\cite{Deguchi2023DynNorm,Wang2023ExpertGuide},
or the trace of Neural Tangent Kernel matrices~\cite{Wang2022NTK}.


\subsection{Proposed method: integration of normalized distance and adaptive weight}   \label{sec:proposed_method}
\subsubsection{R-function-based distance fields}   \label{sec:r_function}
\paragraph{Fundamental properties of distance functions}   \label{sec:distance_function}
Let $\Phi (\bm{x})$ be the exact distance function (EDF) from a point $\bm{x} \in \Omega$ to the boundary $\Gamma = \partial \Omega$,
and $\phi (\bm{x})$ represent the approximate distance function (ADF).
By definition, $\Phi$ satisfies the following properties~\cite{Biswas2004}:
\begin{itemize}
    \item[(A)] $\Phi (\bm{x}) = 0 \quad \left( \bm{x} \in \Gamma \right)$
    \item[(B)] $\partial_{\bm{\nu}} \Phi (\bm{x}) = 1 \quad \left( \bm{x} \in \Gamma \right)$
    \item[(C)] $\partial_{\bm{\nu}}^{M} \Phi (\bm{x}) = 0 \quad \left( \bm{x} \in \Gamma, M \in \mathbb{Z}_{\ge 2} \right)$
\end{itemize}
where $\bm{\nu}$ is the inward unit normal vector on $\Gamma$ ($\bm{\nu} = - \bm{n}$).
Since $\phi$ is an approximation of $\Phi$, it is desirable that it possesses similar properties.
Specifically, $\phi$ is required to satisfy properties (A) and (B) as $\Phi$ does,
while property (C) is relaxed to a finite order:
\begin{itemize}
    \item[(a)] $\phi (\bm{x}) = 0 \quad \left( \bm{x} \in \Gamma \right)$
    \item[(b)] $\partial_{\bm{\nu}} \phi (\bm{x}) = 1 \quad \left( \bm{x} \in \Gamma \right)$
    \item[(c)] $\partial_{\bm{\nu}}^{k} \phi (\bm{x}) = 0 \quad \left( \bm{x} \in \Gamma, k = 2, \ldots, m \in \mathbb{Z}_{\ge 2}, \text{for some finite } m \right)$
\end{itemize}
Such a function $\phi$ is said to be normalized to the $m$-th order
or an $m$-th order approximation of $\Phi$~\cite{Rvachev2001,Biswas2004}.
In the following, we denote $\phi = \phi^{(m)}$ to emphasize the order of normalization, if necessary.

\paragraph{Distance to a line segment}   \label{sec:distance_line}
\begin{figure}[tpb]
    \centering
    \begin{minipage}[b]{.24\linewidth}
        \centering
        \includegraphics[width=.99\linewidth]{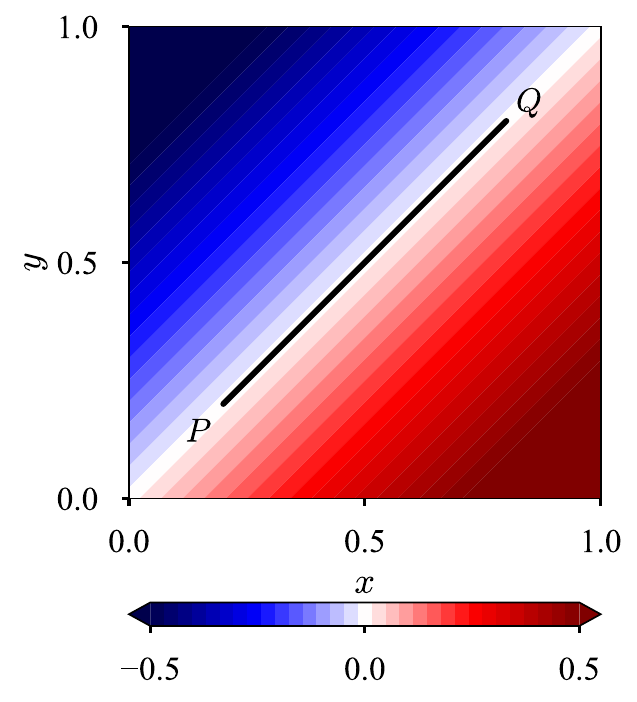}
        \subcaption{SDF, $s_{PQ}$}
    \end{minipage}
    \begin{minipage}[b]{.24\linewidth}
        \centering
        \includegraphics[width=.99\linewidth]{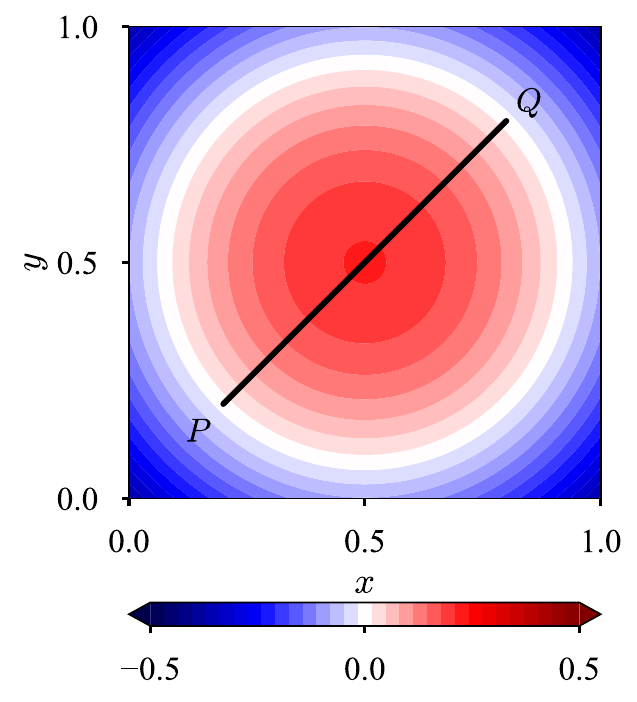}
        \subcaption{TF, $t_{PQ}$}
    \end{minipage}
    \begin{minipage}[b]{.24\linewidth}
        \centering
        \includegraphics[width=.99\linewidth]{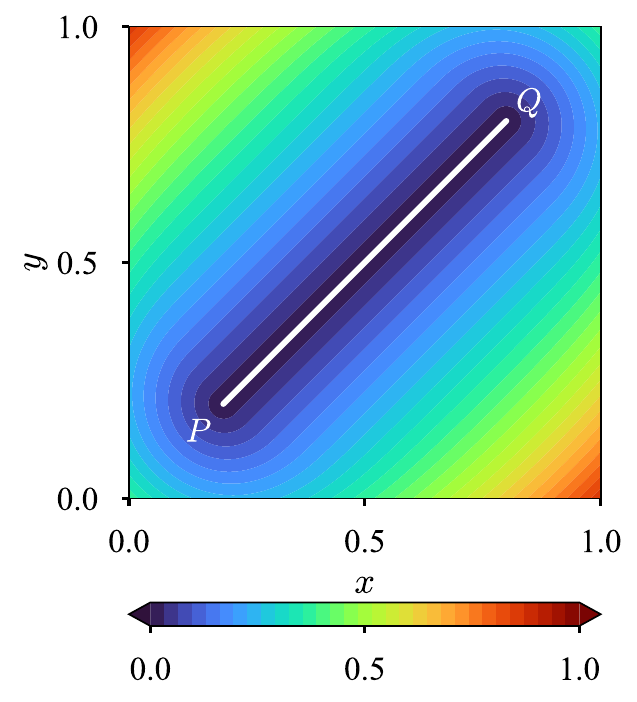}
        \subcaption{ADF, $\phi_{PQ}$}
    \end{minipage}
    \begin{minipage}[b]{.24\linewidth}
        \centering
        \includegraphics[width=.99\linewidth]{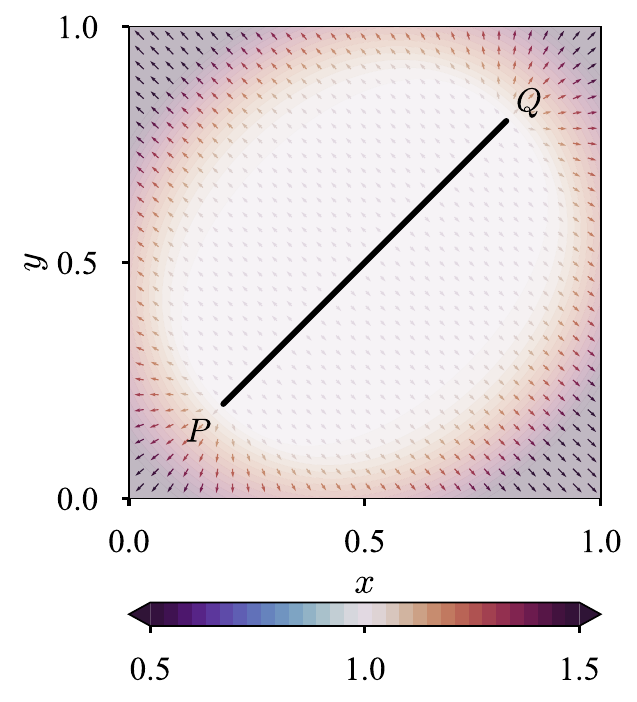}
        \subcaption{Gradient, $\nabla \phi_{PQ}$}
    \end{minipage}
    \caption{
        \emph{Distance to a line segment.}
        Representations of functions defined in
        Equations~\eqref{eq:signed_distance},~\eqref{eq:trimming},~\eqref{eq:approximate_distance}, and its gradient.
        The line segment $S_{PQ}$, realized as the intersection of
        $\mathbb{S} = \left\{ \bm{x} \in \mathbb{R}^2 \mid s_{PQ} \left( \bm{x} \right) = 0 \right\}$
        and
        $\mathbb{T} = \left\{ \bm{x} \in \mathbb{R}^2 \mid t_{PQ} \left( \bm{x} \right) \ge 0 \right\}$,
        is represented by $\mathbb{P} = \left\{ \bm{x} \in \mathbb{R}^2 \mid \phi_{PQ} \left( \bm{x} \right) = 0 \right\}$.
        $\nabla \phi_{PQ}$ offers the unit normal vector in the vicinity of $S_{PQ}$.
    }
    \label{fig:line_segment}
\end{figure}

We begin by considering the distance function to a line segment.
Let $P \left( = \bm{x}_P \right)$ and $Q \left( = \bm{x}_Q \right)$ be the endpoints of the line segment $S_{PQ}$,
with $S^{\prime}_{PQ}$ denoting the infinite line passing through $P$ and $Q$.
First, we define the following signed distance function (SDF) $s_{PQ}$:
\begin{equation}
    s_{PQ} \left( \bm{x} \right)
    = \frac{1}{\|\bm{x}_{PQ}\|} s^{\prime}_{PQ}
    = \frac{1}{\|\bm{x}_{PQ}\|} \bm{n}_{PQ} \cdot (\bm{x} - \bm{x}_P),
    \label{eq:signed_distance}
\end{equation}
where $\| \cdot \|$ denotes the $L^2$-norm,
$s^{\prime}_{PQ}$ is the signed distance to the infinite line $S^{\prime}_{PQ}$,
$\bm{x}_{PQ} = \bm{x}_Q - \bm{x}_P$ is the vector from $P$ to $Q$,
and $\bm{n}_{PQ}$ is the unit normal to $S^{\prime}_{PQ}$.
The function $s_{PQ}$ can be viewed as the signed distance function to the infinite line $S^{\prime}_{PQ}$,
normalized by the length of the segment of interest, $S_{PQ}$.
With these definitions,
the line segment $S_{PQ}$ is realized as a subset of
$\mathbb{S} = \left\{ \bm{x} \in \mathbb{R}^d \mid s_{PQ} \left( \bm{x} \right) = 0 \right\}$.
Next, we introduce the trimming function (TF) $t_{PQ}$:
\begin{equation}
    t_{PQ} ( \bm{x} )
    = \frac{1}{\| \bm{x}_{PQ} \|}
        \left(
            \left( \frac{\| \bm{x}_{PQ} \|}{2} \right)^2
            - \| \bm{x} - \bm{x}_M \|
        \right),
    \label{eq:trimming}
\end{equation}
where $\bm{x}_M = (\bm{x}_P + \bm{x}_Q) / 2$ is the midpoint of $S_{PQ}$.
The region $\mathbb{T} = \left\{ \bm{x} \in \mathbb{R}^d \mid t_{PQ} \left( \bm{x} \right) \ge 0 \right\}$
corresponds to the circular disk whose radius is $\| \bm{x}_{PQ} \| / 2$ and center is $\bm{x}_M$.
Finally, interpreting the line segment $S_{PQ}$ as the intersection of $\mathbb{S}$ and $\mathbb{T}$,
we define the sufficiently smooth approximate distance function (ADF) to the segment $S_{PQ}$ as follows~\cite{Sheiko1982}:
\begin{equation}
    \phi_{PQ} ( \bm{x} )
    = \left(
        s_{PQ}^2
        + \left(
            \frac{\left( s_{PQ}^4 + t_{PQ}^2 \right)^{1/2} - t_{PQ}}{2}
        \right)^2
    \right)^{1/2}.
    \label{eq:approximate_distance}
\end{equation}
Figure~\ref{fig:line_segment} illustrates
the signed distance function $s_{PQ}$,
the trimming function $t_{PQ}$,
the approximate distance function $\phi_{PQ}$ to the line segment $S_{PQ}$,
and its gradient $\nabla \phi_{PQ}$.
The segment $S_{PQ}$ lies in the region
where both $s_{PQ} = 0$ and $t_{PQ} \ge 0$ hold,
and is implicitly represented by $\phi_{PQ} = 0$.
Moreover, $\nabla \phi_{PQ}$ serves as the unit normal vector in the vicinity of $S_{PQ}$,
which is beneficial for considering Neumann boundary conditions.

\paragraph{Distance to the boundary: normalized approach}  \label{sec:distance_domain}
\begin{figure}[tpb]
    \centering
    \begin{minipage}[b]{.24\linewidth}
        \centering
        \includegraphics[width=.99\linewidth]{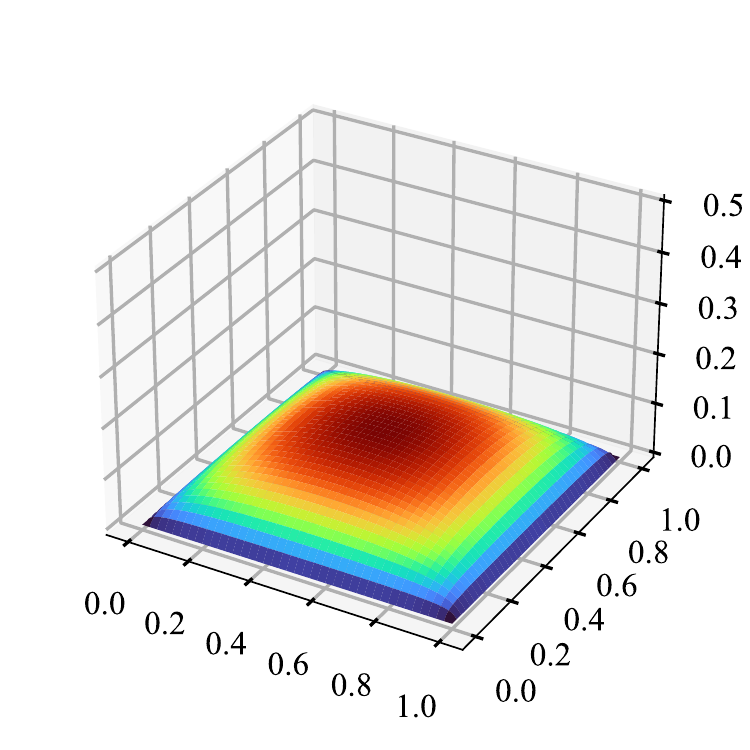}
        \subcaption{ADF, $\phi^{(1)}$}
    \end{minipage}
    \begin{minipage}[b]{.24\linewidth}
        \centering
        \includegraphics[width=.99\linewidth]{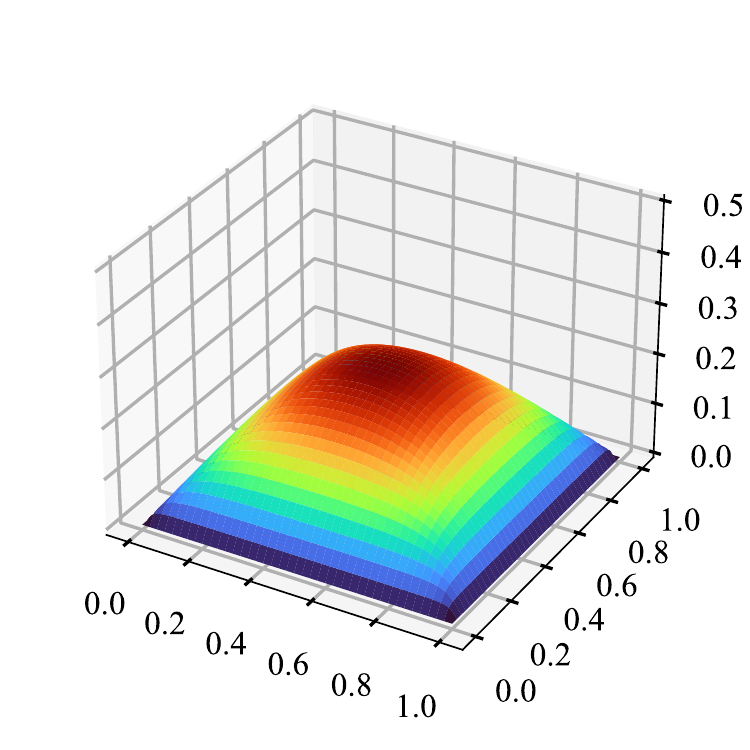}
        \subcaption{ADF, $\phi^{(2)}$}
    \end{minipage}
    \begin{minipage}[b]{.24\linewidth}
        \centering
        \includegraphics[width=.99\linewidth]{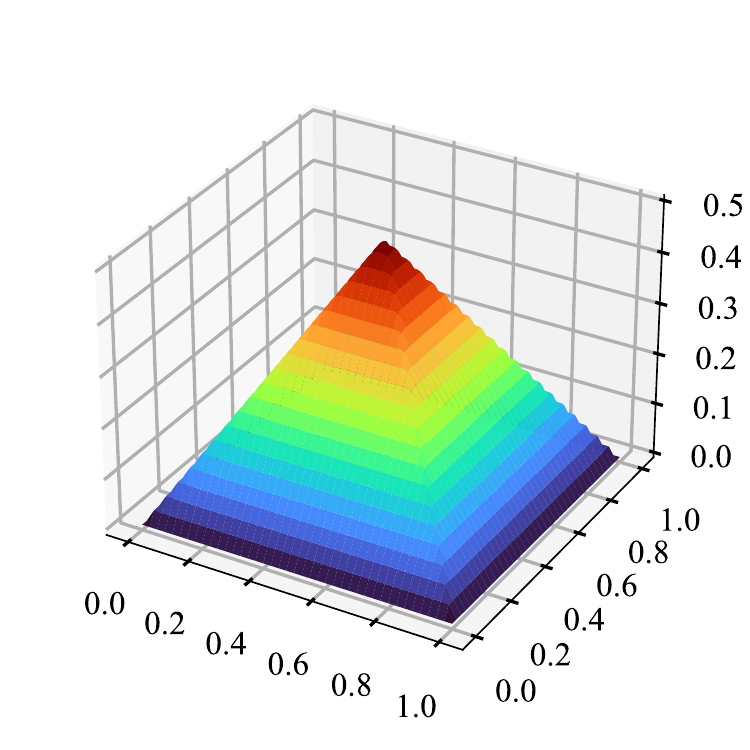}
        \subcaption{ADF, $\phi^{(64)}$}
    \end{minipage}
    \begin{minipage}[b]{.24\linewidth}
        \centering
        \includegraphics[width=.99\linewidth]{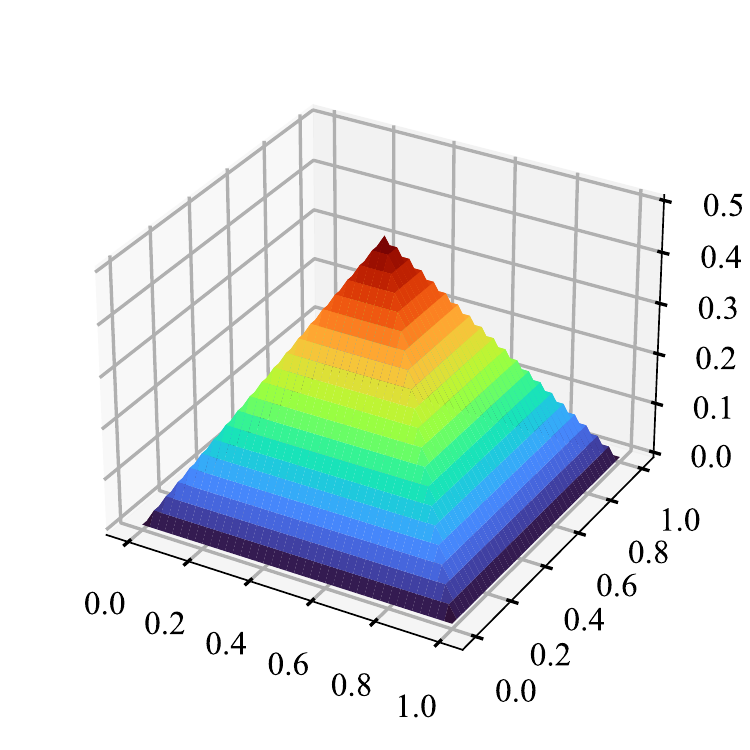}
        \subcaption{EDF, $\Phi$}
    \end{minipage}
    \caption{
        \emph{Distance to the boundary: square.}
        Normalized ADFs are smooth approximations of the EDF,
        and approach the EDF as the normalization order is increased.
    }
    \label{fig:square}
\end{figure}

\begin{figure}[tpb]
    \centering
    \begin{minipage}[b]{.24\linewidth}
        \centering
        \includegraphics[width=.99\linewidth]{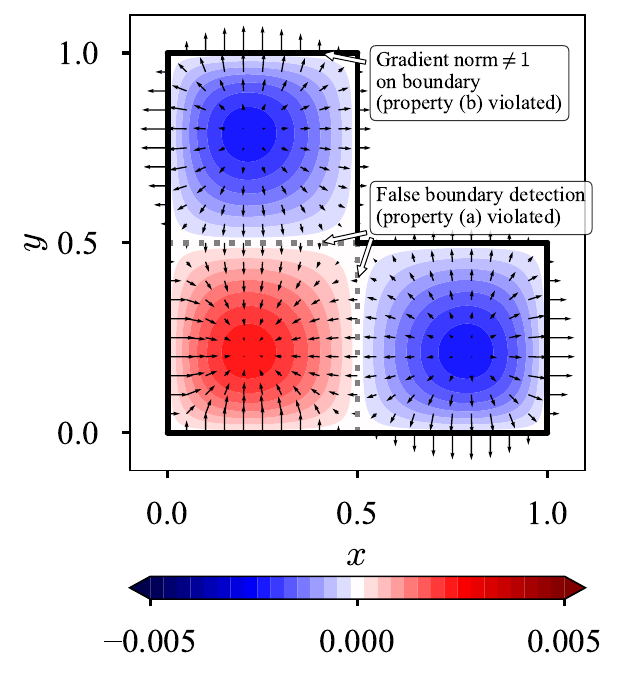}
    \end{minipage}
    \begin{minipage}[b]{.24\linewidth}
        \centering
        \includegraphics[width=.99\linewidth]{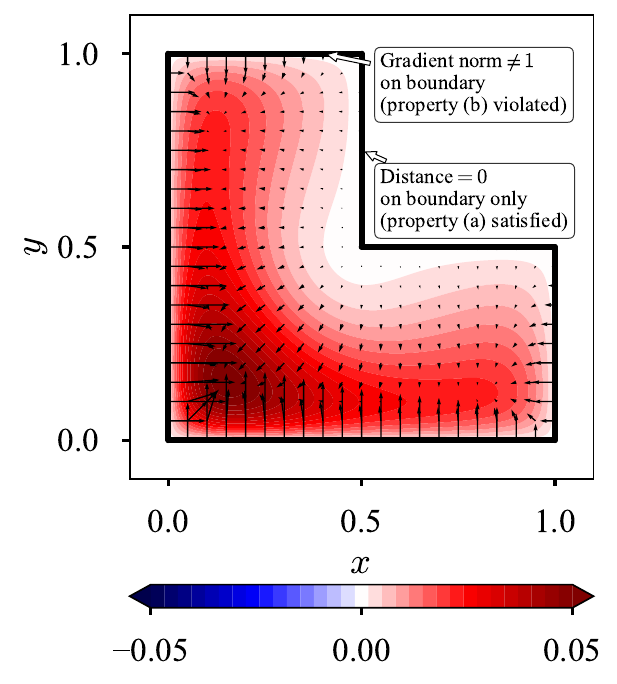}
    \end{minipage}
    \begin{minipage}[b]{.24\linewidth}
        \centering
        \includegraphics[width=.99\linewidth]{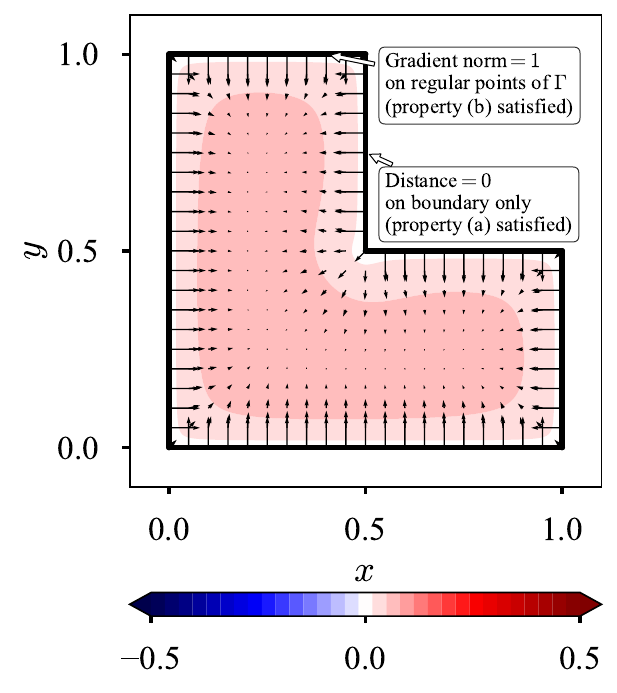}
    \end{minipage}
    \begin{minipage}[b]{.24\linewidth}
        \centering
        \includegraphics[width=.99\linewidth]{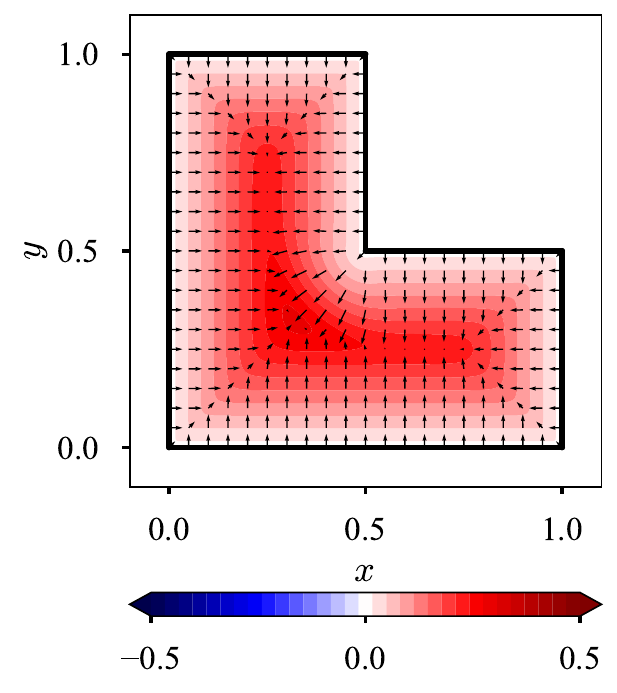}
    \end{minipage}
    \\
    \begin{minipage}[b]{.24\linewidth}
        \centering
        \includegraphics[width=.99\linewidth]{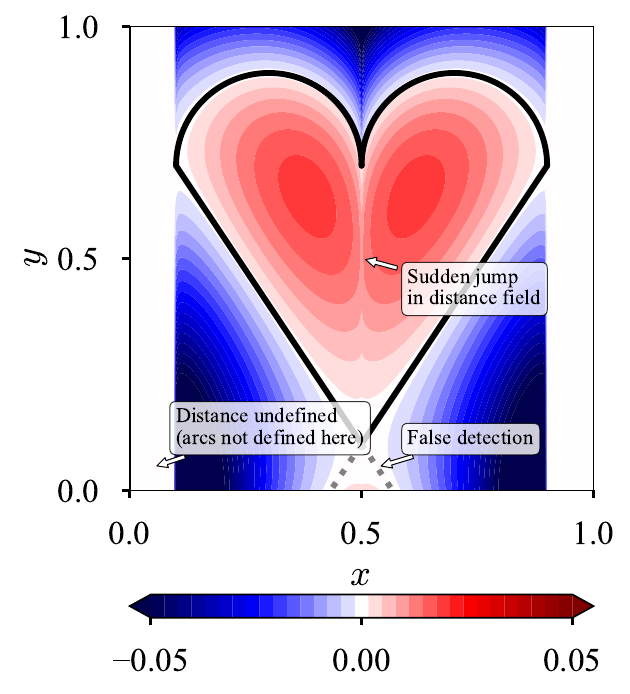}
        \subcaption{Na\"ive, $\psi$}
    \end{minipage}
    \begin{minipage}[b]{.24\linewidth}
        \centering
        \includegraphics[width=.99\linewidth]{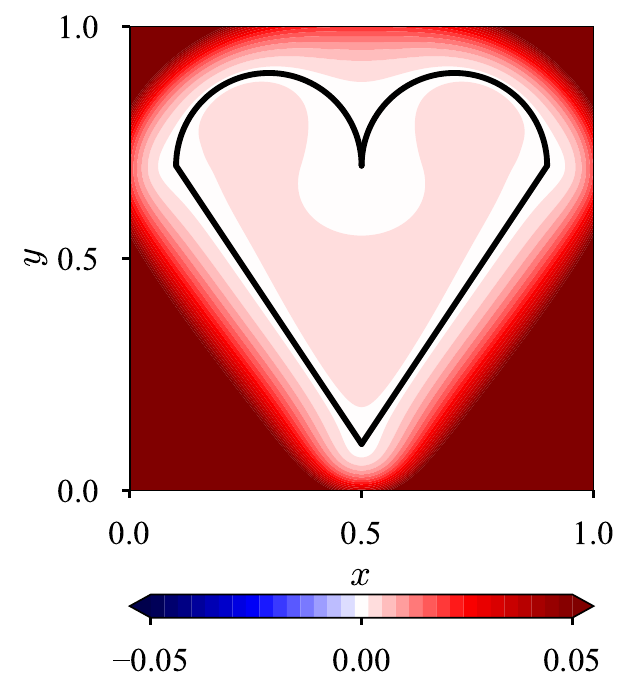}
        \subcaption{Non-normalized, $\varphi$}
    \end{minipage}
    \begin{minipage}[b]{.24\linewidth}
        \centering
        \includegraphics[width=.99\linewidth]{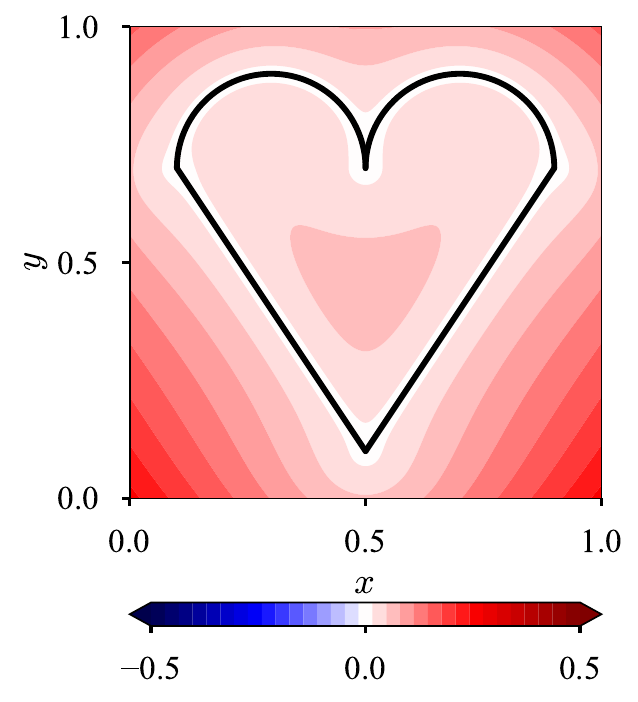}
        \subcaption{Normalized, $\phi^{(1)}$}
    \end{minipage}
    \begin{minipage}[b]{.24\linewidth}
        \centering
        \includegraphics[width=.99\linewidth]{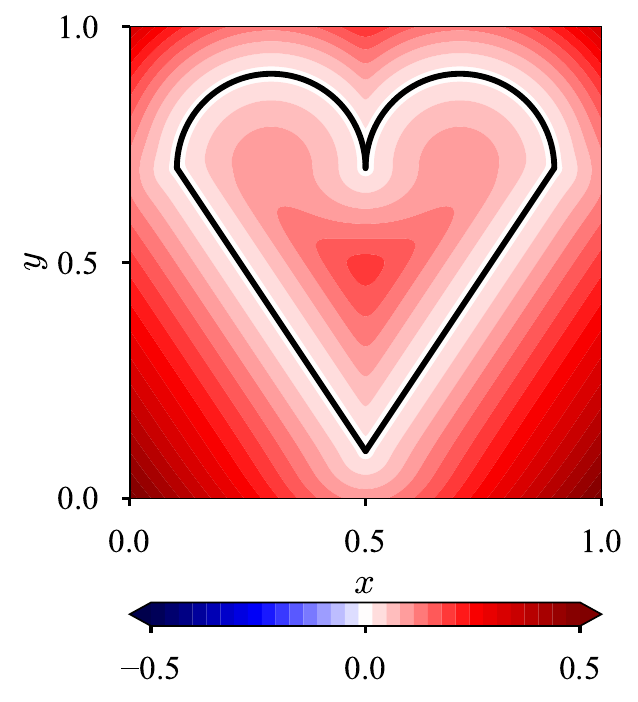}
        \subcaption{Normalized, $\phi^{(8)}$}
    \end{minipage}
    \caption{
        \emph{Distance to the boundary: non-convex shapes.}
        (Top) L-shaped geometry.
        (Bottom) heart-shaped geometry.
        Colors represent the distance fields, and arrows indicate the gradients
        (arrow lengths are scaled only in a relative sense in each panel).
        The na\"ive approach (first column) fails to capture the non-convex geometry
        with undesired zero level sets emerging inside the domain.
        The non-normalized formulation (second column) is able to represent non-convex domains,
        but the gradients are not normalized (i.e., they do not have unit length on the boundary).
        The normalized formulations (third and fourth columns) accurately capture the geometry
        while providing unit normals at the boundary.
    }
    \label{fig:non_convex}
\end{figure}

\begin{figure}[tpb]
    \centering
    \begin{minipage}[b]{.8\linewidth}
        \centering
        \includegraphics[width=.99\linewidth]{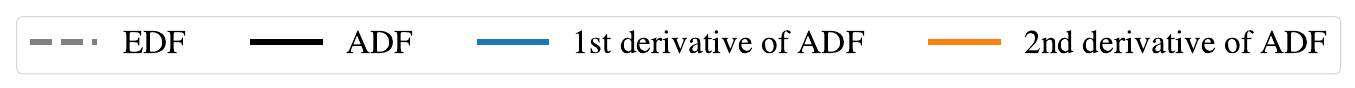}
    \end{minipage}
    \\
    \begin{minipage}[b]{.24\linewidth}
        \centering
        \includegraphics[width=.99\linewidth]{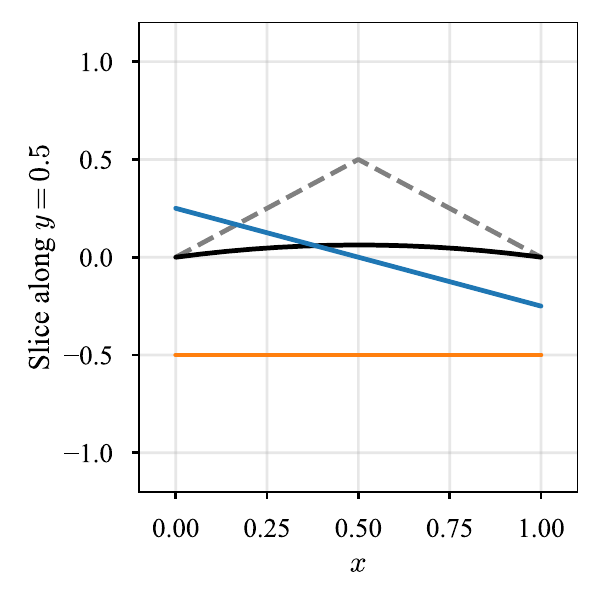}
    \end{minipage}
    \begin{minipage}[b]{.24\linewidth}
        \centering
        \includegraphics[width=.99\linewidth]{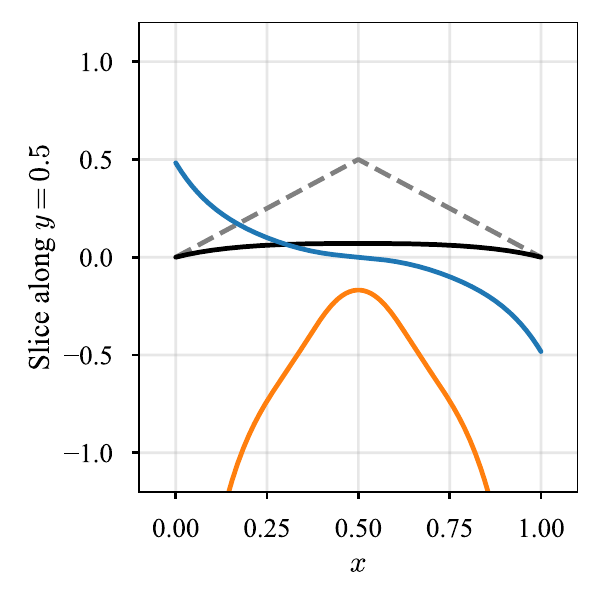}
    \end{minipage}
    \begin{minipage}[b]{.24\linewidth}
        \centering
        \includegraphics[width=.99\linewidth]{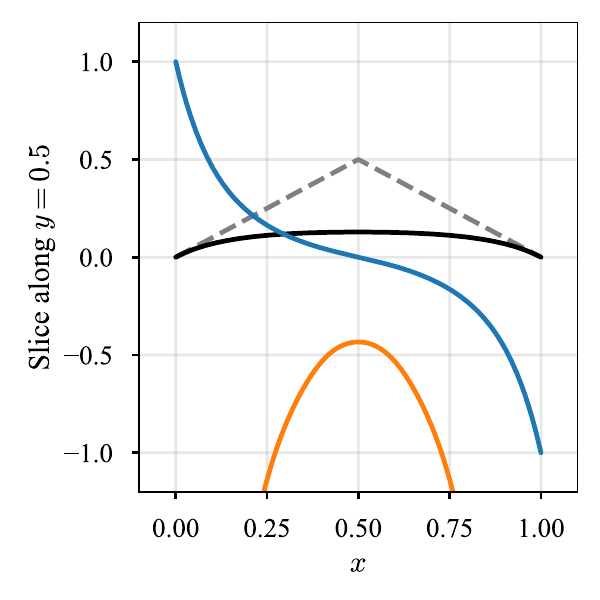}
    \end{minipage}
    \begin{minipage}[b]{.24\linewidth}
        \centering
        \includegraphics[width=.99\linewidth]{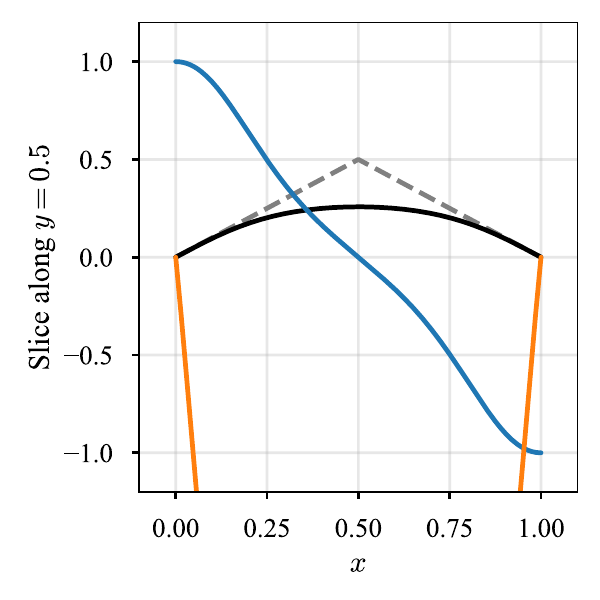}
    \end{minipage}
    \\
    \begin{minipage}[b]{.24\linewidth}
        \centering
        \includegraphics[width=.99\linewidth]{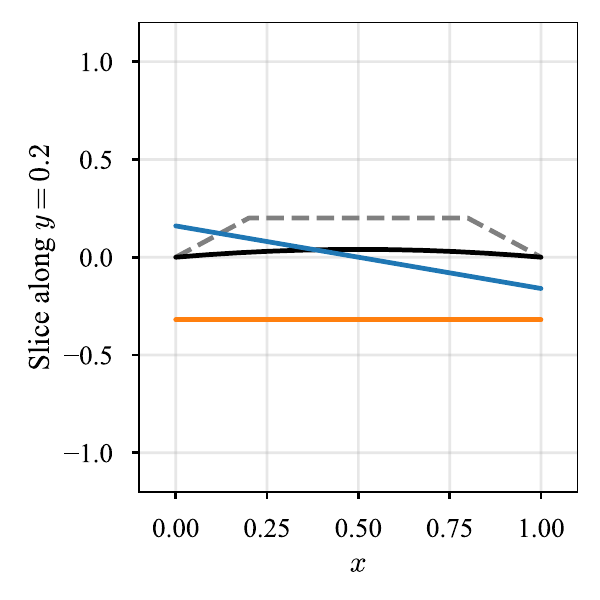}
        \subcaption{Na\"ive, $\psi$}
    \end{minipage}
    \begin{minipage}[b]{.24\linewidth}
        \centering
        \includegraphics[width=.99\linewidth]{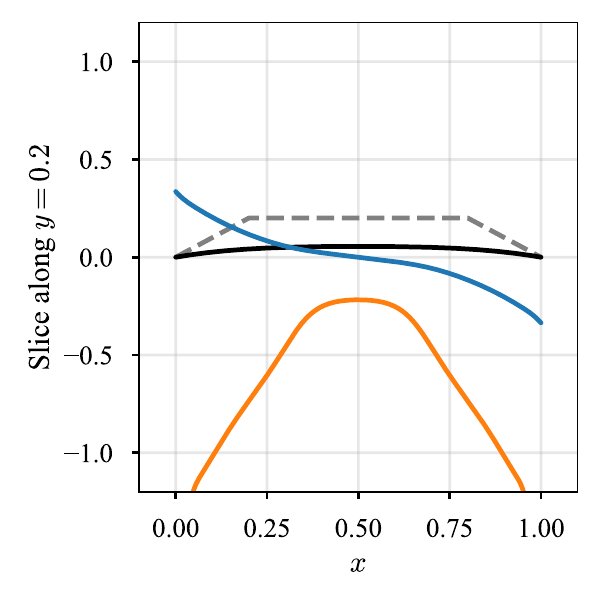}
        \subcaption{Non-normalized, $\varphi$}
    \end{minipage}
    \begin{minipage}[b]{.24\linewidth}
        \centering
        \includegraphics[width=.99\linewidth]{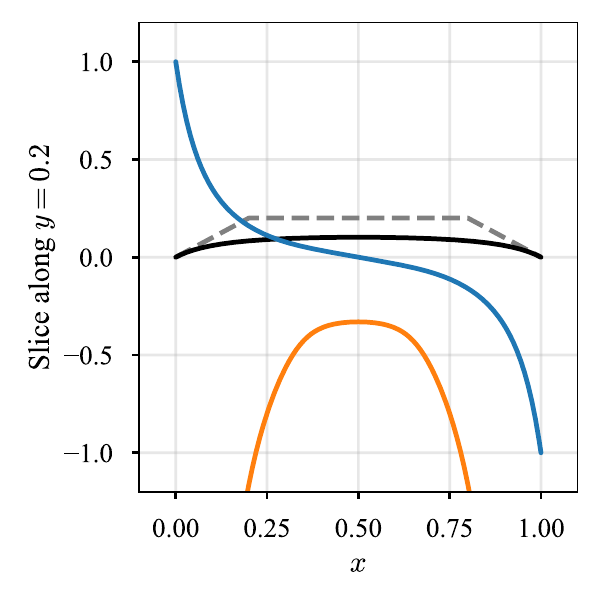}
        \subcaption{Normalized, $\phi^{(1)}$}
    \end{minipage}
    \begin{minipage}[b]{.24\linewidth}
        \centering
        \includegraphics[width=.99\linewidth]{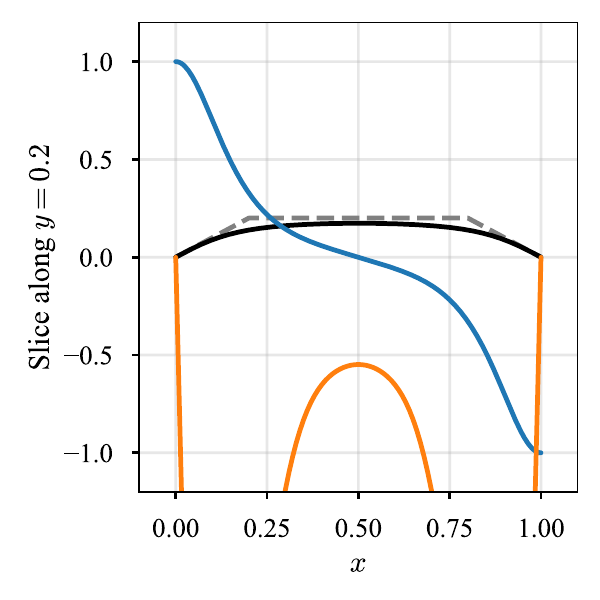}
        \subcaption{Normalized, $\phi^{(2)}$}
    \end{minipage}
    \caption{
        \emph{Distance to the boundary: square (slice).}
        (Top) slice along $y = 0.5$.
        (Bottom) slice along $y = 0.2$.
        The first derivative of the normalized ADFs form the inward unit normal at the boundary (property (b)),
        whereas the na\"ive and non-normalized ADFs fail to do so.
        Simply scaling the na\"ive or non-normalized ADF ($\psi$ or $\varphi$)
        by a constant does not resolve this issue.
    }
    \label{fig:square_slice}
\end{figure}

Assuming that the boundary $\Gamma$ is composed of piecewise segments ($\Gamma = \bigcup_{i} \Gamma_{i}$),
the normalized ADF to $\Gamma$ is constructed via the following joining operation~\cite{Biswas2004,Millan2015}:
\begin{equation}
    \phi^{(m)}
    \coloneq \frac{\prod_{i} \phi_i}{\left( \sum_{i} \phi_{i}^{m} \right)^{1/m}}
    = \frac{1}{\left( \sum_{i} \phi_{i}^{-m} \right)^{1/m}},
    \label{eq:normalized_adf}
\end{equation}
where $\phi_i$ is the ADF to $\Gamma_i$ ($\phi_i = 0$ on $\Gamma_i$, as defined in Equation~\eqref{eq:approximate_distance}),
and $m \left( \in \mathbb{Z}_{>0} \right)$ is the normalization order associated with the property (c).
Equation~\eqref{eq:normalized_adf} is formally known as the R-equivalence operation.
While another joining, the R-conjunction operation, is also available,
we employ the R-equivalence operation in this study
due to its associative property, which the R-conjunction lacks.
Figure~\ref{fig:square} shows the ADFs with different normalization orders and the EDF to a unit square.
As depicted, the ADFs serve as smooth approximations of the EDF.
The ADFs $\phi^{(m)}$ approach the EDF $\Phi$ as the normalization order $m$ increases, as described in~\cite{Biswas2004}.
This is further confirmed by comparing $\phi^{(64)}$ and $\Phi$ in Figure~\ref{fig:square}.
It is also noteworthy that the ADF obtained by Equation~\eqref{eq:normalized_adf} generalizes to
curves~\cite{Biswas2004},
complex non-convex geometries~\cite{Shapiro2002SAGE,Tsukanov2003MeshfreeFluid},
and higher-dimensional field
such as 3D objects~\cite{Biswas2004,Shapiro2007} and 4D cubes~\cite{Sukumar2022MixedBC}.

The choice of the normalization order $m$ in Equation~\eqref{eq:normalized_adf} is arbitrary,
and $m = 2$ is commonly used in computational mechanics practice~\cite{Millan2015}.
Our preliminary numerical experiments using $m = 1, 2, 4, 8$ found that
$m = 1$ and $m = 2$ achieve similar accuracy and convergence,
while $m = 4$ and $m = 8$ can degrade the convergence speed (see Appendix~\ref{sec:appendix_distance}).
Reflecting this observation and the previous studies~\cite{Sukumar2022MixedBC,Berrone2023BCEnforcement},
we use $m = 1$ for the remainder of this study.


\paragraph{Distance to the boundary: potential alternatives}  \label{sec:distance_domain_alternative}
As outlined in Section~\ref{sec:introduction},
several alternatives could be considered to realize the distance to the boundary $\Gamma = \bigcup_{i} \Gamma_{i}$.
In particular, the community has favored the na\"ive approach ((3-i), as introduced in Section~\ref{sec:introduction}),
under the assumption that the domain is convex (see, e.g.,~\cite{Lagaris1998,Lu2021HardDesign,Li2024HardAdvcDiff}).
In this formulation, the domain is detected as the intersection of half-spaces,
each of which represented by the SDF to an infinite line (or plane) containing $\Gamma_i$.
For instance, in the case of a unit square, the domain is represented by the intersection of four half-spaces:
\begin{equation}
    \psi
    \coloneq \prod_{i} s_i,
    \label{eq:naive_adf}
\end{equation}
where $\psi$ is the na\"ive ADF,
and each $s_i$ is defined as in Equation~\eqref{eq:signed_distance},
namely $s_1 = x$, $s_2 = 1 - x$, $s_3 = y$, and $s_4 = 1 - y$.
This formulation can be extended to other polygons and polyhedra, as long as they are convex.
Nonetheless, it is insufficient for non-convex geometries, as it does not account for segment endpoints (or plane edges).
Figure~\ref{fig:non_convex}~(a, top) illustrates
the distance field generated by Equation~\eqref{eq:naive_adf} for an L-shaped domain,
where $s_5 = 0.5 - x$ and $s_6 = 0.5 - y$ are included to the previous four $s_i$.
One can observe that false boundary detections (zero level sets) appear inside the domain,
due to the presence of a concave corner (specifically, $s_5$ and $s_6$).
The corresponding gradients also exhibit inconsistent scale and direction on the boundary, violating property (b).
We further applied this approach to represent a heart-shaped geometry (Figure~\ref{fig:non_convex}~(a, bottom)),
by defining SDFs for the top humps and the bottom cusp separately, and combining them through Equation~\eqref{eq:naive_adf}.
Despite this careful construction,
the resulting distance field shows a sudden jump inside the domain, undefined regions outside the arcs, and undesired boundary detections.
In addition to these geometric shortcomings,
applying this na\"ive formulation to Neumann and mixed conditions
remain challenging due to the lack of a well-defined unit normal on the boundary.
These observations suggest that
while the na\"ive approach (3-i) may suffice for convex domains with relatively simple BCs, such as homogeneous Dirichlet conditions,
it is unsuitable for complex geometries and a wide range of boundary value problems.

As another alternative, one may consider:
\begin{equation}
    \varphi
    \coloneq \prod_{i} \phi_i,
    \label{eq:non_normalized_adf}
\end{equation}
where $\varphi$ denotes the non-normalized ADF,
and each $\phi_i$ is defined in Equation~\eqref{eq:approximate_distance}.
By using $\phi_i$ instead of $s_i$,
this formulation is aware of segment endpoints (plane edges)
and can therefore represent non-convex geometries, as shown in Figure~\ref{fig:non_convex}~(b).
Equation~\eqref{eq:non_normalized_adf} is known as another class of R-equivalence operation,
and $\varphi$ satisfies $\varphi = 0$ on $\Gamma$ (property (a)).
However, it should be noted that this ADF does not satisfy property (b) ($\partial_{\bm{\nu}} \varphi \neq 1$ on $\Gamma$, see Figure~\ref{fig:non_convex}~(b, top)),
and is therefore called non-normalized in the framework of R-functions~\cite{Biswas2004}.
In summary,
this non-normalized formulation (3-ii) offers greater geometric flexibility than the na\"ive approach (3-i),
but remains limited in enforcing various boundary conditions.

To further assess whether the ADFs
in Equations~\eqref{eq:naive_adf},~\eqref{eq:non_normalized_adf}, and~\eqref{eq:normalized_adf}
satisfy properties (a), (b), and (c),
Figure~\ref{fig:square_slice} compares
na\"ive, non-normalized, and normalized distance fields (to the first and second orders),
and their derivatives for a unit square (slices along $y = 0.5$ and $y = 0.2$ are shown).
Both $\phi^{(1)}$ and $\phi^{(2)}$ (formulation (3-iii), Equation~\eqref{eq:normalized_adf}) satisfy properties (a) and (b),
with $\phi^{(2)}$ also satisfying (c) up to the second order.
In contrast, $\psi$ ((3-i), Equation~\eqref{eq:naive_adf})
and $\varphi$ ((3-ii), Equation~\eqref{eq:non_normalized_adf}) only meet property (a).
Moreover, from Figure~\ref{fig:square_slice}~(a) and (b),
one can see that simply scaling $\psi$ or $\varphi$ by a constant,
as done in~\cite{Lu2021HardDesign}, does not help to enforce property (b).

Based on these observations,
we employ Equation~\eqref{eq:normalized_adf} in this study
to define the ADFs and unit normal vectors on the boundary,
rather than Equations~\eqref{eq:naive_adf} or~\eqref{eq:non_normalized_adf},
in order to address mixed boundary value problems.
Therefore, we use the term `hard imposition' to refer specifically to the normalized formulation,
i.e., Equation~\eqref{eq:normalized_adf} (approach (3-iii)) in the remainder of this paper.
This choice becomes further beneficial for imposing inhomogeneous Dirichlet and mixed boundary conditions,
as discussed in the following section.


\paragraph{Approximate solution structure}
\begin{figure}[tpb]
    \centering
    \begin{minipage}[b]{.49\linewidth}
        \centering
        \includegraphics[width=.8\linewidth]{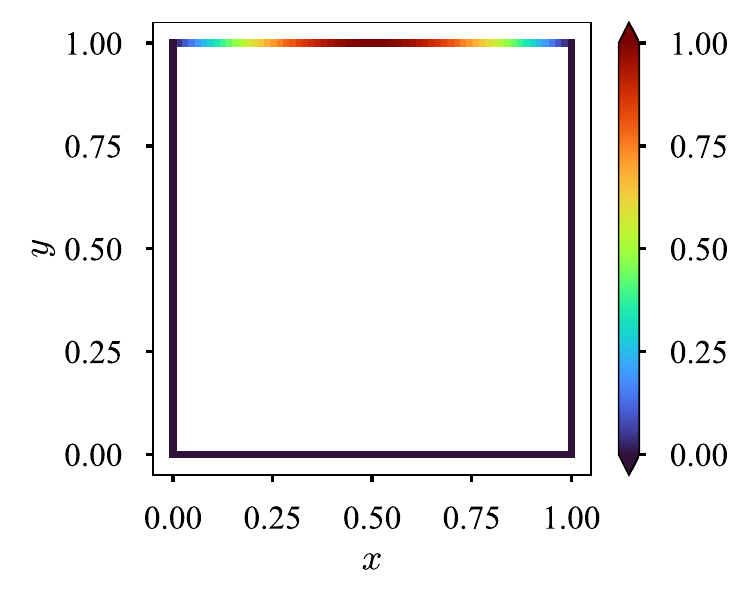}
        \subcaption{Dirichlet condition, $g_D: \Gamma_D \to \mathbb{R}$}
    \end{minipage}
    \begin{minipage}[b]{.49\linewidth}
        \centering
        \includegraphics[width=.8\linewidth]{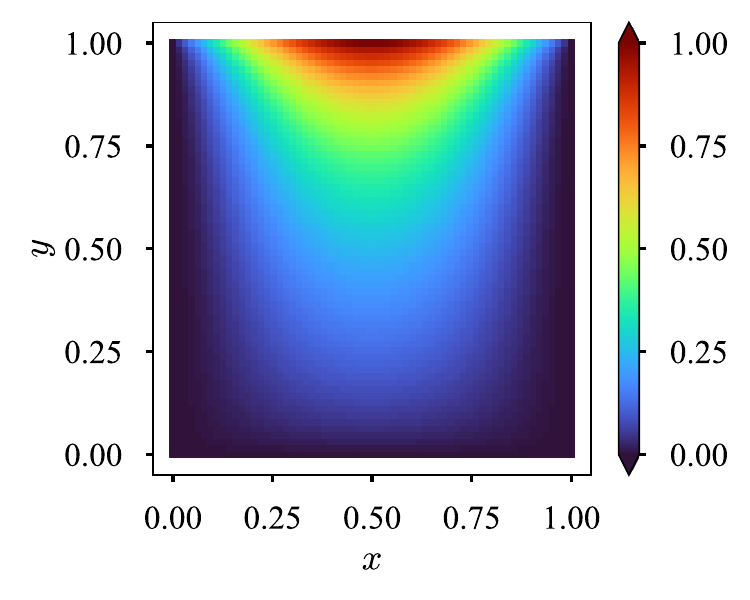}
        \subcaption{Interpolant, $\bar{g}_D: \bar{\Omega} \to \mathbb{R}$}
    \end{minipage}
    \caption{
        \emph{Transfinite interpolation of Dirichlet boundary condition.}
        With the ADF $\phi$ vanishing on the boundary,
        the trial function $\tilde{u} = \bar{g}_D + \phi \hat{u}$
        exactly satisfies the Dirichlet boundary condition, $g_D$.
    }
    \label{fig:transfinite_interpolation}
\end{figure}

For simplicity, we only discuss the treatment of Dirichlet boundary condition.
However, the method can be extended to Neumann, Robin, and mixed boundary conditions,
as described in~\cite{Sukumar2022MixedBC,Shapiro2002SAGE,Tsukanov2003MeshfreeFluid,Rvachev2000}
and demonstrated in Section~\ref{sec:results}.

As mentioned in Section~\ref{sec:introduction},
boundary conditions can be imposed in either soft~\cite{Raissi2019PINN} or hard manner~\cite{Lagaris1998,Sun2020Surrogate,Sukumar2022MixedBC},
in the context of PINN.
In the soft imposition approach, the raw output of the neural network serves as the approximate solution:
\begin{equation}
    u
    \simeq \mathcal{I} \left( \hat{u} \right)
    \coloneqq \hat{u},
    \label{eq:soft_imposition}
\end{equation}
where $\mathcal{I} \left( \cdot \right)$ is the identity operator.
Since no explicit treatment is applied and the neural network $\hat{u}$ is not constrained,
boundary conditions are enforced through the additional penalty terms in the loss function.

In contrast, the hard imposition approach directly enforces the boundary condition by constructing
the approximate solution through the following projection:
\begin{equation}
    u
    \simeq \mathcal{P} \left( \hat{u} \right)
    \coloneqq \tilde{u}
    \coloneqq \bar{g}_{D} + \phi_{D} \hat{u},
    \label{eq:hard_imposition}
\end{equation}
where $\bar{g}_{D}$ is an interpolation of the Dirichlet condition ($\bar{g}_{D} = g_{D}$ on $\Gamma_{D}$),
and $\phi_{D}$ is the approximate distance function (ADF) to $\Gamma_{D}$.
With $\phi_{D}$ vanishing on $\Gamma_{D}$,
$\tilde{u}$ readily satisfies the imposed Dirichlet boundary condition.
When the Dirichlet boundary $\Gamma_{D}$ consists of multiple segments $\Gamma_{i}$ ($i = 1, \ldots, N_{D}$)
(i.e., Dirichlet conditions are prescribed on multiple segments),
the interpolant $\bar{g}_{D}$ can be constructed by transfinite interpolation~\cite{Rvachev2001}:
\begin{align}
    \bar{g}_{D}
    &= \sum_{i} w_i g_{i},
    \\
    w_i
    &= \frac{\phi_i^{-\mu_i}}{\sum_j \phi_j^{-\mu_j}}
    = \frac{\prod_{j; j \neq i} \phi_j^{\mu_j}}{\sum_k \prod_{j; j \neq k} \phi_j^{\mu_j}},
\end{align}
where $g_{i}$ is the Dirichlet BC prescribed on $\Gamma_{i}$,
$\mu_i$ is the interpolation order ($\bar{g}_{D}$ is $(\mu_i - 1)$-times continuously differentiable on $\Gamma_{i}$),
and $w_i$ forms a partition of unity.

For example, given the Dirichlet conditions:
\begin{alignat}{2}
    g_{1} &= 0            && \quad \text{on} \quad \Gamma_{1} = \{ (x, 0) \mid x \in (0, 1) \}, \label{eq:Gamma_1} \\
    g_{2} &= 0            && \quad \text{on} \quad \Gamma_{2} = \{ (1, y) \mid y \in (0, 1) \}, \label{eq:Gamma_2} \\
    g_{3} &= \sin (\pi x) && \quad \text{on} \quad \Gamma_{3} = \{ (x, 1) \mid x \in (0, 1) \}, \label{eq:Gamma_3} \\
    g_{4} &= 0            && \quad \text{on} \quad \Gamma_{4} = \{ (0, y) \mid y \in (0, 1) \}, \label{eq:Gamma_4}
\end{alignat}
the corresponding interpolant is shown in Figure~\ref{fig:transfinite_interpolation},
demonstrating how $g_{D}$ is smoothly extended to $\bar{g}_{D}$.
Notably, the projection in Equation~\eqref{eq:hard_imposition} restricts the admissible solution space,
ensuring that the trial function is selected from
$U = \lbrace u: \Omega \to \mathbb{R} \mid u = g_{D} \text{ on } \Gamma_{D} \rbrace$
which is analogous to the approach used in classical numerical methods such as the Galerkin method.
This direct enforcement of boundary conditions allows for improved accuracy and stability
compared to the penalty-based soft imposition approach.


\subsubsection{Adaptive weight tuning for gradient imbalance}  \label{sec:adaptive_weight}
The formulation described in Section \ref{sec:r_function} ensures that
the approximate solution inherently satisfies the prescribed boundary conditions.
Consequently, $\mathcal{L}_{\text{BC}}$ in Equation~\eqref{eq:loss} becomes identically zero.
This implies that, for forward problems, minimizing $\mathcal{L}_{\text{PDE}}$ alone is sufficient,
akin to classical numerical methods.
However, in inverse problems, PINN must incorporate observed data through Equation~\eqref{eq:loss_data}.
In this case, the loss function with hard boundary condition enforcement takes the form:
\begin{equation}
    \mathcal{L} (\bm{\theta})
    = \tilde{\mathcal{L}}_{\text{PDE}} (\bm{\theta})
        + \lambda_{\text{Data}} \tilde{\mathcal{L}}_{\text{Data}} (\bm{\theta}),
    \label{eq:loss_hard}
\end{equation}
where $\tilde{\mathcal{L}}_{\text{PDE}}$ and $\tilde{\mathcal{L}}_{\text{Data}}$
are the modified PDE and data loss functions, respectively,
obtained by replacing $\hat{u}$ with $\tilde{u}$ in Equations~\eqref{eq:loss_pde} and~\eqref{eq:loss_data}.

A critical challenge in optimizing multi-objective functions, such as Equation~\eqref{eq:loss_hard},
is the risk of convergence to undesired, suboptimal Pareto solutions due to gradient imbalance,
which can further lead to the emergence of trivial, constant solutions, known as failure modes~\cite{Krishnapriyan2021FailureModes,Rohrhofer2023IEEE,Deguchi2023DynNorm,Daw2023PropagationFailures}.
In such cases, appropriate normalization is crucial to balance the gradients of multiple loss terms.
In the context of PINN, the PDE and data loss terms are typically of different magnitudes,
and their gradients may not be directly comparable.
To address this issue,
we employ an adaptive weight tuning method known as dynamic normalization~\cite{Deguchi2023DynNorm}.
This method adaptively adjusts the weight $\lambda_{\text{Data}}$ in Equation~\eqref{eq:loss_hard}
to balance the gradients of each loss term during the process of optimization, eliminating the need for manual tuning.

We first consider the first-order Taylor series expansion of the loss function around the current parameter $\bm{\theta}^{(n)}$:
\begin{equation}
    \mathcal{L} (\bm{\theta}^{(n)} + \Delta \bm{\theta}) - \mathcal{L} (\bm{\theta}^{(n)})
    = \Delta \bm{\theta}^{\top} \nabla_{\bm{\theta}} \mathcal{L} (\bm{\theta}^{(n)}),
    \label{eq:loss_taylor}
\end{equation}
where $\Delta \bm{\theta} = \bm{\theta}^{(n+1)} - \bm{\theta}^{(n)}$ represents the update.
In gradient descent, the update is given by:
\begin{equation}
    \begin{split}
        \Delta \bm{\theta}
        &= \bm{\theta}^{(n+1)} - \bm{\theta}^{(n)} \\
        &= - \eta^{(n)} \nabla_{\bm{\theta}} \mathcal{L} (\bm{\theta}^{(n)}),
    \end{split}
\end{equation}
where $\eta^{(n)}$ is the learning rate.
Substituting this into Equation~\eqref{eq:loss_taylor} yields:
\begin{equation}
    \Delta \tilde{\mathcal{L}}_{\text{PDE}} + \Delta \tilde{\mathcal{L}}_{\text{Data}}
    = - \eta^{(n)} \| \nabla_{\bm{\theta}} \tilde{\mathcal{L}}_{\text{PDE}} + \lambda_{\text{Data}} \nabla_{\bm{\theta}} \tilde{\mathcal{L}}_{\text{Data}} \|^{2}.
    \label{eq:loss_taylor2}
\end{equation}
Here, $\Delta \tilde{\mathcal{L}}_{\text{PDE}}$ and $\Delta \tilde{\mathcal{L}}_{\text{Data}}$
denote the changes in the PDE and data loss terms, respectively.
For simplicity, we assume that each gradient is orthogonal to the other.
This assumption may not strictly hold,
as $\bm{\theta}$ is shared across multiple objectives and gradient correlations can arise.
Nonetheless, it offers a tractable simplification to facilitate analysis,
and prior studies have shown that correcting magnitude imbalances
typically plays a more critical role in practice~\cite{Wang2021GradPath,Maddu2022InvDirichlet,Rohrhofer2021arXiv}.
If directional conflicts become pronounced or pose significant challenges in optimization,
projection techniques such as those presented in~\cite{Yu2020PCGrad,Li2023AGDA} can be employed to mitigate them.
Under the aforementioned orthogonality assumption, we split Equation~\eqref{eq:loss_taylor2} into:
\begin{align}
    \Delta \tilde{\mathcal{L}}_{\text{PDE}}
    &= - \eta^{(n)} \| \nabla_{\bm{\theta}} \tilde{\mathcal{L}}_{\text{PDE}} \|^{2},
    \label{eq:loss_taylor_pde}
    \\
    \Delta \tilde{\mathcal{L}}_{\text{Data}}
    &= - \eta^{(n)} \lambda_{\text{Data}}^{2} \| \nabla_{\bm{\theta}} \tilde{\mathcal{L}}_{\text{Data}} \|^{2}.
    \label{eq:loss_taylor_data}
\end{align}
Based on the analysis in~\cite{Wang2021GradPath,Rohrhofer2023IEEE},
it is desirable to balance the reduction of the PDE and data loss
such that they are reduced at similar rates, i.e., $\Delta \tilde{\mathcal{L}}_{\text{PDE}} \sim \Delta \tilde{\mathcal{L}}_{\text{Data}}$.
This translates to the following adaptive weight tuning strategy:
\begin{equation}
    \lambda_{\text{Data}}^{(n)}
    = \frac{\| \nabla_{\bm{\theta}} \tilde{\mathcal{L}}_{\text{PDE}} \|}{\| \nabla_{\bm{\theta}} \tilde{\mathcal{L}}_{\text{Data}} \|}.
    \label{eq:weight1}
\end{equation}
In practice, the exponential decay is applied to stabilize the fluctuations~\cite{Maddu2022InvDirichlet,Wang2021GradPath,Wang2023ExpertGuide}:
\begin{equation}
    \hat{\lambda}_{\text{Data}}^{(n)}
    = \beta \hat{\lambda}_{\text{Data}}^{(n-1)} + (1 - \beta) \lambda_{\text{Data}}^{(n)},
    \label{eq:weight2}
\end{equation}
where $\beta \in [0, 1)$ is the decay rate.
Our preliminary numerical experiments indicate that
$\beta$ should be set to a value close to 1 to ensure stability.
However, as is well known, setting $\beta$ too close to 1 can lead to slow convergence.
As an illustration, applying Equation~\eqref{eq:weight2} recursively, one obtains:
\begin{align}
    \hat{\lambda}_{\text{Data}}^{(n)}
    &= \beta^{n} \hat{\lambda}_{\text{Data}}^{(0)} + (1 - \beta) \sum_{k=1}^{n} \beta^{n-k} \lambda_{\text{Data}}^{(k)}, \\
    \begin{split}
        \therefore
        \mathbb{E} \left[ \hat{\lambda}_{\text{Data}}^{(n)} \right]
        &= \beta^{n} \hat{\lambda}_{\text{Data}}^{(0)}
            + (1 - \beta) \sum_{k=1}^{n} \beta^{n-k} \mathbb{E} \left[ \lambda_{\text{Data}}^{(k)} \right] \\
        &= \beta^{n} \hat{\lambda}_{\text{Data}}^{(0)}
            + \mathbb{E} \left[ \lambda_{\text{Data}}^{(n)} \right] (1 - \beta) \sum_{k=1}^{n} \beta^{n-k} \\
        &= \beta^{n} \hat{\lambda}_{\text{Data}}^{(0)}
            + \mathbb{E} \left[ \lambda_{\text{Data}}^{(n)} \right] (1 - \beta^{n}).
    \end{split}
\end{align}
One can instantly realize that $\hat{\lambda}_{\text{Data}}^{(n)}$
is biased towards its initial value $\hat{\lambda}_{\text{Data}}^{(0)}$,
particularly when $\beta$ is close to 1 and $n$ is not sufficiently large.
To remove this initialization bias, we set $\hat{\lambda}_{\text{Data}}^{(0)} = 0$
and introduce bias correction as in~\cite{Kingma2015Adam}:
\begin{equation}
    \tilde{\lambda}_{\text{Data}}^{(n)}
    = \frac{\hat{\lambda}_{\text{Data}}^{(n)}}{1 - \beta^{n}},
    \label{eq:weight3}
\end{equation}
ergo $\mathbb{E} [ \tilde{\lambda}_{\text{Data}}^{(n)} ] = \mathbb{E} [ \lambda_{\text{Data}}^{(n)} ]$,
whereas $\mathbb{E} [ \hat{\lambda}_{\text{Data}}^{(n)} ] \neq \mathbb{E} [ \lambda_{\text{Data}}^{(n)} ]$.
Equation~\eqref{eq:weight3} ensures that the weight is properly adjusted,
even in the early stages of optimization and when $\beta$ is near 1.
Despite the presence of numerous studies on adaptive weight tuning in the context of PINN
~\cite{Wang2021GradPath,Maddu2022InvDirichlet,Jin2021NSFnets,Wang2023ExpertGuide},
discussion on this initialization bias is often overlooked.
Bias correction technique in Equation~\eqref{eq:weight3} is crucial for the appropriate adjustment,
ensuring that $\Delta \tilde{\mathcal{L}}_{\text{PDE}} \sim \Delta \tilde{\mathcal{L}}_{\text{Data}}$ is satisfied from the initial stage.
The effectiveness of bias-corrected dynamic normalization has already been demonstrated in~\cite{Deguchi2023DynNorm,Deguchi2023Keisan}
for both forward and inverse problems, and is further confirmed in Section~\ref{sec:results}.


\subsubsection{Positivity enforcement for inverse analysis}  \label{sec:positivity_enforcement}
When applied to inverse problems,
PINNs are typically equipped with additional trainable parameters
to represent the physical quantities of interest.
For instance, when identifying the diffusion coefficient $\kappa$ in a diffusion equation,
an additional parameter $\hat{\kappa}$ is introduced to represent and estimate $\kappa$
from the available observations.
This approach has been widely used in the literature,
as exemplified in~\cite{Raissi2019PINN,Kharazmi2021hpVPINN,Jagtap2020CPINN,Lu2021DeepXDE}.
Since $\hat{\kappa}$ is a trainable parameter,
it is updated alongside the network parameters (weights and biases).
This allows $\hat{\kappa}$ to take any real value, including negative values.
Such an approach can be viewed as a class of soft imposition,
as the sign of $\hat{\kappa}$ is not explicitly constrained
and is only expected to be learned through the optimization process.
From a physical point of view, however,
these quantities must possess specific signs to ensure physical validity.
In the case of diffusion example, the diffusion coefficient $\kappa$ must remain positive.

We address this issue by enforcing the positivity of the physical quantities
through the following transformation:
\begin{equation}
    \kappa
    \simeq \tilde{\kappa}
    \coloneqq h_{+} (\hat{\kappa}).
    \label{eq:positivity_enforcement}
\end{equation}
Here, $h_{+} (\cdot)$ is a bijective function that maps any real value to a positive real realm
($h_{+}: \mathbb{R} \to \mathbb{R}_{+}$).
We refer to this transformation as positivity enforcement.
In this study, we utilize the exponential function $\exp (\cdot)$ as $h_{+} (\cdot)$,
since it showed superior convergence in the prior experiments,
where we compared against softplus and rectified power unit, $\text{RePU}_{p=2} = \max (0, \cdot)^{2}$.
It is important to note that
this transformation is not limited to scalar parameters
but can also be applied to space-dependent physical quantities,
ensuring physically meaningful estimations across various applications.


\section{Numerical results}   \label{sec:results}
This section presents the results of several numerical experiments.
Throughout the experiments, we employed MLPs with a depth of 5 and a width of 64,
initialized by the Glorot method~\cite{Glorot2010}.
This choice is based on prior studies demonstrating its effectiveness across a broad range of problems~\cite{Sun2020Surrogate,Jin2021NSFnets,Rohrhofer2023IEEE}.
The Adam optimizer~\cite{Kingma2015Adam} was used for training,
with exponential decay rates of 0.9 and 0.999 for the first and second moment estimates, respectively.

As is well known, the choice of the activation function
significantly affects the convergence and stability.
Common choices include the hyperbolic tangent function
and the rectified linear unit (ReLU, $\max (0, \cdot)$).
However, the ReLU and some of its variants~\cite{Maas2013LeakyReLU,He2015,Clevert2016ELU}
are generally unsuitable for PINNs because their higher-order derivatives vanish over a wide range of inputs.
This limitation hinders the accurate computation of higher-order derivatives of the solution,
leading to inaccurate PDE residual calculations~\cite{Deguchi2021Doboku}.
Several studies have also reported numerical results that the ReLU leads to poor convergence in PINNs~\cite{Jagtap2020GAAF,Haghighat2021SolidMechanics}.
Recent research has proposed the rectified power unit ($\text{RePU}_{p} = \max (0, \cdot)^{p}, p \in \mathbb{Z}_{\ge 2}$)
as an alternative~\cite{Sukumar2022MixedBC,Samaniego2020Energy}.
However, RePU networks are prone to vanishing gradients for inputs less than 1
and exploding gradients for inputs greater than 1, particularly in deeper networks.
Indeed, Sukumar and Srivastava~\cite{Sukumar2022MixedBC} and Samaniego et al.~\cite{Samaniego2020Energy}
utilized RePU networks with a depth of 3,
which is relatively shallow in the literature.
Our preliminary experiments also confirmed that the RePU network with increased depth (4, 5, and 6)
exhibited vanishing and exploding gradient issues, leading to instability.
Consequently, we evaluated and compared the hyperbolic tangent,
Sigmoid Linear Unit (SiLU)~\cite{Elfwing2017SiLU,Ramachandran2017Swish},
and Gaussian Error Linear Unit (GELU)~\cite{Hendrycks2016GELU} activation functions.
These functions are $C^{\infty}$-continuous and have non-vanishing higher-order derivatives,
which are essential for properly calculating the PDE residuals.

Furthermore, several techniques have been proposed to enhance PINN performance,
including adaptive activation functions~\cite{Jagtap2020GAAF,Jagtap2020LAAF},
gradient enhancement~\cite{Yu2022GradEnhc},
and self-adaptive soft attention mechanisms~\cite{McClenny2023SA}.
However, based on the authors' experience,
the accuracy improvements achieved by these methods are often marginal
and highly sensitive to hyperparameter selection
(see Appendix~\ref{sec:appendix_improvement} for numerical results).
Hence, we focus on the fundamental aspects of PINN,
such as boundary condition treatment and adaptive weight tuning.

We have implemented all the experiments using TensorFlow~\cite{Abadi2015TensorFlow}
\footnote{
    \url{https://github.com/ShotaDeguchi/PINN_HardBC_DynNorm}.
}.


\subsection{Forward problem: Poisson equation with mixed boundary conditions}   \label{sec:results_poisson}


\begin{figure}[tpb]
    \centering
    \begin{minipage}[b]{.99\linewidth}
        \centering
        \includegraphics[width=.99\linewidth]{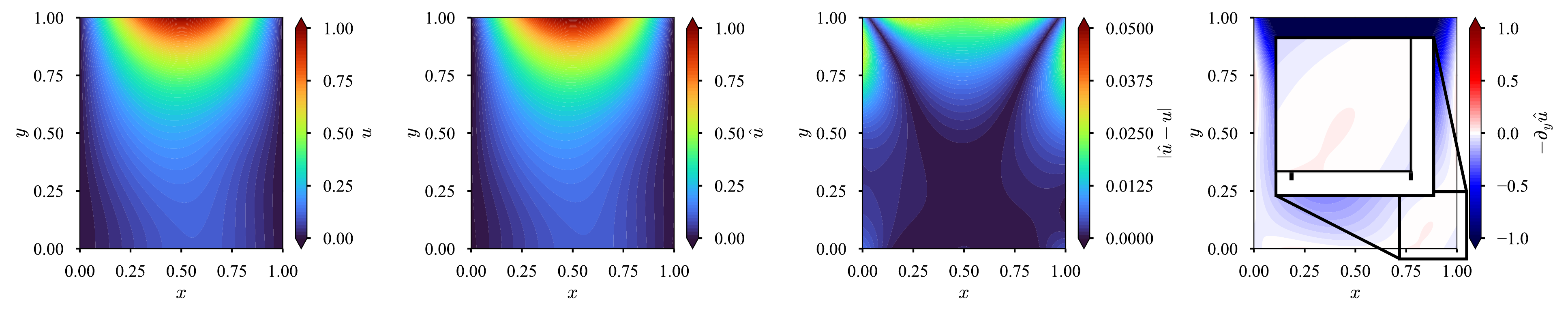}
        \subcaption{
            Soft imposition (weight parameter: $\lambda_{\text{BC}} = 1$).
            Relative $L^2$ error: $2.88 \times 10^{-2}$.
        }
    \end{minipage}
    \\
    \begin{minipage}[b]{.99\linewidth}
        \centering
        \includegraphics[width=.99\linewidth]{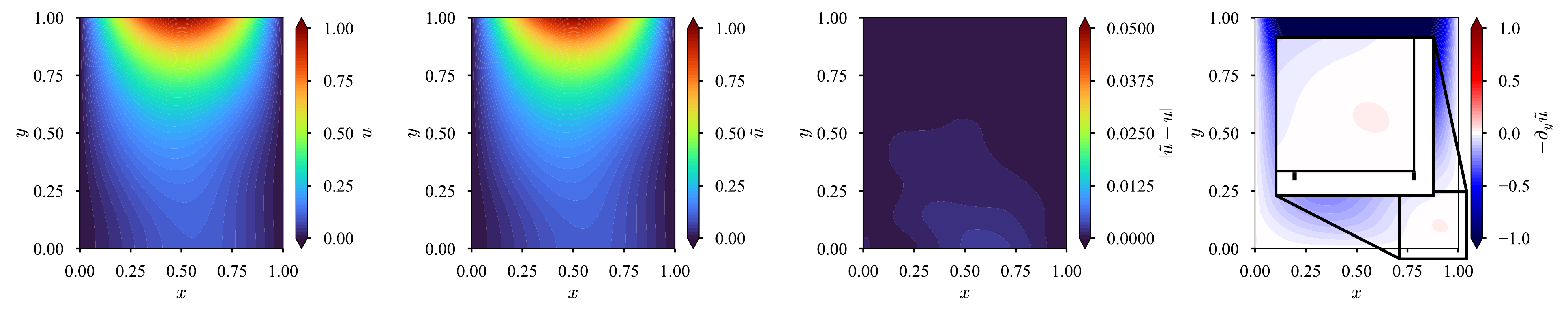}
        \subcaption{
            Hard imposition (normalization order: $m = 1$).
            Relative $L^2$ error: $3.13 \times 10^{-3}$.
        }
    \end{minipage}
    \caption{
        \emph{Poisson equation with mixed boundary conditions: homogeneous Neumann condition case.}
        (From left to right)
        Reference solution,
        approximate solution,
        absolute error against the reference solution,
        first derivative of the approximate solution along the negative $y$-axis.
        Results with GELU activation function.
        Soft imposition fails to enforce the Dirichlet (third column) and Neumann (fourth column) boundary conditions,
        while hard imposition successfully enforces both.
    }
    \label{fig:poisson_hmg_solution}
\end{figure}

\begin{figure}[tpb]
    \centering
    \begin{minipage}[b]{.49\linewidth}
        \centering
        \includegraphics[width=.99\linewidth]{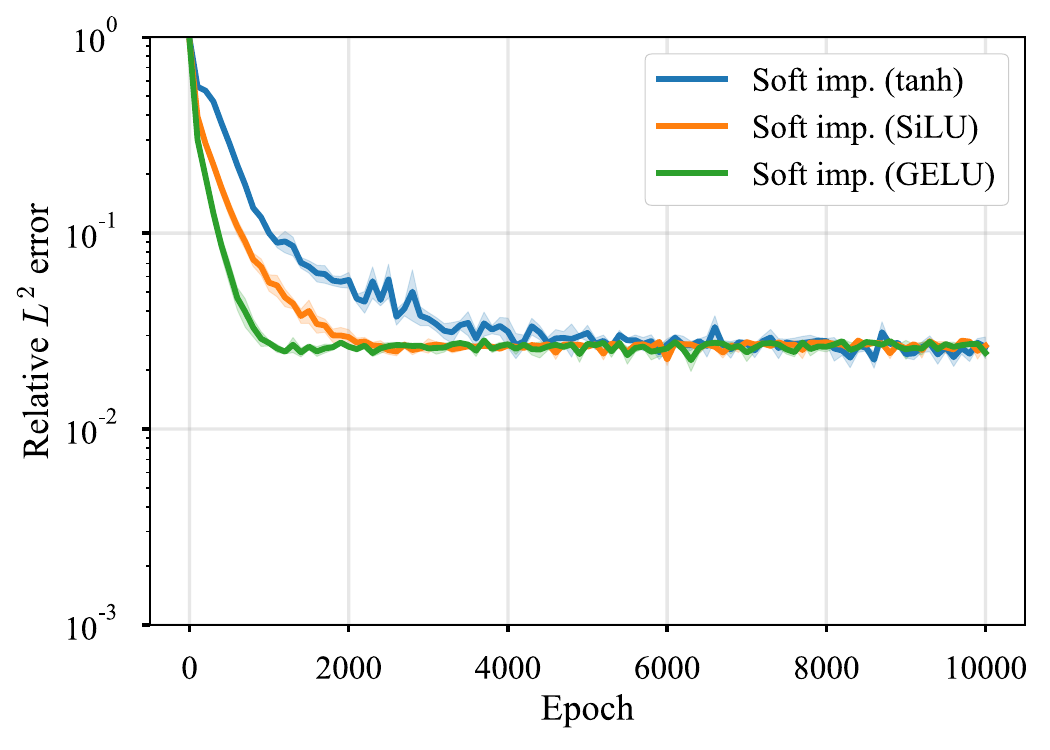}
        \subcaption{
            Soft imposition (weight parameter: $\lambda_{\text{BC}} = 1$)
        }
    \end{minipage}
    \begin{minipage}[b]{.49\linewidth}
        \centering
        \includegraphics[width=.99\linewidth]{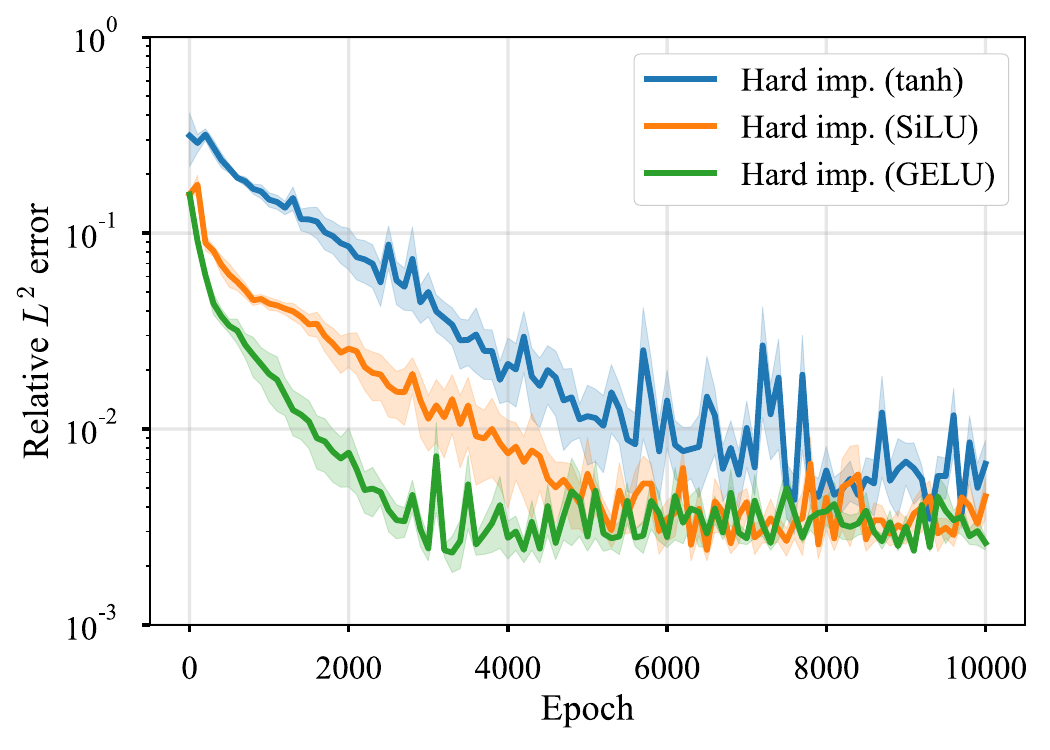}
        \subcaption{
            Hard imposition (normalization order: $m = 1$)
        }
    \end{minipage}
    \caption{
        \emph{Poisson equation with mixed boundary conditions: homogeneous Neumann condition case.}
        Relative $L^2$ error of the approximate solution with different activation functions.
        The solid line represents the mean, and the shaded area represents the standard error
        over 5 independent runs with i.i.d. initialization.
        Soft imposition quickly plateaus regardless of the activation function.
    }
    \label{fig:poisson_hmg_log}
\end{figure}

\begin{table}[tpb]
    \centering
    \caption{
        \emph{Poisson equation with mixed boundary conditions: homogeneous Neumann condition case.}
        Relative $L^2$ error of the approximate solution with different activation functions.
        Mean and standard error over 5 independent runs with i.i.d. initialization.
        All values are scaled by $10^{-3}$.
    }
    \label{tab:poisson_hmg_summary}
    \begin{tabular}{l|ccc}
        \toprule
        Activation function & Soft imposition  & Hard imposition \\
        \midrule
        tanh                & $26.26 \pm 3.23$ & $6.65 \pm 2.08$ \\
        SiLU                & $26.64 \pm 0.63$ & $4.54 \pm 1.05$ \\
        GELU                & $24.50 \pm 1.64$ & $2.62 \pm 0.21$ \\
        \bottomrule
    \end{tabular}
\end{table}

\paragraph{Homogeneous Neumann condition case}
We first consider the forward problem of the Poisson equation with mixed boundary conditions
to assess the effectiveness of soft and hard boundary condition impositions.
In addition, we compare the approximation performance of the three aforementioned activation functions: tanh, SiLU, and GELU.
The domain is a unit square $\Omega = (0, 1)^2$,
and the boundary is partitioned into Dirichlet and Neumann boundaries:
$\Gamma_{D} = \Gamma_{2} \cup \Gamma_{3} \cup \Gamma_{4}$ and
$\Gamma_{N} = \Gamma \setminus \Gamma_{D}$
(identical to Equations~\eqref{eq:Gamma_1}--\eqref{eq:Gamma_4},
except for Equation~\eqref{eq:Gamma_1} being replaced by the Neumann boundary).
The source and the boundary conditions are set as follows:
\begin{align}
    f
    &= \sin \left( 2 \pi (x + y) \right),
    \\
    g_{D}
    &= \begin{cases}
        \sin \left( \pi x \right) & \text{on} \quad \Gamma_{3}, \\
        0                         & \text{on} \quad \Gamma_{2} \cup \Gamma_{4},
    \end{cases}
    \\
    g_{N}
    &= 0.
\end{align}
The reference solution was obtained using the finite difference method (FDM)
with a spatial resolution of $\Delta x = \Delta y = 5 \times 10^{-3}$.
Although the FDM solution contains numerical errors,
we treat it as the reference as it satisfies the boundary conditions precisely.
For both soft and hard boundary condition impositions,
the number of collocation points was set to $N_{\text{PDE}} = 4,096$.
Additionally, $N_{\text{DBC}} = 256 \times 3$ and $N_{\text{NBC}} = 256$ points
were added for the soft imposition.
The learning rate of the Adam optimizer was set to $10^{-3}$.
The weight parameter $\lambda_{\text{BC}}$ in the soft imposition was set to unity,
and the normalization order $m$ in the hard imposition was set to 1,
based on numerical tests in Appendix~\ref{sec:appendix_distance}.

Figure~\ref{fig:poisson_hmg_solution} presents
reference solution,
approximate solution,
absolute error,
and the first derivative of the approximate solution along the negative $y$-direction,
so that one can observe how well the Dirichlet and Neumann boundary conditions are satisfied.
Although the second column may appear similar,
the third and fourth columns reveal a significant difference.
The soft imposition fails to properly enforce the prescribed Dirichlet and Neumann boundary conditions,
whereas the hard imposition does so effectively, as evidenced by the absolute error and derivative plots.
The relative $L^2$ error further emphasizes the importance of boundary condition treatment.
The error of the hard imposition is $3.13 \times 10^{-3}$,
which is an order of magnitude lower than that of the soft imposition, $2.88 \times 10^{-2}$.
Figure~\ref{fig:poisson_hmg_log} shows the relative $L^2$ error of the approximate solution
with different activation functions and boundary condition impositions.
We observe from Figure~\ref{fig:poisson_hmg_log}~(a)
that the soft imposition exhibits a plateau regardless of the activation function,
which is known as pitfall of soft boundary condition enforcement~\cite{Sun2020Surrogate}.
This behavior is likely due to the presence of multiple competing objectives,
making it difficult for the network to learn the correct solutions~\cite{Rohrhofer2023IEEE}.
In contrast, the hard imposition demonstrates superior convergence,
achieving a relative error of $\mathcal{O} (10^{-3})$ (Figure~\ref{fig:poisson_hmg_log}~(b)).
Comparing the convergence behavior of different activation functions,
we find that both soft and hard impositions converge more rapidly with SiLU and GELU than with tanh.
This should be attributed to the non-saturating property of SiLU and GELU,
which allows the network to learn more effectively.
Moreover, GELU exhibits slightly faster convergence than SiLU,
which is likely due to its steeper gradient around the origin.
Furthermore, Table~\ref{tab:poisson_hmg_summary} summarizes the relative error
with different activation functions and boundary condition impositions.
One can observe that the hard imposition consistently outperforms the soft imposition
across all activation functions, confirming the effectiveness of the hard boundary condition imposition in PINN.


\begin{figure}[tpb]
    \centering
    \begin{minipage}[b]{.99\linewidth}
        \centering
        \includegraphics[width=.99\linewidth]{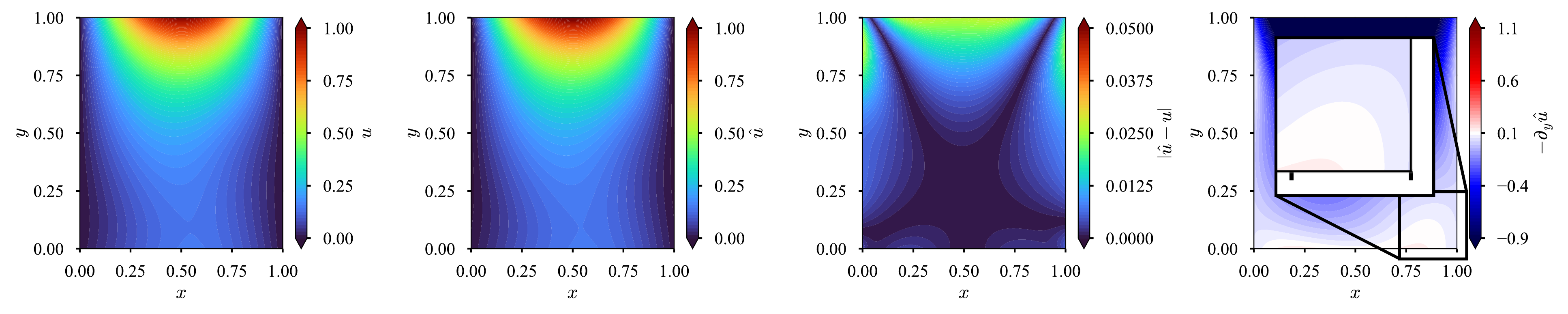}
        \subcaption{
            Soft imposition (weight parameter: $\lambda_{\text{BC}} = 1$).
            Relative $L^2$ error: $2.87 \times 10^{-2}$.
        }
    \end{minipage}
    \\
    \begin{minipage}[b]{.99\linewidth}
        \centering
        \includegraphics[width=.99\linewidth]{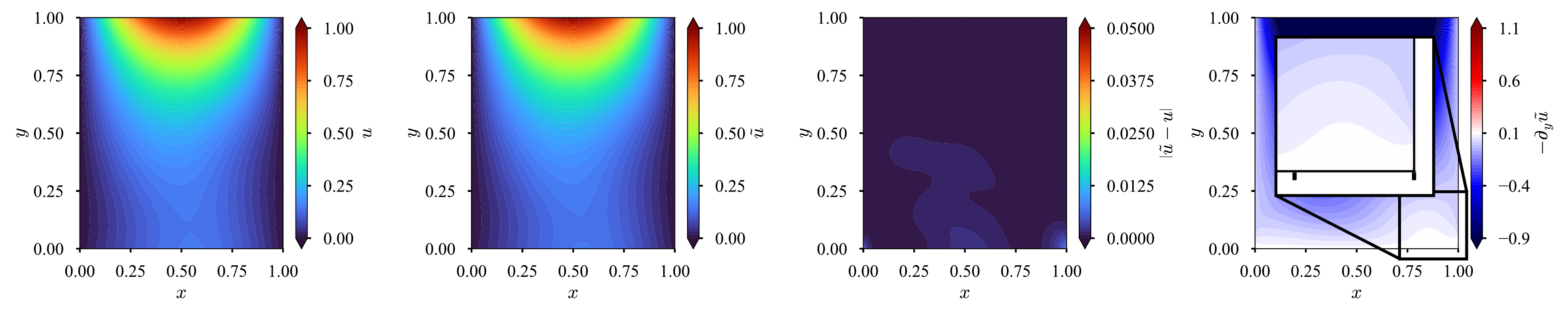}
        \subcaption{
            Hard imposition (normalization order: $m = 1$).
            Relative $L^2$ error: $2.11 \times 10^{-3}$.
        }
    \end{minipage}
    \caption{
        \emph{Poisson equation with mixed boundary conditions: inhomogeneous Neumann condition case.}
        (From left to right)
        Reference solution,
        approximate solution,
        absolute error against the reference solution,
        first derivative of the approximate solution along the negative $y$-axis.
        Results with GELU activation function.
        Again, soft imposition struggles to enforce the both Diriichlet and Neumann boundary conditions,
        which can be seen from the third and fourth columns.
    }
    \label{fig:poisson_inhmg_solution}
\end{figure}

\begin{table}[tpb]
    \centering
    \caption{
        \emph{Poisson equation with mixed boundary conditions: inhomogeneous Neumann condition case.}
        Relative $L^2$ error of the approximate solution with different activation functions.
        Mean and standard error over 5 independent runs with i.i.d. initialization.
        All values are scaled by $10^{-3}$.
    }
    \label{tab:poisson_inhmg_summary}
    \begin{tabular}{l|ccc}
        \toprule
        Activation function & Soft imposition  & Hard imposition \\
        \midrule
        tanh                & $27.89 \pm 2.90$ & $13.23 \pm 7.42$ \\
        SiLU                & $27.25 \pm 1.00$ & $ 4.79 \pm 1.00$ \\
        GELU                & $27.59 \pm 0.77$ & $ 3.34 \pm 0.61$ \\
        \bottomrule
    \end{tabular}
\end{table}

\paragraph{Inhomogeneous Neumann condition case}
\begin{figure}[tpb]
    \centering
    \begin{minipage}[b]{.49\linewidth}
        \centering
        \includegraphics[width=.99\linewidth]{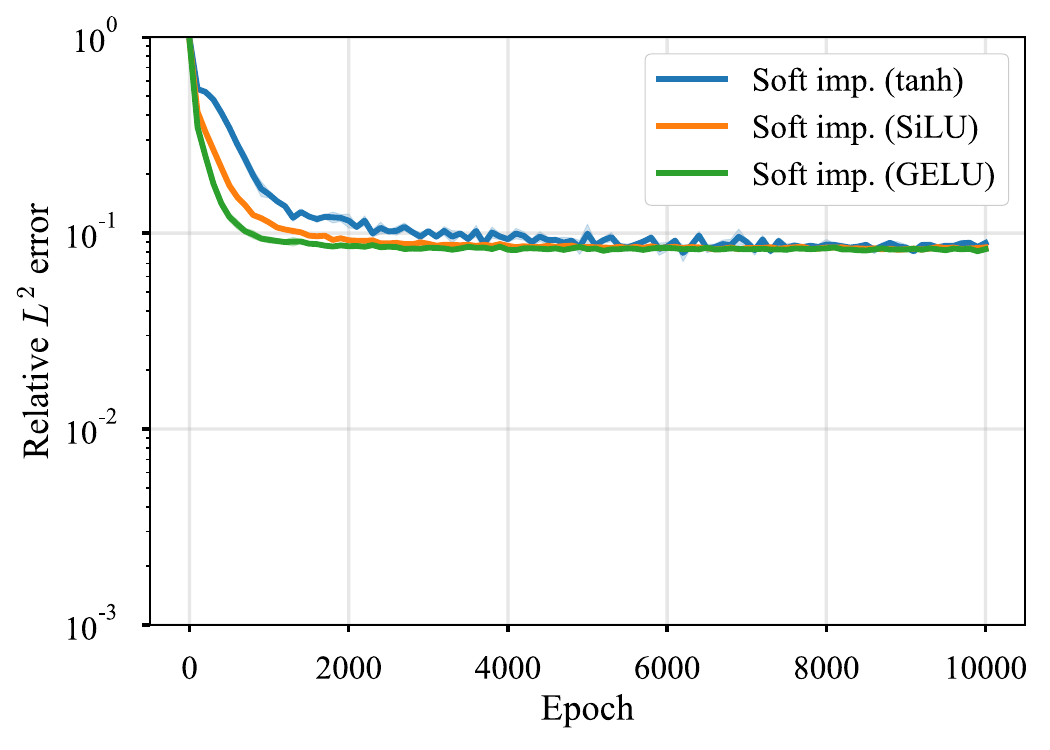}
        \subcaption{
            Soft imposition (weight parameter: $\lambda_{\text{BC}} = 1$)
        }
    \end{minipage}
    \begin{minipage}[b]{.49\linewidth}
        \centering
        \includegraphics[width=.99\linewidth]{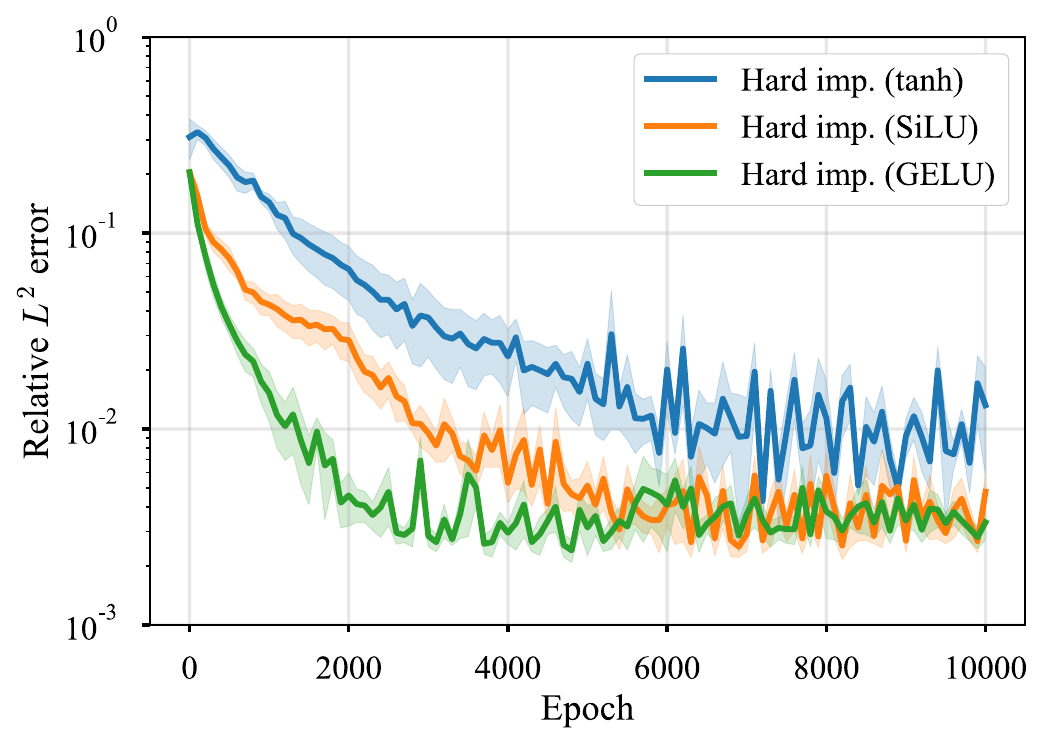}
        \subcaption{
            Hard imposition (normalization order: $m = 1$)
        }
    \end{minipage}
    \caption{
        \emph{Poisson equation with mixed boundary conditions: inhomogeneous Neumann condition case.}
        Relative $L^2$ error of the approximate solution with different activation functions.
        The solid line represents the mean, and the shaded area represents the standard error
        over 5 independent runs with i.i.d. initialization.
        Soft imposition quickly plateaus regardless of the activation function.
    }
    \label{fig:poisson_inhmg_log}
\end{figure}

We further consider the forward problem of the Poisson equation with inhomogeneous Neumann boundary condition.
The domain, source, and Dirichlet boundary conditions remain identical to the previous case,
but the Neumann condition is modified to $g_{N} = 0.1$.
Other settings,
such as the number of collocation points, weights and normalization orders,
are consistent with the homogeneous Neumann condition case.

Figure~\ref{fig:poisson_inhmg_solution} presents
reference,
approximate solution,
absolute error,
and the first derivative of the approximate solution along the negative $y$-direction
to visually realize how well the Neumann condition is satisfied.
Again, the soft imposition fails to meet the prescribed boundary conditions,
whereas the hard imposition effectively handles the inhomogeneous Neumann condition.
One can also observe that
the soft imposition suffers from the same 'plateau' issue as in the homogeneous Neumann condition case,
while the hard imposition is free from this trouble (Figure~\ref{fig:poisson_inhmg_log}).
Table~\ref{tab:poisson_inhmg_summary} summarizes the relative error
with different activation functions and boundary condition impositions.
These results further confirm the effectiveness and applicability of the hard BC imposition.
Based on these results,
GELU demonstrates superior performance among the three activation functions considered.
Therefore, we will employ GELU in the subsequent experiments.


\subsection{Inverse problem: shear-driven cavity flow of an incompressible fluid}   \label{sec:results_cavity}

To evaluate the applicability of hard imposition and adaptive weight tuning in inverse analysis,
we consider the shear-driven cavity flow of an incompressible fluid.
The governing equations are the continuity and momentum equations with no external force:
\begin{alignat}{2}
    \nabla \cdot \bm{u}
    &= 0
    && \quad \text{in} \quad \Omega,
    \label{eq:cavity_continuity}
    \\
    (\bm{u} \cdot \nabla) \bm{u}
    &= - \nabla p + \frac{1}{\mathrm{Re}} \nabla^2 \bm{u}
    && \quad \text{in} \quad \Omega,
    \label{eq:cavity_momentum}
    \\
    \bm{u}
    &= (1, 0)^{\top}
    && \quad \text{on} \quad \Gamma_{3},
    \label{eq:cavity_driving}
    \\
    \bm{u}
    &= (0, 0)^{\top}
    && \quad \text{on} \quad \Gamma_{1} \cup \Gamma_{2} \cup \Gamma_{4},
    \label{eq:cavity_stationary}
\end{alignat}
where $\bm{u} = (u, v)^{\top}$ is the velocity, $p$ is the pressure,
and $\mathrm{Re}$ is the Reynolds number, set to 1,000 and 5,000 in this study.
Boundary conditions~\eqref{eq:cavity_driving} and~\eqref{eq:cavity_stationary} correspond to
the driving and the stationary walls, respectively
(definitions of $\Gamma_{1}$--$\Gamma_{4}$ are consistent with  those in Equations~\eqref{eq:Gamma_1}--\eqref{eq:Gamma_4}).
The reference solution was obtained using the finite difference method on a staggered Arakawa B-type grid~\cite{Arakawa1977Grid}.
Advection was discretized with a third-order upwind scheme~\cite{Kawamura1984,Kawamura1986},
while diffusion and pressure gradient were approximated using second-order central differences.
Velocity and pressure were coupled via Chorin's projection method~\cite{Chorin1968Projection}.
The spatial resolution was set to $\Delta x = \Delta y = 5 \times 10^{-3}$,
which was confirmed to be sufficiently fine for the Reynolds numbers considered.
The time step was selected as $\Delta t = 1 \times 10^{-3}$
in accordance with the stability criterion for explicit time marching~\cite{Courant1967,Neumann1950}.
The steady state was assumed when $\| u^{(n+1)} - u^{(n)} \| / \| u^{(n)} \| \le 1 \times 10^{-8}$ was met.
We have confirmed that the obtained numerical solution
is in good agreement with the previous studies~\cite{Ghia1982Cavity,Erturk2005Cavity}
and is treated as the reference solution for the inverse analysis.
With the obtained reference solutions,
we formulate the inverse problem of identifying the Reynolds number $\mathrm{Re}$ and the pressure $p$
from limited observations of the velocity field $\bm{u}$ (no pressure data is provided to the network).
Collocation points were randomly drawn from the domain with $N_{\text{PDE}} = 4,096$ and $N_{\text{Data}} = 256$,
with an additional $N_{\Gamma} = 256 \times 4$ boundary points for the soft imposition.
The learning rate was initially set to $10^{-3}$,
and an exponential decay schedule was applied with a decay factor of 0.9 every 2,000 iterations.
For both soft and hard boundary condition impositions, bias-corrected dynamic normalization
was employed to adaptively adjust the weights for data loss (Equation~\eqref{eq:loss_data})
and divergence loss (the residual of Equation~\eqref{eq:cavity_continuity}).
The decay rates were chosen based on our previous studies~\cite{Deguchi2023DynNorm,Deguchi2023Keisan},
which indicated that higher decay rates are more effective.
While higher decay rates might typically lead to slower convergence,
the introduced bias correction (Equation~\eqref{eq:weight3}) effectively mitigates this issue,
which is a unique feature of our method to differentiate from similar approaches, such as~\cite{Wang2021GradPath,Maddu2022InvDirichlet,Jin2021NSFnets,Wang2023ExpertGuide}.
For the soft imposition approach, the weight for boundary loss was also adaptively adjusted by dynamic normalization.

\begin{figure}[tpb]
    \centering
    \begin{minipage}[b]{.9\linewidth}
        \centering
        \includegraphics[width=.99\linewidth]{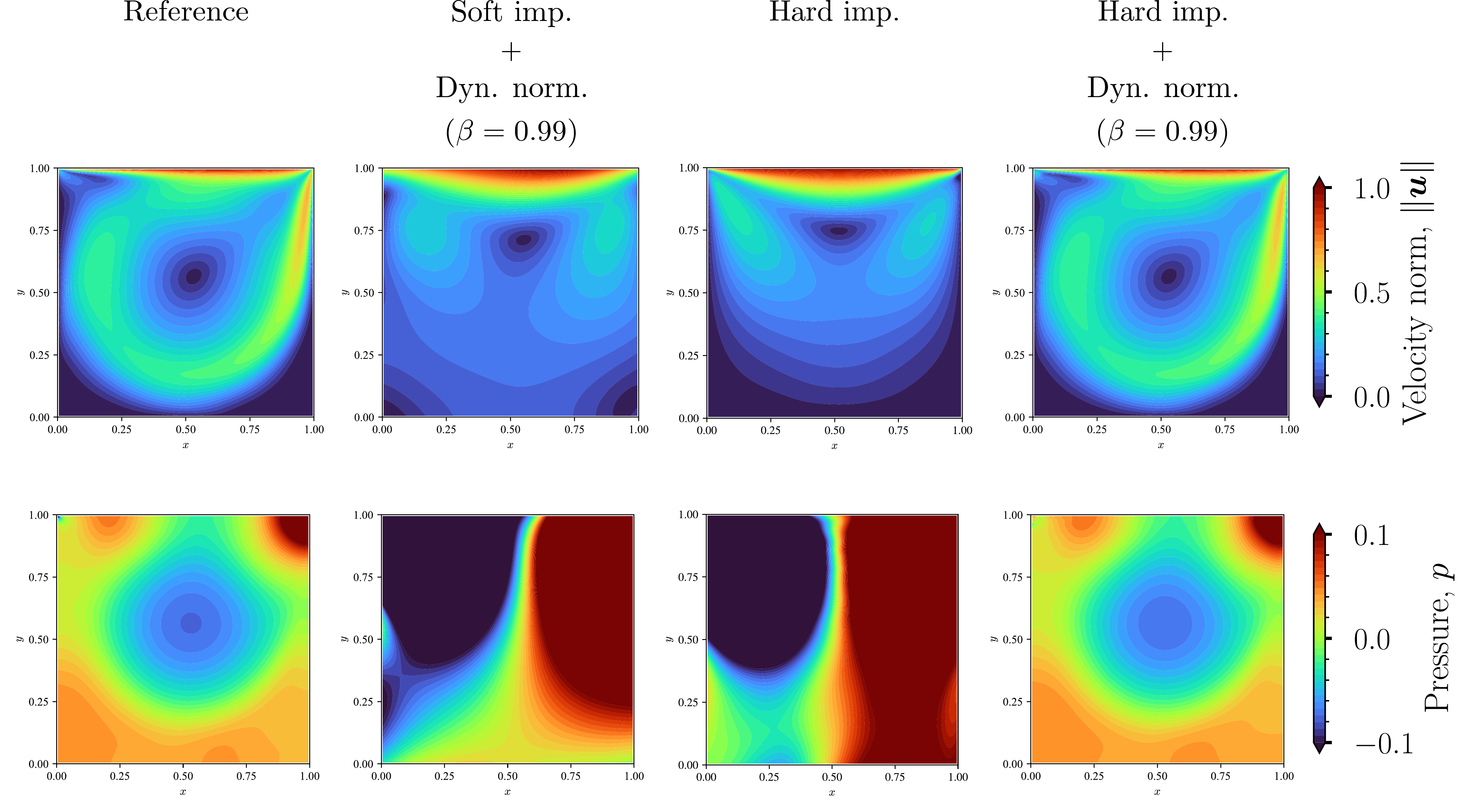}
        \subcaption{Reference solution, reconstructed velocity, and inferred pressure}
    \end{minipage}
    \\
    \vspace{5mm}
    \begin{minipage}[b]{.49\linewidth}
        \centering
        \includegraphics[width=.99\linewidth]{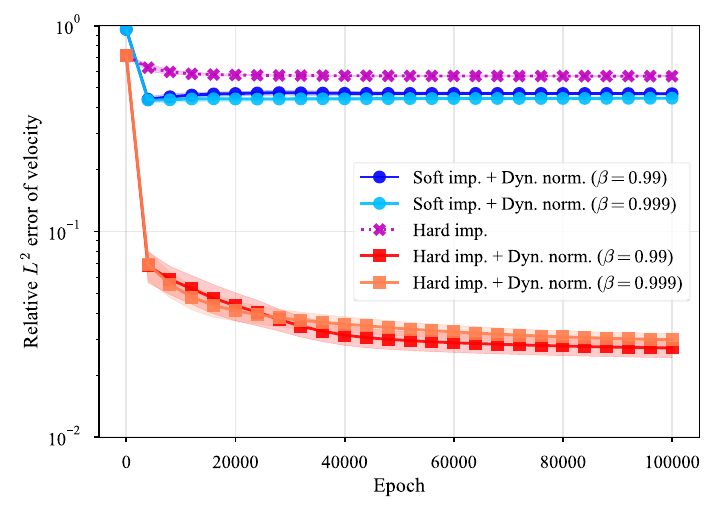}
        \subcaption{Relative $L^2$ error of velocity}
    \end{minipage}
    \begin{minipage}[b]{.49\linewidth}
        \centering
        \includegraphics[width=.99\linewidth]{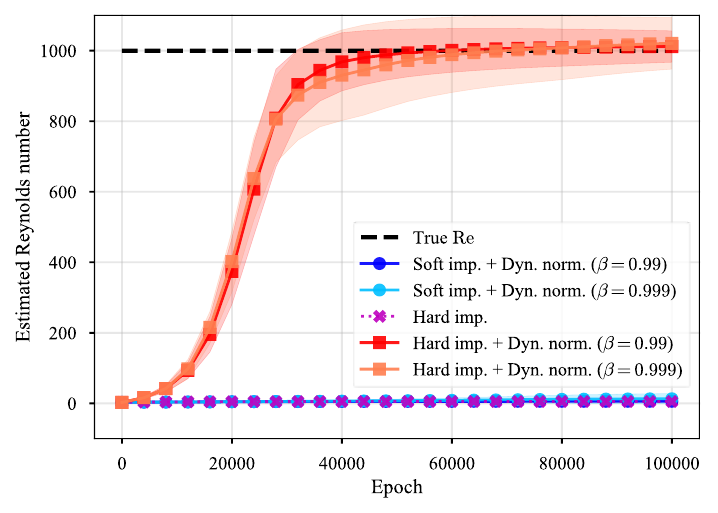}
        \subcaption{Identification of Reynolds number}
    \end{minipage}
    \caption{
        \emph{Shear-driven cavity flow of an incompressible fluid: $\mathit{Re = 1,000}$ case.}
        The application of hard imposition alone remains ineffective.
        Accurate solutions are achieved by combining hard imposition and dynamic normalization.
        In panels (b) and (c), the solid (or dotted) line represents the mean over 5 i.i.d. runs,
        and the shaded area represents the standard error in (b) and the standard deviation in (c).
    }
    \label{fig:cavity_Re1000}
\end{figure}

\begin{figure}[tpb]
    \centering
    \begin{minipage}[b]{.9\linewidth}
        \centering
        \includegraphics[width=.99\linewidth]{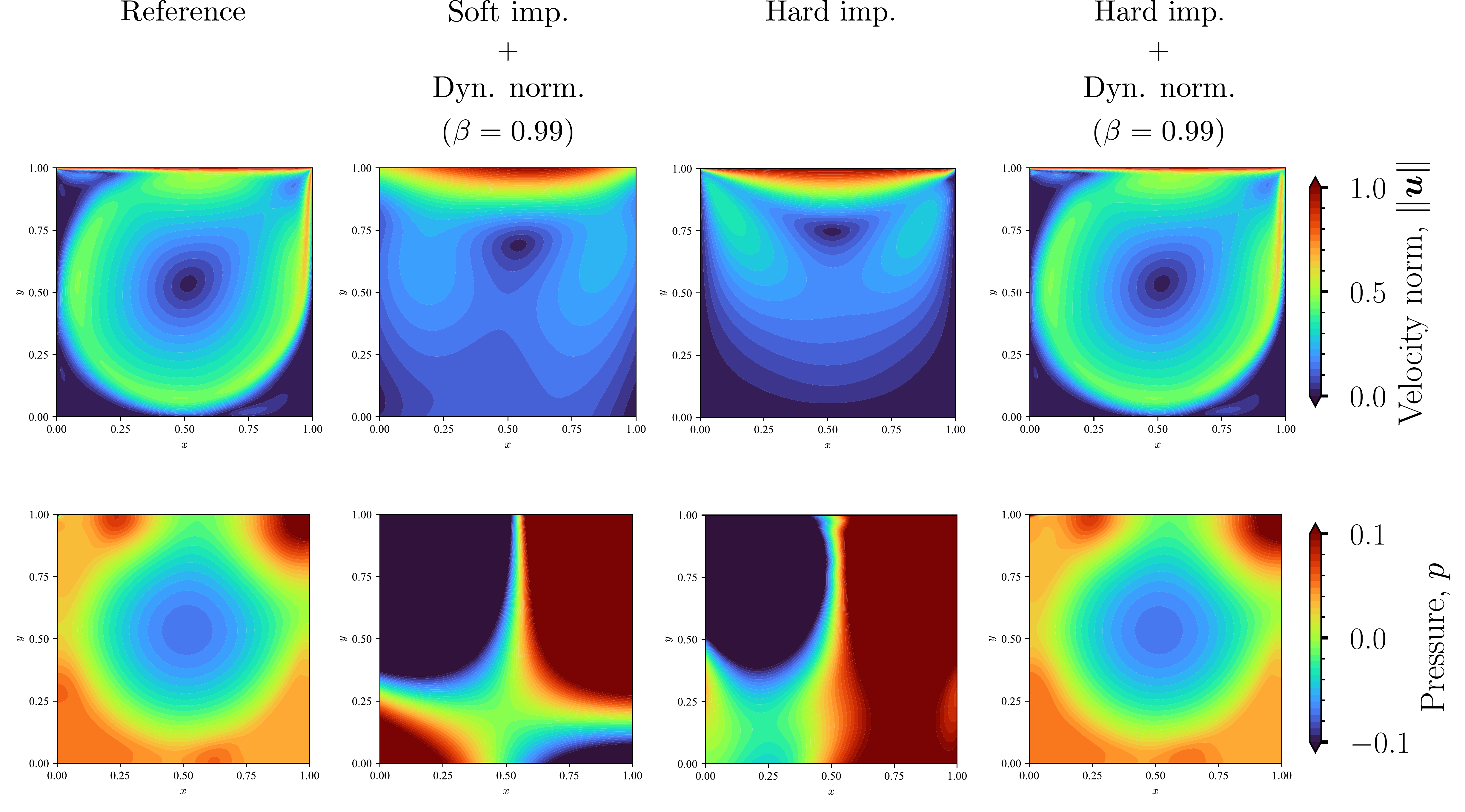}
        \subcaption{Reference solution, reconstructed velocity, and inferred pressure}
    \end{minipage}
    \\
    \vspace{5mm}
    \begin{minipage}[b]{.49\linewidth}
        \centering
        \includegraphics[width=.99\linewidth]{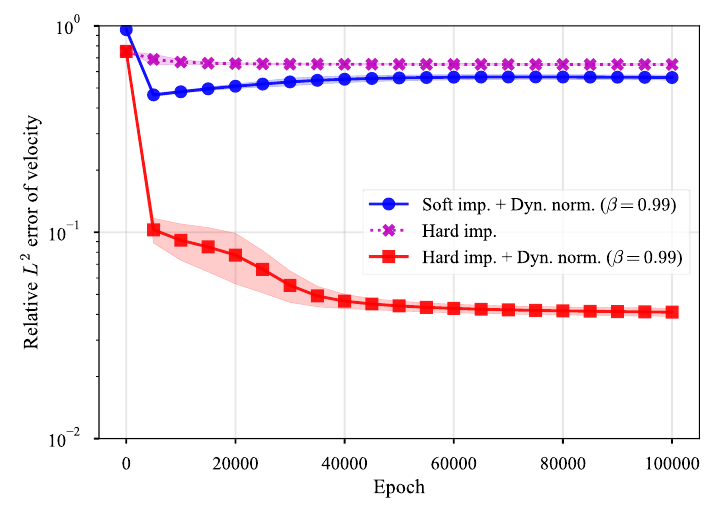}
        \subcaption{Relative $L^2$ error of velocity}
    \end{minipage}
    \begin{minipage}[b]{.49\linewidth}
        \centering
        \includegraphics[width=.99\linewidth]{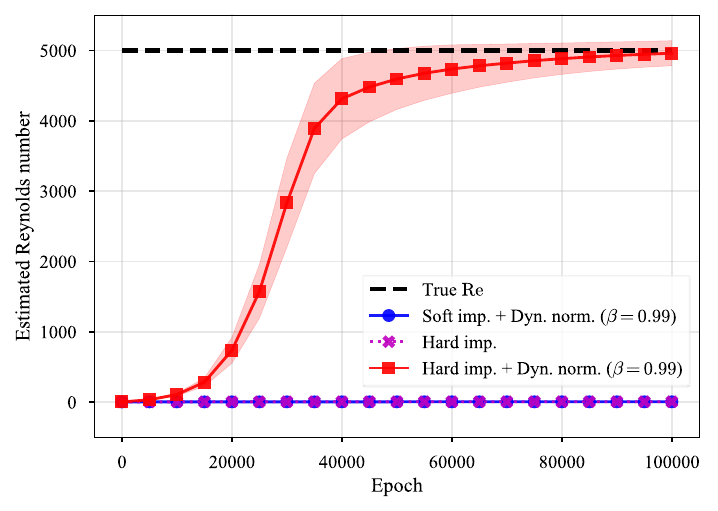}
        \subcaption{Identification of Reynolds number}
    \end{minipage}
    \caption{
        \emph{Shear-driven cavity flow of an incompressible fluid: $\mathit{Re = 5,000}$ case.}
        Similar to the $\mathrm{Re} = 1,000$ case, the sole application of hard imposition fails.
        The combination of hard imposition and dynamic normalization achieves accurate inverse analysis.
        As in Figure \ref{fig:cavity_Re1000},
        panels (b) and (c) depict the mean (solid or dotted line)
        with shaded regions representing the standard error in (b) and the standard deviation in (c) over 5 i.i.d. runs.
    }
    \label{fig:cavity_Re5000}
\end{figure}

Figures~\ref{fig:cavity_Re1000} and~\ref{fig:cavity_Re5000} present
the reconstructed velocity,
inferred pressure,
estimated Reynolds number,
and relative $L^2$ error of velocity field
for the $\mathrm{Re} = 1,000$ and $5,000$ cases, respectively.
The importance of boundary condition treatment is highlighted,
especially by comparing the velocity norm.
Specifically, we observe that the soft imposition fails to recover the velocity field,
particularly due to the violation of the prescribed no-slip boundary condition on $\Gamma_{1} \cup \Gamma_{2} \cup \Gamma_{4}$.
Consequently, the soft imposition approach also fails to estimate the pressure distribution;
the output is inaccurate and physically unrealistic.
Given that PINN performance in inverse analysis is highly dependent on boundary condition treatment~\cite{Sukumar2022MixedBC},
this failure, primarily attributed to improper boundary condition enforcement,
underscores the critical impact of accurate approximate solution construction on inverse analysis.

However, we stress that the sole application of the normalized distance field remains insufficient.
In the third column of Figures~\ref{fig:cavity_Re1000}~(a) and~\ref{fig:cavity_Re5000}~(a),
it can be observed that the hard imposition approach precisely satisfies the boundary conditions (Equations~\eqref{eq:cavity_driving}--\eqref{eq:cavity_stationary});
however, the reconstructed velocity field resembles that of a highly viscous fluid,
which is not consistent with the references at the corresponding Reynolds numbers.
This discrepancy arises from the inability of the hard imposition alone to adequately balance data and PDE losses,
thereby preventing the effective integration of observed data into the network.
Notably, Figures~\ref{fig:cavity_Re1000}~(b) and~\ref{fig:cavity_Re5000}~(b) further indicate
that the plateau issue persists in both the soft and hard approaches within this inverse analysis.

In contrast, the combination of hard imposition and dynamic normalization achieves a balance between multiple losses,
leading to accurate reconstruction of the velocity field and pressure distribution,
as demonstrated in the fourth column of Figures~\ref{fig:cavity_Re1000}~(a) and~\ref{fig:cavity_Re5000}~(a).
The relative error in the velocity field is significantly reduced,
not only when compared to soft imposition but also in comparison to the results obtained with hard imposition alone (Figures~\ref{fig:cavity_Re1000}~(b) and~\ref{fig:cavity_Re5000}~(b)).
The Reynolds numbers are also accurately estimated for both cases,
as depicted in Figures~\ref{fig:cavity_Re1000}~(c) and~\ref{fig:cavity_Re5000}~(c).
Notably, the successful inverse analysis achieved by our proposed framework at a Reynolds number of $5,000$,
without the need for curriculum training,
represents a significant advancement in the application of PINNs to challenging regimes.
It is well-known that directly training PINNs for high Reynolds number flows
often leads to instability and convergence to erroneous solutions,
necessitating the use of staged curriculum learning~\cite{Krishnapriyan2021FailureModes}.
Our results,
demonstrating accurate flow field inference without curriculum training,
underscores the robustness and efficacy
achieved by the integration of hard boundary condition enforcement and bias-corrected adaptive weight tuning.


\subsection{Inverse problem: incompressible flow around an obstacle}   \label{sec:results_von_Karman}

\paragraph{Square obstacle case}

\begin{figure}[tpb]
    \centering
    \begin{minipage}[b]{.9\linewidth}
        \centering
        \includegraphics[width=.99\linewidth]{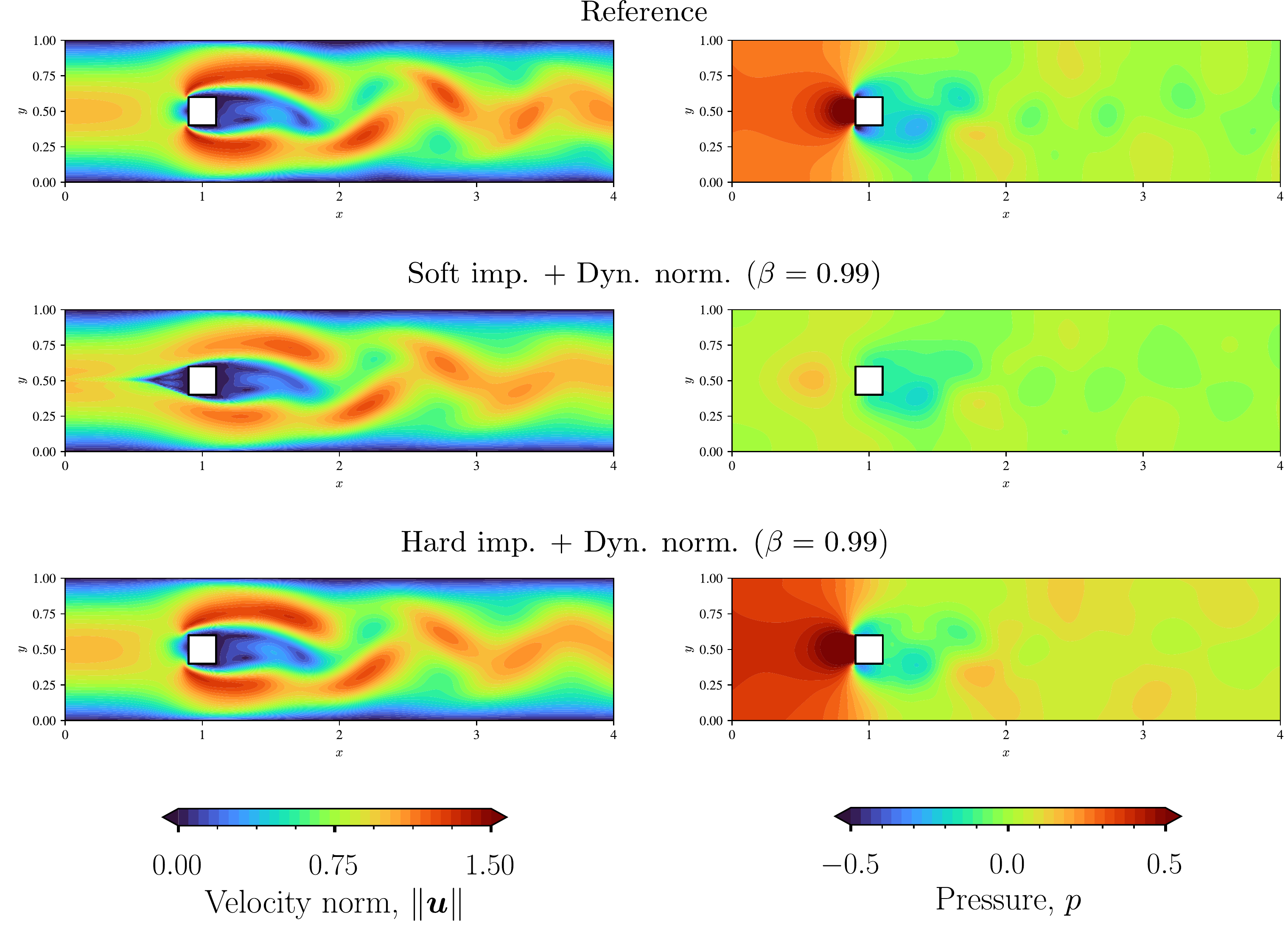}
        \subcaption{
            Velocity norm and pressure.
            (From top to bottom) reference solution, soft and hard impositions with dynamic normalization.
        }
    \end{minipage}
    \\
    \begin{minipage}[b]{.9\linewidth}
        \centering
        \includegraphics[width=.8\linewidth]{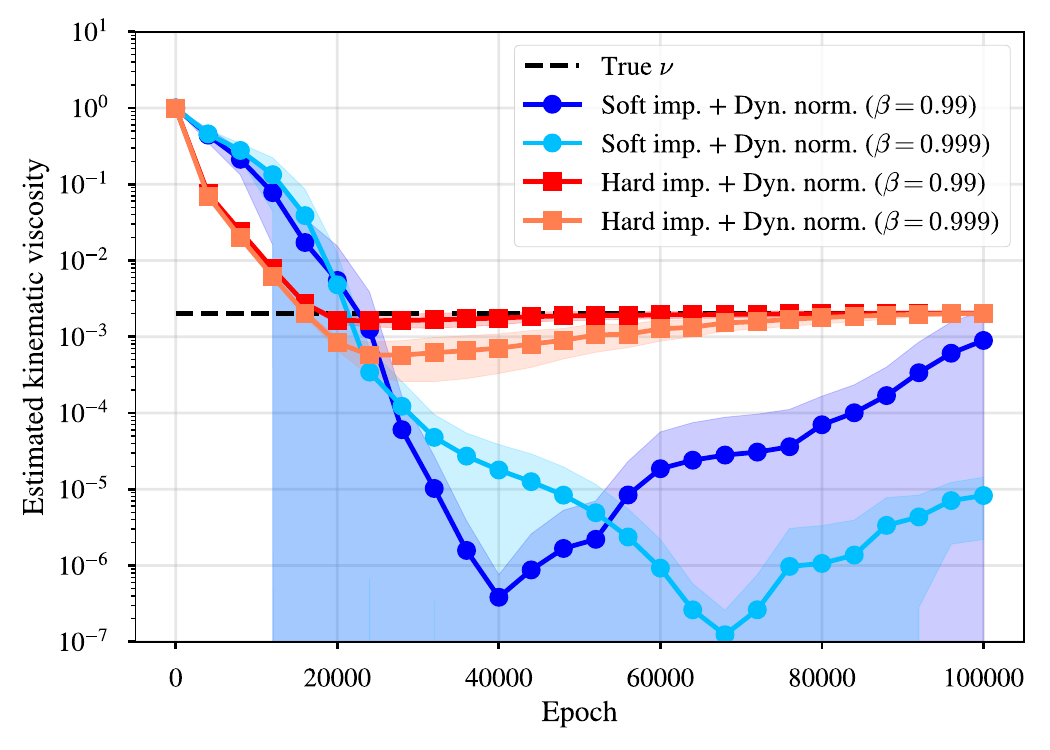}
        \subcaption{
            Estimated kinematic viscosity.
            The solid line and the shaded area represent the mean and the standard deviation over 5 i.i.d. runs, respectively.
            Results without dynamic normalization are not shown for clarity, as they were significantly worse.
        }
    \end{minipage}
    \caption{
        \emph{Incompressible flow around an obstacle: square obstacle case.}
        Reconstructed velocity field, inferred pressure, and estimated kinematic viscosity.
    }
    \label{fig:von_Karman_square}
\end{figure}

\begin{figure}[tpb]
    \centering
    \begin{minipage}[b]{.9\linewidth}
        \centering
        \includegraphics[width=.8\linewidth]{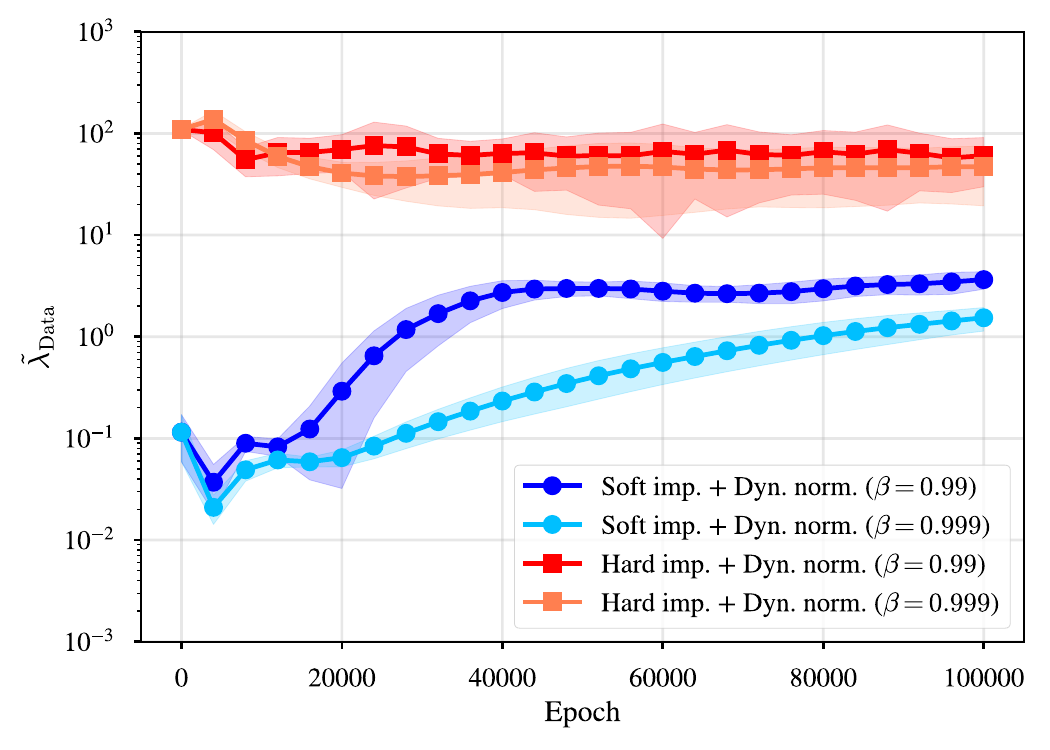}
        \subcaption{Weight assigned to the data loss, $\tilde{\lambda}_{\text{Data}}$}
    \end{minipage}
    \\
    \begin{minipage}[b]{.9\linewidth}
        \centering
        \includegraphics[width=.8\linewidth]{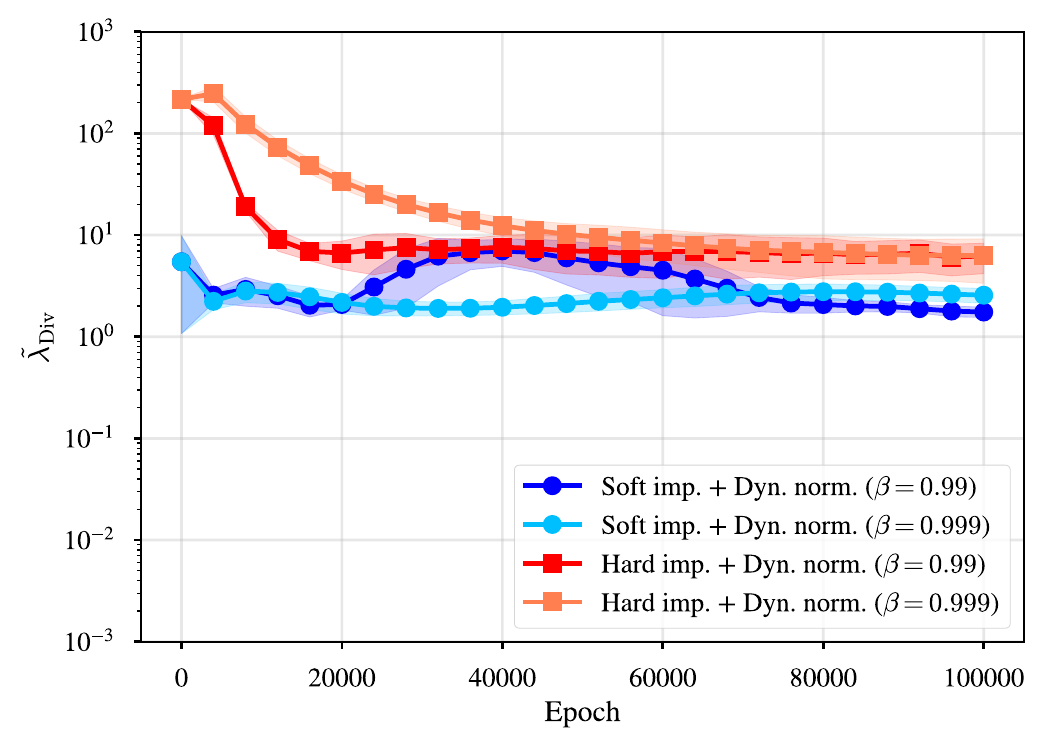}
        \subcaption{Weight assigned to the divergence loss, $\tilde{\lambda}_{\text{Div}}$}
    \end{minipage}
    \caption{
        \emph{Incompressible flow around an obstacle: square obstacle case.}
        Evolution of the adaptive weights assigned to the data and divergence losses.
    }
    \label{fig:von_Karman_square_weight}
\end{figure}

\begin{table}[tpb]
    \centering
    \caption{
        \emph{Incompressible flow around an obstacle: square obstacle case.}
        Estimated kinematic viscosity.
        Mean and standard deviation over 5 i.i.d. runs are shown.
        True value is $2 \times 10^{-3} \ [\mathrm{m^2/s}]$.
    }
    \label{tab:von_Karman_square_summary}
    \begin{tabular}{l|c}
    \toprule
    Method & Estimated kinematic viscosity [$\mathrm{m^2/s}$] \\
    \midrule
    Soft imp.                                              & $7.929 \times 10^{-5} \pm 3.051 \times 10^{-5}$ \\
    Soft imp. + Dyn. norm. ($\beta = 0.99$)                & $8.939 \times 10^{-4} \pm 1.303 \times 10^{-3}$ \\
    Soft imp. + Dyn. norm. ($\beta = 0.999$)               & $8.264 \times 10^{-6} \pm 6.069 \times 10^{-6}$ \\
    Hard imp.                                              & $8.501 \times 10^{-4} \pm 1.384 \times 10^{-3}$ \\
    \textbf{Hard imp. + Dyn. norm. ($\bm{\beta = 0.99}$)}  & $2.038 \times 10^{-3} \pm 1.254 \times 10^{-4}$ \\
    \textbf{Hard imp. + Dyn. norm. ($\bm{\beta = 0.999}$)} & $2.008 \times 10^{-3} \pm 1.166 \times 10^{-4}$ \\
    \bottomrule
    \end{tabular}
\end{table}

Following~\cite{Raissi2019PINN}, we consider an unsteady flow over an obstacle.
The governing equations read as:
\begin{alignat}{2}
    \nabla \cdot \bm{u}
    &= 0
    && \quad \text{in} \quad \Omega \times \mathcal{I},
    \label{eq:von_Karman_continuity}
    \\
    \frac{\partial \bm{u}}{\partial t} + (\bm{u} \cdot \nabla) \bm{u}
    &= - \frac{1}{\rho} \nabla p + \nu \nabla^2 \bm{u}
    && \quad \text{in} \quad \Omega \times \mathcal{I},
    \label{eq:von_Karman_momentum}
\end{alignat}
where $\bm{u} = (u, v)^{\top}$ is the velocity [$\mathrm{m/s}$],
$p$ is the pressure [$\mathrm{Pa}$],
$\rho$ is the density [$\mathrm{kg/m^3}$],
and $\nu$ is the kinematic viscosity [$\mathrm{m^2/s}$].
We set the density and the kinematic viscosity to
$\rho = 1 \ [\mathrm{kg/m^3}]$ and $\nu = 2 \times 10^{-3} \ [\mathrm{m^2/s}]$.
The domain is $4 \ [\mathrm{m}] \times 1 \ [\mathrm{m}]$ rectangle
with a square obstacle of side length $0.2 \ [\mathrm{m}]$ located at $(1, 0.5)$.
No-slip conditions are applied on the obstacle, top, and bottom boundaries.
A parabolic velocity profile with a maximum velocity of $1 \ [\mathrm{m/s}]$ is given on the inlet boundary,
and a zero-gradient condition is imposed on the outlet.
The corresponding Reynolds number is $\mathrm{Re} \sim 100$.
The reference solution was obtained through a FDM simulation,
employing the same computational scheme as in Section~\ref{sec:results_cavity},
with a spatial resolution of $\Delta x = \Delta y = 1 \times 10^{-2} \ [\mathrm{m}]$
and a time step of $\Delta t = 1 \times 10^{-3}  \ [\mathrm{s}]$.
The simulation was executed until the flow reached a periodic state,
and a 5-second segment of this periodic flow was extracted as the reference solution.
Similar to the cavity flow problem,
we consider an inverse problem of identifying the kinematic viscosity $\nu$
and the pressure $p$ from limited observations of the velocity field $\bm{u}$.
This study differs from the previous works~\cite{Raissi2019PINN,Jin2021NSFnets}
by considering the entire computational domain, rather than focusing solely on the wake region.
This highlights the effect of different boundary condition imposition techniques
on the performance of the inverse analysis using PINN.
Here we employ normalized distance functions (approach (3-iii) based on R-functions~\cite{Sheiko1982,Rvachev2001,Shapiro2007}),
which makes it possible to detect both interior and outer boundaries,
unlike na\"ive approaches (3-i) used in ~\cite{Lagaris1998,Lu2021HardDesign,Li2024HardAdvcDiff}
that only handle convex boundaries.
Collocation points were randomly drawn with $N_{\text{PDE}} = 1,024$ and $N_{\text{Data}} = 64$ every $0.1 \ [\mathrm{s}]$,
resulting in a total of $N_{\Omega \times \mathcal{I}} = 52,224$ and $N_{\text{Data} \times \mathcal{I}} = 3,264$.
For soft imposition,
$N_{\Gamma_{\text{inlet}} \times \mathcal{I}} = N_{\Gamma_{\text{outlet}} \times \mathcal{I}} = 5,151$,
$N_{\Gamma_{\text{top}} \times \mathcal{I}} = N_{\Gamma_{\text{bottom}} \times \mathcal{I}} = 20,451$,
$N_{\Gamma_{\text{cylinder}} \times \mathcal{I}} = 4,284$ points were added for the boundary condition enforcement.
For this problem, we only employed GELU, as it demonstrated the best performance in previous sections.
The Adam optimizer with a learning rate of $1 \times 10^{-3}$ was used.

Figure~\ref{fig:von_Karman_square}~(a) shows
the reference, reconstructed velocity, and inferred pressure fields,
as well as the estimated kinematic viscosity for the square obstacle case.
While the soft imposition demonstrates reasonable agreement with the reference in the wake region,
it faces significant challenges in accurately capturing the flow pattern near the obstacle.
In particular, the velocity field exhibits non-physical flow separation upstream of the object,
with its magnitude being underestimated both near the obstacle and within the wake region.
Furthermore, the pressure field shows a non-physical distribution,
with the peak point shifted upstream from the obstacle.
Conversely, the combination of hard imposition and dynamic normalization captures the flow structure accurately,
including the region around the cylinder.
Although a slight underestimation of the velocity magnitude is observed in the wake,
the discrepancy is considerably smaller than that of the soft imposition.
The pressure field, which was not provided to the network, is also inferred with high accuracy,
with the peak pressure point correctly located on the obstacle surface.
It is important to note that, due to the nature of the incompressible flow,
the pressure is determined only up to an additive constant~\cite{ChorinMarsden1993}.
Therefore, the slight shift observed in the inferred pressure
under hard imposition and dynamic normalization is not a concern.

The results of the kinematic viscosity estimation
are shown in Figure~\ref{fig:von_Karman_square}~(b)
and summarized in Table~\ref{tab:von_Karman_square_summary}.
Figure~\ref{fig:von_Karman_square}~(b) reveals that
the soft imposition yields highly unstable viscosity estimates,
rendering it unsuitable for practical applications.
As can be seen in Table~\ref{tab:von_Karman_square_summary},
the soft imposition, even with dynamic normalization,
fails to accurately identify the kinematic viscosity,
undershooting the true value by orders of magnitude.
In contrast, the integration of hard imposition and dynamic normalization
provides stable and accurate viscosity estimates, closely matching the true value.
Notably, the estimation error in the viscosity is only 1.9\% for $\beta = 0.99$ and 0.4\% for $\beta = 0.999$,
where $\beta$ is the decay rate of dynamic normalization.
Furthermore, training did not converge in the absence of dynamic normalization,
as indicated in Table~\ref{tab:von_Karman_square_summary}.
These findings underscore the critical role of hard boundary condition imposition
and dynamic normalization in enhancing the performance of PINNs in inverse analysis.
In addition, Figure~\ref{fig:von_Karman_square_weight} illustrates
the evolution of the adaptive weights assigned to the data and divergence losses.
One can observe that the weights are adaptively adjusted from the early stages of training,
highlighting the effectiveness of bias-corrected dynamic normalization in balancing the data and divergence losses.


\paragraph{Heart-shaped obstacle case}
\begin{figure}[tpb]
    \centering
    \begin{minipage}[b]{.99\linewidth}
        \centering
        \includegraphics[width=.99\linewidth]{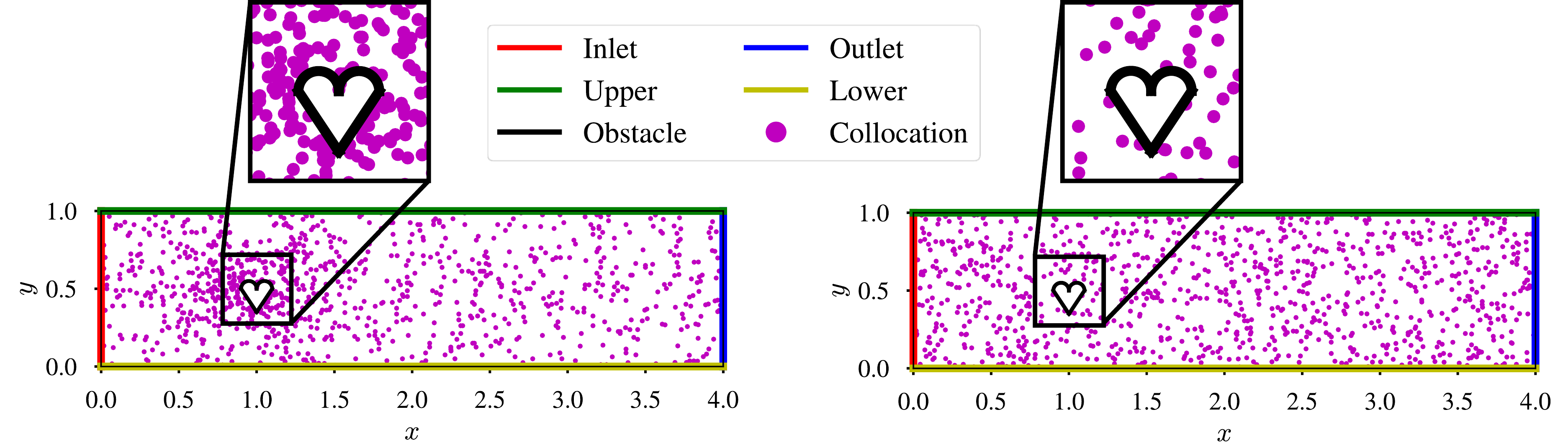}
        \subcaption{Distribution of collocation points}
    \end{minipage}
    \\
    \vspace{5mm}
    \begin{minipage}[b]{.99\linewidth}
        \centering
        \includegraphics[width=.99\linewidth]{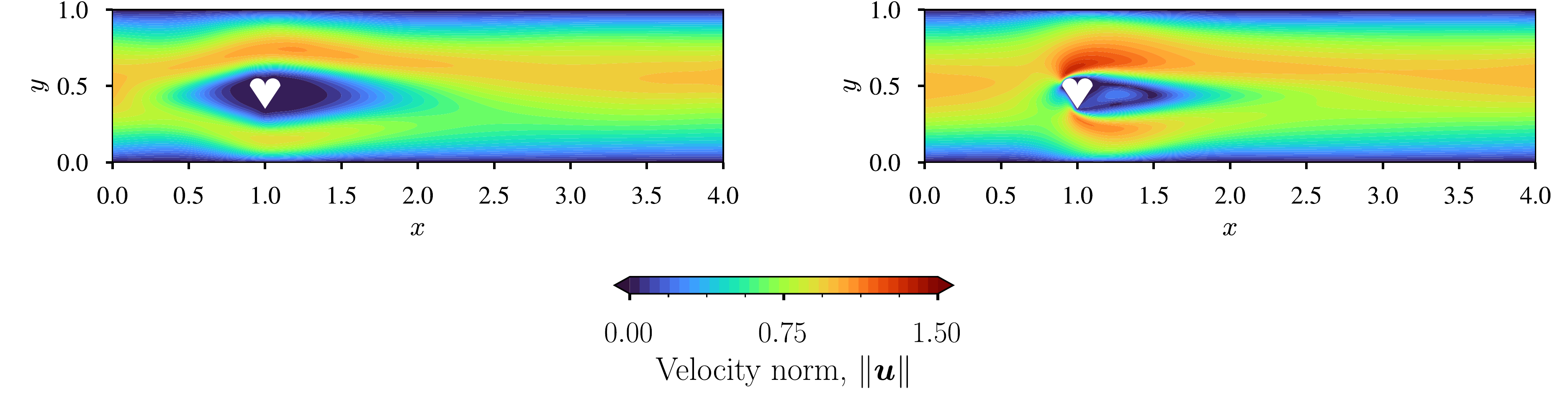}
        \subcaption{Reconstructed velocity field}
    \end{minipage}
    \caption{
        \emph{Incompressible flow around an obstacle: heart-shaped obstacle case.}
        Comparison of collocation point sampling strategy.
        (Left) non-uniform distribution, where FEM nodes are used as collocation points.
        (Right) uniform distribution, where collocation points are drawn from the uniform probability density.
    }
    \label{fig:von_Karman_heart_collocation_schematic}
\end{figure}

\begin{figure}[tpb]
    \centering
    \begin{minipage}[b]{.9\linewidth}
        \centering
        \includegraphics[width=.99\linewidth]{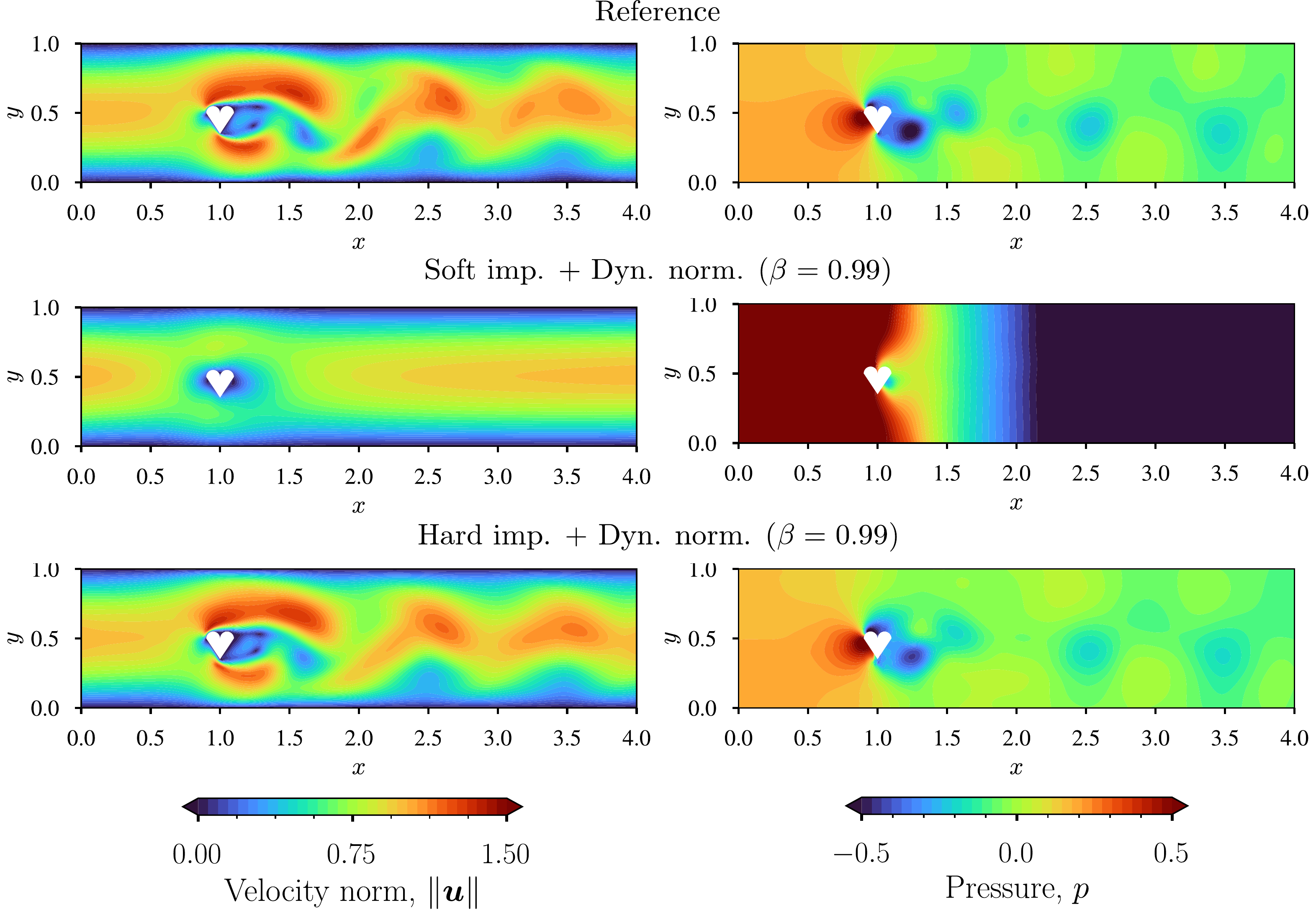}
        \subcaption{
            Velocity norm and pressure.
            (From top to bottom) reference solution, soft and hard impositions with dynamic normalization.
        }
    \end{minipage}
    \\
    \begin{minipage}[b]{.9\linewidth}
        \centering
        \includegraphics[width=.8\linewidth]{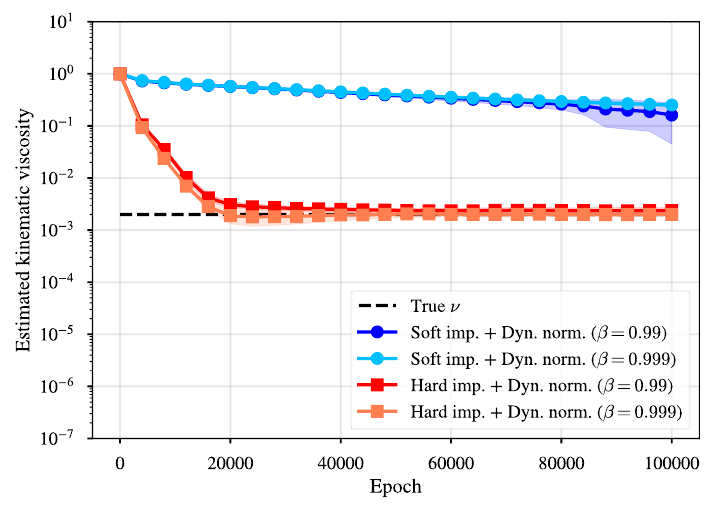}
        \subcaption{
            Estimated kinematic viscosity.
            The solid line and the shaded area represent the mean and the standard deviation over 5 i.i.d. runs, respectively.
            Results without dynamic normalization are not shown for clarity, as they did not provide satisfactory results.
        }
    \end{minipage}
    \caption{
        \emph{Incompressible flow around an obstacle: heart-shaped obstacle case.}
        Reconstructed velocity field, inferred pressure, and estimated kinematic viscosity.
    }
    \label{fig:von_Karman_heart}
\end{figure}

\begin{figure}[tpb]
    \centering
    \begin{minipage}[b]{.9\linewidth}
        \centering
        \includegraphics[width=.8\linewidth]{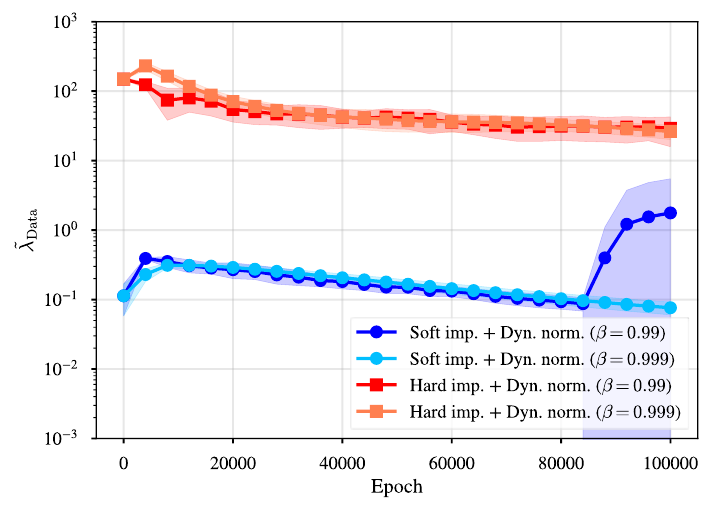}
        \subcaption{Weight assigned to the data loss, $\tilde{\lambda}_{\text{Data}}$}
    \end{minipage}
    \\
    \begin{minipage}[b]{.9\linewidth}
        \centering
        \includegraphics[width=.8\linewidth]{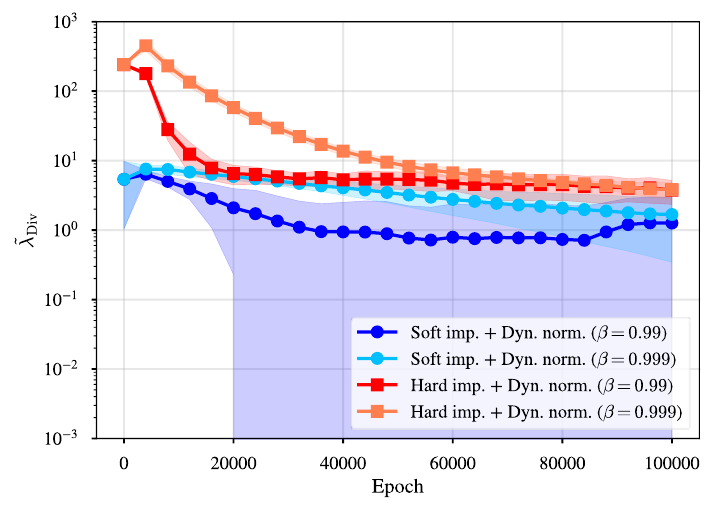}
        \subcaption{Weight assigned to the divergence loss, $\tilde{\lambda}_{\text{Div}}$}
    \end{minipage}
    \caption{
        \emph{Incompressible flow around an obstacle: heart-shaped obstacle case.}
        Evolution of the adaptive weights assigned to the data and divergence losses.
    }
    \label{fig:von_Karman_heart_weight}
\end{figure}

\begin{table}[tpb]
    \centering
    \caption{
        \emph{Incompressible flow around an obstacle: heart-shaped obstacle case.}
        Estimated kinematic viscosity.
        Mean and standard deviation over 5 i.i.d. runs are shown.
        True value is $2 \times 10^{-3} \ [\mathrm{m^2/s}]$.
    }
    \label{tab:von_Karman_heart_summary}
    \begin{tabular}{l|c}
    \toprule
    Method & Estimated kinematic viscosity [$\mathrm{m^2/s}$] \\
    \midrule
    Soft imp                                               & $4.609 \times 10^{-3} \pm 1.038 \times 10^{-4}$ \\
    Soft imp. + Dyn. norm. ($\beta = 0.99$)                & $1.611 \times 10^{-1} \pm 1.163 \times 10^{-1}$ \\
    Soft imp. + Dyn. norm. ($\beta = 0.999$)               & $2.529 \times 10^{-1} \pm 2.223 \times 10^{-2}$ \\
    Hard imp.                                              & $2.532 \times 10^{-3} \pm 9.627 \times 10^{-4}$ \\
    \textbf{Hard imp. + Dyn. norm. ($\bm{\beta = 0.99}$)}  & $2.369 \times 10^{-3} \pm 3.335 \times 10^{-4}$ \\
    \textbf{Hard imp. + Dyn. norm. ($\bm{\beta = 0.999}$)} & $1.999 \times 10^{-3} \pm 1.066 \times 10^{-4}$ \\
    \bottomrule
    \end{tabular}
\end{table}

The methodology is further extended to the flow around a heart-shaped obstacle.
The governing equations are the same as in the previous case
(Equations~\eqref{eq:von_Karman_continuity} and~\eqref{eq:von_Karman_momentum}).
The domain is a $4 \ [\mathrm{m}] \times 1 \ [\mathrm{m}]$ rectangle,
and the square cylinder is replaced with a heart-shaped obstacle.
The same kinematic viscosity and boundary conditions are applied,
and the geometric parameters are chosen so that the same Reynolds number is maintained.
The reference solution was obtained using a P2/P1 finite element method~\cite{Hecht2012FreeFEM}
on a triangular mesh to satisfy the inf-sup condition,
ensuring the stability of the numerical solutions~\cite{Brenner2008}.
The mesh was progressively refined towards the obstacle,
resulting in the characteristic (maximum) element lengths of $2.4 \times 10^{-2} \ [\mathrm{m}]$.
The problem was solved with varying mesh resolutions,
confirming the convergence of the numerical results with the current resolution.
The flow exhibits periodic, yet asymmetric behavior, attributed to the obstacle's geometry.
The obtained numerical solution is used as the reference for the inverse analysis.
The same number of collocation points and identical settings for the learning rate,
dynamic normalization, and activation function were used as in the square obstacle case.

In the context of computational fluid dynamics,
it is common practice to employ dense and refined meshes near obstacles with complex geometries
so that one can accurately capture steep gradients of the flow.
Motivated by this, we first compared the PINN's performance using collocation points
taken from the finite element nodes (increased density near the obstacle)
and those uniformly drawn from the domain (see Figure~\ref{fig:von_Karman_heart_collocation_schematic}~(a)).
For this preliminary test,
we used hard imposition approach without dynamic normalization
to focus on the effect of the collocation point distribution.
Velocity reconstruction results are presented in Figure~\ref{fig:von_Karman_heart_collocation_schematic}~(b).
Although both are not quite satisfactory,
the uniform distribution of collocation points yields better results.
This result is fairly natural from Monte Carlo perspective,
as each collocation point holds a representative volume to discretize the integrals in the loss function.
Indeed, low-discrepancy sequences (e.g., lattice) are known to generally facilitate quicker convergence~\cite{Matsubara2023GLT}.
For non-uniform distributions,
techniques such as volumetric averaging~\cite{Song2024VWPINN} can be incorporated
to adjust the representative volume based on collocation point density.
However, a comprehensive study of non-uniform collocation points is beyond the scope of this work.
Based on this observation,
we proceed with uniformly distributed collocation points for the problem with the heart-shaped obstacle.

Figure~\ref{fig:von_Karman_heart} shows the velocity field, pressure, and estimated kinematic viscosity
obtained using the soft and hard impositions with dynamic normalization.
From Figure~\ref{fig:von_Karman_heart}~(a),
one can observe that the solution obtained with soft imposition,
even with dynamic normalization,
does not exhibit the shedding structure,
resembling a highly viscous or steady flow.
This observation aligns with~\cite{Chuang2022ExperienceReport},
and is attributed to improper boundary condition treatment.
Results with the hard imposition clearly reproduce the vortex shedding,
and the unobserved pressure is also accurately inferred.
The kinematic viscosity estimation is illustrated in Figure~\ref{fig:von_Karman_heart}~(b)
and summarized in Table~\ref{tab:von_Karman_heart_summary}.
One can further confirm that the unobserved viscosity is accurately identified
by the integration of hard imposition and dynamic normalization.
Notably, neither distance function-based enforcement nor dynamic normalization alone is sufficient;
both techniques are necessary to achieve accurate results
(see Table~\ref{tab:von_Karman_heart_summary} and Figure~\ref{fig:von_Karman_heart_collocation_schematic}~(b) (right)).
Once again, the distance field constructed upon R-functions~\cite{Sheiko1982,Rvachev2001,Shapiro2007}
enables the detection of both outer and interior boundaries,
and further extends to non-convex geometries, such as the heart-shaped domain in this problem.
This capability is a unique feature of R-function-based distance fields,
whereas approaches in~\cite{Lagaris1998,Sun2020Surrogate,Lu2021HardDesign,Tang2023PolyPINN} cannot handle,
as they implicitly assume convex hull boundaries.
Furthermore, Figure~\ref{fig:von_Karman_heart_weight} illustrates the evolution of the adaptive weights built by bias-corrected dynamic normalization.
The weights exhibit stable behavior when combined with the distance functions,
which is similar to the behavior of adaptive activation functions observed in the channel flow problems by Sun et al.~\cite{Sun2020Surrogate}.
These results suggest that hard boundary condition enforcement contributes to the stability of the adaptive weights.


\section{Conclusion}   \label{sec:conclusion}
In this study,
we investigated the impact of boundary condition imposition techniques
on the performance of physics-informed neural networks
for both forward and inverse problems of partial differential equations.
Although penalty-based soft enforcement is prevalent in the literature,
its performance is highly sensitive to the weight parameter,
requiring very careful tuning~\cite{E2018DRM,Chen2020DGMvsDRM}.
In contrast, distance function-based hard enforcement
directly encodes the boundary conditions into the approximate function structure,
ensuring that the solution inherently satisfies the prescribed conditions.
In addition, we have reviewed the properties of exact and approximate distance functions~\cite{Biswas2004}
while exploring the use of distance functions in the context of boundary value problems.
In the framework of R-functions, the distance field can be normalized,
providing a smooth and differentiable representation of distance and unit normal vectors on the boundaries.
These properties are crucial for applications
to homogeneous/inhomogeneous Dirichlet, Neumann, and mixed boundary conditions~\cite{Shapiro2002SAGE,Tsukanov2003MeshfreeFluid,Millan2015,Sukumar2022MixedBC,Berrone2023BCEnforcement}
and enable the detection of arbitrary polygonal and non-convex boundaries,
overcoming limitations of other (na\"ive and non-normalized) distance function-based approaches such as those presented in~\cite{Lagaris1998,Lu2021HardDesign,Li2024HardAdvcDiff}.

The effectiveness of normalized distance-based hard imposition was first verified with a forward problem of the Poisson equation
and subsequently extended to inverse problems of incompressible flow in a cavity and around an obstacle.
For forward problems, soft imposition approach quickly reached a plateau,
which is consistent result with~\cite{Sun2020Surrogate}.
In contrast, hard imposition outperformed soft enforcement across various activation functions,
achieving faster convergence and lower error.
The same architecture was then applied to inverse problems,
however, we found that the sole application of hard imposition to explicitly enforce the boundary conditions
was insufficient to achieve accurate inverse analysis.
To address this, we introduced dynamic normalization~\cite{Deguchi2023DynNorm},
which adaptively adjusts the weights assigned to the data and divergence losses,
and combined it with hard imposition.
The proposed framework,
incorporating hard imposition (R-function-based normalized distance)
and dynamic normalization (bias-corrected adaptive weight),
provided a robust and efficient inverse analysis approach.
This study underscores the importance of boundary condition treatment
and the efficacy of adaptive weight tuning
in improving the performance of PINNs, particularly for inverse problems.

In this study,
we have demonstrated the applicability of the proposed method to steady and unsteady problems.
The methodology can be further extended to initial-boundary value problems
by interpreting the initial condition as a special case of Dirichlet condition
on the spatio-temporal plane~\cite{Sun2020Surrogate,Berrone2023BCEnforcement,Kharazmi2021hpVPINN,Matsubara2023GLT}
and extending the R-function-based distance fields to spatio-temporal domains
(e.g.,~\cite{Sanchez2015SpaceTimeRfunction,Mulroy2022RfunctionSoftRobots}).
Notably, for time-dependent problems where the solution exhibits dynamic, multi-scale, or turbulent behavior,
one could consider loss function formulations that explicitly account for temporal causality~\cite{Wang2024Causality}
to further enhance the network to learn the temporal evolution.

The presented method can be further improved by exploring various aspects.
For instance, the neural network structure employed was a simple MLP.
Future investigations could explore the integration with advanced network architectures and preprocessing techniques,
such as basis transformation~\cite{Wang2021FourierFeature,Tang2023PolyPINN,Wang2024PirateNets},
domain decomposition~\cite{Jagtap2020XPINN,Moseley2023FBPINN,Li2023D3M},
and mixed-variable formulations~\cite{Rao2020TAML,Haghighat2021SolidMechanics},
to more effectively capture complex flow structures.
Furthermore, future work could explore the integration of different architectures,
such as convolutional neural networks to efficiently capture spatial features~\cite{Gao2021PhyGeoNet}
or attention mechanisms~\cite{Vaswani2017Transformer} to automatically detect important regions of the domain~\cite{McClenny2023SA}.
The applicability of the current R-function-based hard imposition primarily focused on fixed boundaries.
Extending this framework to handle domain changes, such as moving boundaries,
presents another avenue for future research,
which could potentially be addressed by constructing time-varying distance functions~\cite{Sanchez2015SpaceTimeRfunction}.


\subsubsection*{Acknowledgements}
This work was supported by
Japan Society for the Promotion of Science (JSPS) KAKENHI Grant Number JP25KJ0026, JP23KK0182, JP23K17807, JP23K26356, JP23K24857, JP23KJ1685,
and Japan Science and Technology Agency (JST) Support for Pioneering Research Initiated by the Next Generation (SPRING) Grant Number JPMJSP2136.

\subsubsection*{CRediT author contribution statement}
\textbf{S. Deguchi:}
Conceptualization,
Methodology,
Validation,
Formal analysis,
Investigation,
Data Curation,
Writing - Original Draft,
Funding acquisition.

\noindent
\textbf{M. Asai:}
Conceptualization,
Methodology,
Resources,
Writing - Review \& Editing,
Supervision,
Project administration,
Funding acquisition.

\subsubsection*{Data availability}
The code will be made available at:
\url{https://github.com/ShotaDeguchi/PINN_HardBC_DynNorm}
upon publication.

\subsubsection*{Conflict of interest}
The authors have no competing interests to declare that are relevant to the content of this article.


\begin{appendices}

\section{Numerical study on several improvement techniques}   \label{sec:appendix_improvement}
\renewcommand{\theequation}{A.\arabic{equation}}
\renewcommand{\thefigure}{A.\arabic{figure}}
\renewcommand{\thetable}{A.\arabic{table}}
\setcounter{equation}{0}
\setcounter{figure}{0}
\setcounter{table}{0}

Several techniques have been proposed to improve the performance of PINN.
In this section, we report numerical results for some of these techniques
applied to the Poisson equation with mixed boundary conditions
(described in Section~\ref{sec:results_poisson}).

\paragraph{Adaptive activation function}
There have been many studies on adaptive activation functions in the literature of neural networks~\cite{Chen1996AGtanh,Ramachandran2017Swish}.
Similarly, Jagtap et al.~\cite{Jagtap2020GAAF,Jagtap2020LAAF} proposed adaptive activation functions in the context of PINNs,
in which the forward pass is modified as follows:
\begin{equation}
    \mathbf{z}^{(l)}
    = \sigma^{(l)} \left(
        s^{(l)} \left(
            \mathbf{W}^{(l)} \mathbf{z}^{(l-1)} + \mathbf{b}^{(l)}
        \right)
    \right),
    \label{eq:adaptive_activation}
\end{equation}
where a learnable scaling factor $s^{(l)}$ is introduced in each layer.
Equation~\eqref{eq:adaptive_activation} is known as layer-wise locally adaptive activation function (L-LAAF)~\cite{Jagtap2020LAAF}.
With this modification, the set of trainable parameters becomes
$\bm{\theta} = \left\{ \mathbf{W}^{(l)}, \bm{b}^{(l)}, s^{(l)} \right\}_{l=1}^L \setminus \left\{ s^{(L)} \right\}$.
The core idea behind this modification is to adaptively adjust the slope of the activation functions.
Specifically, backpropagated gradients are amplified if $s^{(l)} \ge 1$ and attenuated if $s^{(l)} < 1$.
To encourage $s^{(l)}$ to increase, thereby accelerating optimization,
Jagtap et al.~\cite{Jagtap2020LAAF} introduced the following slope recovery term into the loss function:
\begin{equation}
    \mathcal{L}_{\text{SR}}
    = \frac{1}{1 / (L-1) \sum_{l=1}^{L-1} \exp \left( s^{(l)} \right)}.
\end{equation}
As $\mathcal{L}_{\text{SR}}$ decreases, $s^{(l)}$ increase,
which enhances the gradients and accelerates convergence.

\paragraph{Gradient enhancement}
Yu et al.~\cite{Yu2022GradEnhc} proposed a gradient enhancement technique, termed gPINN,
where an additional loss term is introduced to penalize the derivatives of the PDE residual so that they approach zero.
In particular, they introduced the following gradient enhancement term:
\begin{equation}
    \mathcal{L}_{\text{GE}}
    = \sum_{k=1}^{d} \int_{\Omega}
        \left|
            \frac{\partial}{\partial x_k}
            \left( - \nabla^2 \hat{u} \left( \bm{x}; \bm{\theta} \right) - f \left( \bm{x} \right) \right)
        \right|^2
    \, d\bm{x}.
\end{equation}
While higher-order derivatives can be included to further enforce smoothness,
the computational cost increases rapidly with the order of differentiation.
Moreover, the effectiveness of gradient enhancement is highly sensitive to the weight $\lambda_{\text{GE}}$,
and improper tuning of $\lambda_{\text{GE}}$ may lead to degraded performance,
whereas gPINN could be effective for scarce data and help prevent oscillation due to its smoothing effect~\cite{Mohammadian2023gPINN,Eshkofti2023gPINN}.

\paragraph{Self-adaptive soft attention mechanism}
McClenny \& Braga-Neto~\cite{McClenny2023SA} proposed self-adaptive PINN (SA-PINN),
in which a set of learnable weights is introduced to adaptively determine the importance of each collocation point.
In SA-PINN, the loss function is modified as follows:
\begin{align}
    \mathcal{L} \left( \bm{\theta}, \bm{\lambda} \right)
    &= \mathcal{L}_{\text{PDE}} \left( \bm{\theta}, \bm{\lambda}_{\text{PDE}} \right)
        + \mathcal{L}_{\text{BC}} \left( \bm{\theta}, \bm{\lambda}_{\text{DBC}}, \bm{\lambda}_{\text{NBC}} \right)
        + \mathcal{L}_{\text{Data}} \left( \bm{\theta}, \bm{\lambda}_{\text{Data}} \right),
    \label{eq:self_adaptive_loss}
    \\
    \mathcal{L}_{\text{PDE}} \left( \bm{\theta}, \bm{\lambda}_{\text{PDE}} \right)
    &= \frac{1}{N_{\text{PDE}}} \sum_{i=1}^{N_{\text{PDE}}}
        m \left( \lambda_{\text{PDE}, i} \right)
        \left|
            -\nabla^2 \hat{u} \left( \bm{x}_i; \bm{\theta} \right) - f \left( \bm{x}_i \right)
        \right|^2,
    \\
    \begin{split}
        \mathcal{L}_{\text{BC}} \left( \bm{\theta}, \bm{\lambda}_{\text{DBC}}, \bm{\lambda}_{\text{NBC}} \right)
        &= \frac{1}{N_{\text{DBC}}} \sum_{i=1}^{N_{\text{DBC}}}
            m \left( \lambda_{\text{DBC}, i} \right)
            \left|
                \hat{u} (\bm{x}_i; \bm{\theta}) - g_D (\bm{x}_i)
            \right|^2
        \\
        &\quad + \frac{1}{N_{\text{NBC}}} \sum_{i=1}^{N_{\text{NBC}}}
            m \left( \lambda_{\text{NBC}, i} \right)
            \left|
                \bm{n} \cdot \nabla \hat{u} (\bm{x}_i; \bm{\theta}) - g_N (\bm{x}_i)
            \right|^2,
    \end{split}
    \\
    \mathcal{L}_{\text{Data}} \left( \bm{\theta}, \bm{\lambda}_{\text{Data}} \right)
    &= \frac{1}{N_{\text{Data}}} \sum_{i=1}^{N_{\text{Data}}}
        m \left( \lambda_{\text{Data}, i} \right)
        \left|
            \hat{u} \left( \bm{x}_i; \bm{\theta} \right) - u_{\text{Data}} \left( \bm{x}_i \right)
        \right|^2,
\end{align}
where $\bm{\lambda} = \left\{ \bm{\lambda}_{\text{PDE}}, \bm{\lambda}_{\text{DBC}}, \bm{\lambda}_{\text{NBC}}, \bm{\lambda}_{\text{Data}} \right\}$
is a set of non-negative learnable weights assigned to each collocation point.
$m: \mathbb{R}_{\ge 0} \to \mathbb{R}_{> 0}$ is a smooth, strictly monotonically increasing function,
typically chosen as $m \left( \cdot \right) = \max \left( 0, \cdot \right)^2$~\cite{McClenny2023SA,Linka2022BPINNCovid}.
The training of the self-adaptive PINN is formulated as
$\min_{\bm{\theta}} \max_{\bm{\lambda}} \mathcal{L} \left( \bm{\theta}, \bm{\lambda} \right)$,
where $\bm{\theta}$ is updated using gradient descent,
while $\bm{\lambda}$ is updated by gradient ascent algorithm:
\begin{align}
    \bm{\theta}^{(n+1)}
    &\leftarrow \bm{\theta}^{(n)} - \eta_{\bm{\theta}} \nabla_{\bm{\theta}} \mathcal{L} \left( \bm{\theta}, \bm{\lambda} \right),
    \\
    \bm{\lambda}^{(n+1)}
    &\leftarrow \bm{\lambda}^{(n)} + \eta_{\bm{\lambda}} \nabla_{\bm{\lambda}} \mathcal{L} \left( \bm{\theta}, \bm{\lambda} \right).
\end{align}
Here, $\eta_{\bm{\theta}}$ and $\eta_{\bm{\lambda}}$ are the learning rates
for $\bm{\theta}$ and $\bm{\lambda}$, respectively.
In~\cite{McClenny2023SA}, authors recommend using a larger learning rate for $\bm{\lambda}$ compared to $\bm{\theta}$,
i.e., $\eta_{\bm{\lambda}} = c_{\text{SA}} \eta_{\bm{\theta}}$, for some hyperparameter $c_{\text{SA}} \ge 1$.
Self-adaptive PINN has been applied to various domains,
such as infection dynamics~\cite{Linka2022BPINNCovid} and seismic wave modeling~\cite{Ding2023SAPINN_Seamic}.

\begin{figure}[tpb]
    \centering
    \begin{minipage}[b]{.49\linewidth}
        \centering
        \includegraphics[width=.99\linewidth]{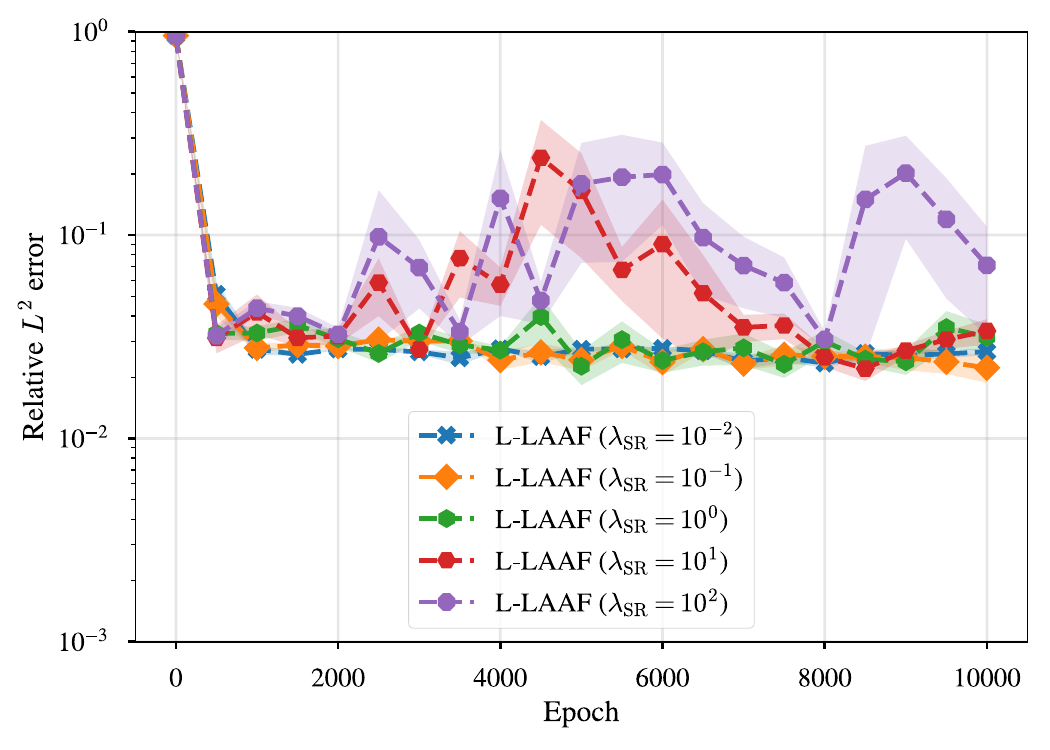}
        \subcaption{L-LAAF}
    \end{minipage}
    \begin{minipage}[b]{.49\linewidth}
        \centering
        \includegraphics[width=.99\linewidth]{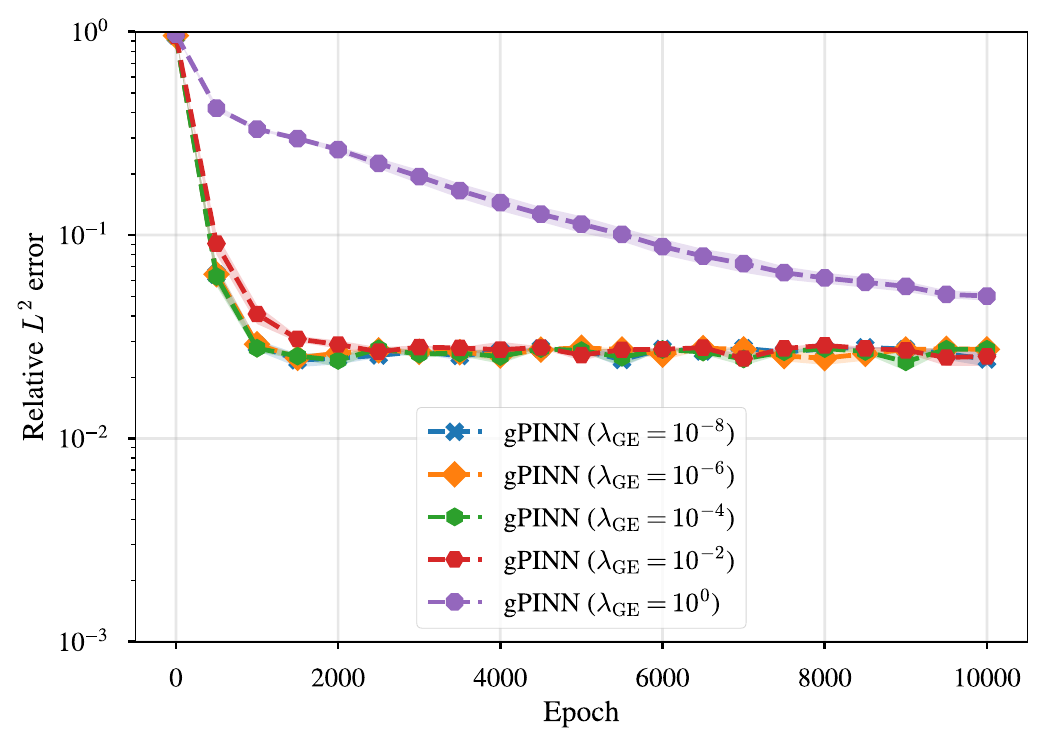}
        \subcaption{gPINN}
    \end{minipage}
    \\
    \begin{minipage}[b]{.49\linewidth}
        \centering
        \includegraphics[width=.99\linewidth]{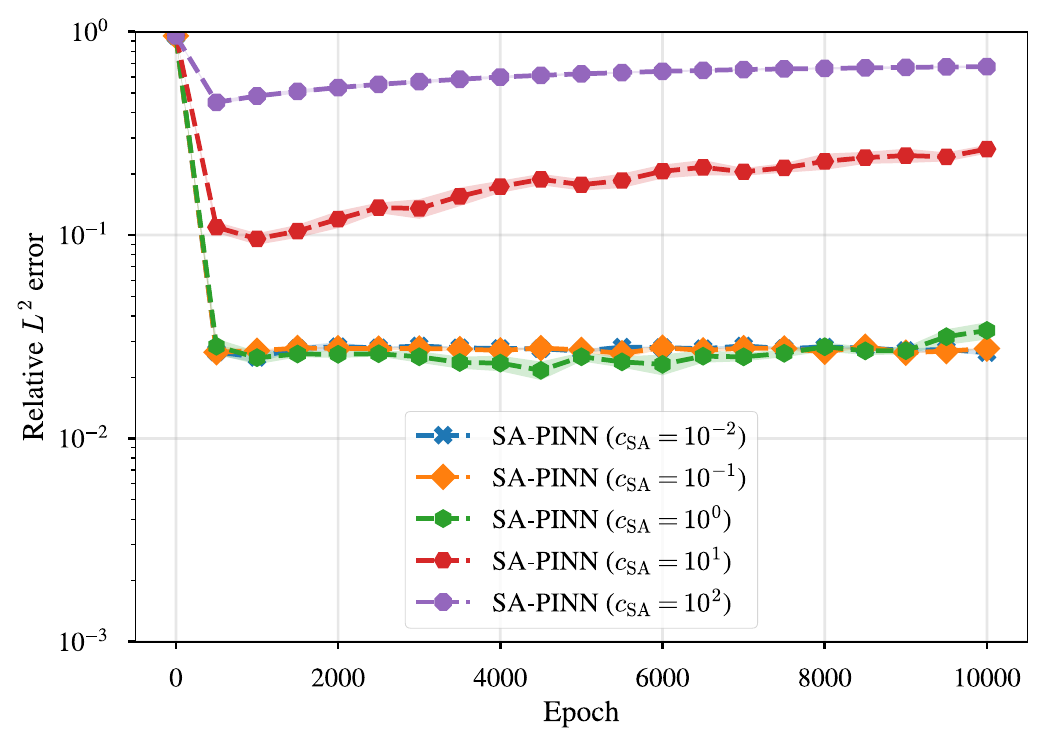}
        \subcaption{SA-PINN}
    \end{minipage}
    \begin{minipage}[b]{.49\linewidth}
        \centering
        \includegraphics[width=.99\linewidth]{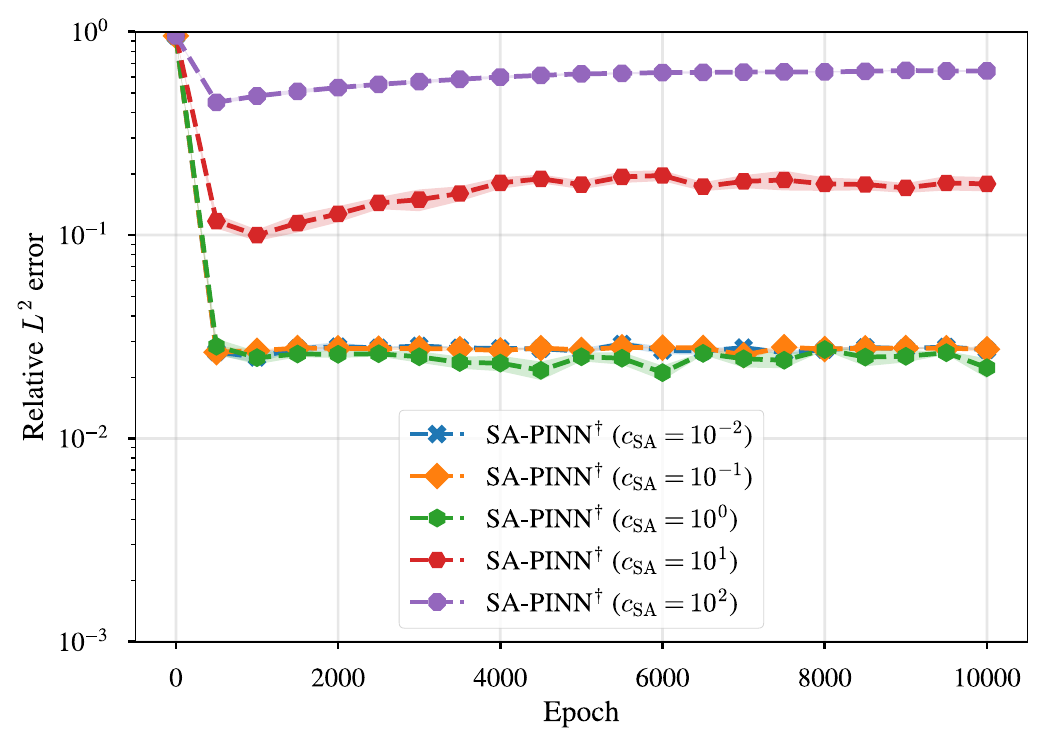}
        \subcaption{SA-PINN$^{\dagger}$}
    \end{minipage}
    \caption{
        \emph{Poisson equation with mixed boundary conditions: homogeneous Neumann condition case (described in Section~\ref{sec:results_poisson}).}
        Relative $L^2$ error of the approximate solution with L-LAAF~\cite{Jagtap2020LAAF}, gPINN~\cite{Yu2022GradEnhc}, and SA-PINN~\cite{McClenny2023SA} using GELU activation function.
        The dashed lines represent the mean, and the shaded areas represent the standard error
        over 5 independent runs with i.i.d. initialization.
        Notably, even with these improvement techniques,
        soft imposition quickly faces a plateau, as observed in Figure~\ref{fig:poisson_hmg_log}~(a).
    }
    \label{fig:poisson_improvement_techniques}
\end{figure}

\begin{table}[tpb]
    \centering
    \caption{
        \emph{Poisson equation with mixed boundary conditions: homogeneous Neumann condition case (described in Section~\ref{sec:results_poisson}).}
        Comparison of L-LAAF~\cite{Jagtap2020LAAF}, gPINN~\cite{Yu2022GradEnhc}, and SA-PINN~\cite{McClenny2023SA}.
        Relative $L^2$ error (mean and standard error) using GELU activation function.
        All values are scaled by $10^{-3}$ for the consistency with Table~\ref{tab:poisson_hmg_summary}.
    }
    \label{tab:poisson_improvement_techniques}
    \begin{subtable}{.24\textwidth}
        \centering
        \caption{L-LAAF}
        \begin{tabular}{c|c}
            \toprule
            $\lambda_{\text{SR}}$ & Rel. $L^2$ Err. \\
            \midrule
            $10^{-2}$ & $26.69 \pm  1.09$ \\
            $10^{-1}$ & $22.24 \pm  3.46$ \\
            $10^{ 0}$ & $31.51 \pm  4.81$ \\
            $10^{ 1}$ & $33.76 \pm  4.81$ \\
            $10^{ 2}$ & $70.99 \pm 39.03$ \\
            \bottomrule
        \end{tabular}
    \end{subtable}
    \hfill
    \begin{subtable}{.24\textwidth}
        \centering
        \caption{gPINN}
        \begin{tabular}{c|c}
            \toprule
            $\lambda_{\text{GE}}$ & Rel. $L^2$ Err. \\
            \midrule
            $10^{-8}$ & $24.46 \pm 1.92$ \\
            $10^{-6}$ & $27.41 \pm 0.27$ \\
            $10^{-4}$ & $27.44 \pm 0.20$ \\
            $10^{-2}$ & $25.33 \pm 2.58$ \\
            $10^{ 0}$ & $50.12 \pm 2.48$ \\
            \bottomrule
        \end{tabular}
    \end{subtable}
    \hfill
    \begin{subtable}{.24\textwidth}
        \centering
        \caption{SA-PINN}
        \begin{tabular}{c|c}
            \toprule
            $c_{\text{SA}}$ & Rel. $L^2$ Err. \\
            \midrule
            $10^{-2}$ & $ 26.42 \pm  0.98$ \\
            $10^{-1}$ & $ 27.66 \pm  0.29$ \\
            $10^{ 0}$ & $ 33.96 \pm  3.30$ \\
            $10^{ 1}$ & $264.84 \pm 13.22$ \\
            $10^{ 2}$ & $673.34 \pm  8.47$ \\
            \bottomrule
        \end{tabular}
    \end{subtable}
    \hfill
    \begin{subtable}{.24\textwidth}
        \centering
        \caption{SA-PINN$^{\dagger}$}
        \begin{tabular}{c|c}
            \toprule
            $c_{\text{SA}}$ & Rel. $L^2$ Err. \\
            \midrule
            $10^{-2}$ & $ 27.28 \pm  0.51$ \\
            $10^{-1}$ & $ 27.43 \pm  0.88$ \\
            $10^{ 0}$ & $ 22.26 \pm  2.48$ \\
            $10^{ 1}$ & $178.51 \pm 14.38$ \\
            $10^{ 2}$ & $641.24 \pm  4.83$ \\
            \bottomrule
        \end{tabular}
    \end{subtable}
\end{table}

Figure~\ref{fig:poisson_improvement_techniques} and Table~\ref{tab:poisson_improvement_techniques}
summarizes the relative $L^2$ error
with L-LAAF~\cite{Jagtap2020LAAF}, gPINN~\cite{Yu2022GradEnhc}, and SA-PINN~\cite{McClenny2023SA}
for the Poisson equation problem with inhomogeneous Dirichlet and homogeneous Neumann conditions.
The problem setup is the same as that in Section~\ref{sec:results_poisson},
and the hyperparameters are selected based on the numerical experiments in the original papers and their recommendations described therein.
For SA-PINN, we considered two distinct strategies:
one where both $\bm{\theta}$ and $\bm{\lambda}$ are updated throughout the optimization (SA-PINN),
and the other where a two-stage optimization is employed (SA-PINN$^{\dagger}$).
In the latter case,
$\bm{\theta}$ and $\bm{\lambda}$ were jointly optimized for the first half of the total training epochs,
after which $\bm{\lambda}$ was frozen,
and only $\bm{\theta}$ was updated for the remaining epochs, following~\cite{McClenny2023SA}.
The results suggest that these methods do not consistently yield substantial improvements,
and their effectiveness is highly sensitive to the hyperparameters.


\section{Numerical study on normalization orders in distance functions}   \label{sec:appendix_distance}
\renewcommand{\theequation}{B.\arabic{equation}}
\renewcommand{\thefigure}{B.\arabic{figure}}
\renewcommand{\thetable}{B.\arabic{table}}
\setcounter{equation}{0}
\setcounter{figure}{0}
\setcounter{table}{0}

\begin{figure}[tbp]
    \centering
    \begin{minipage}[b]{.55\linewidth}
        \centering
        \includegraphics[width=.99\linewidth]{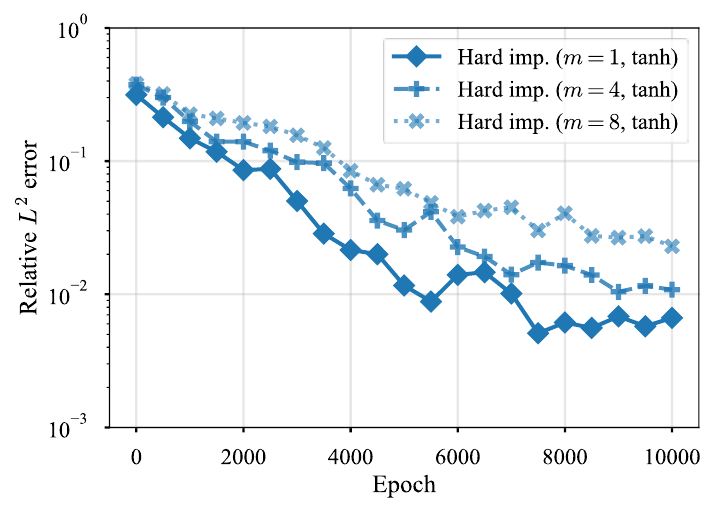}
        \subcaption{tanh}
    \end{minipage}
    \\
    \begin{minipage}[b]{.55\linewidth}
        \centering
        \includegraphics[width=.99\linewidth]{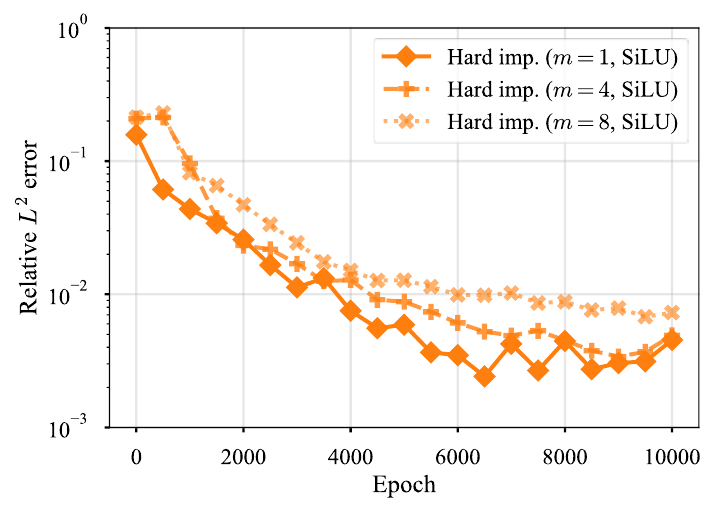}
        \subcaption{SiLU}
    \end{minipage}
    \\
    \begin{minipage}[b]{.55\linewidth}
        \centering
        \includegraphics[width=.99\linewidth]{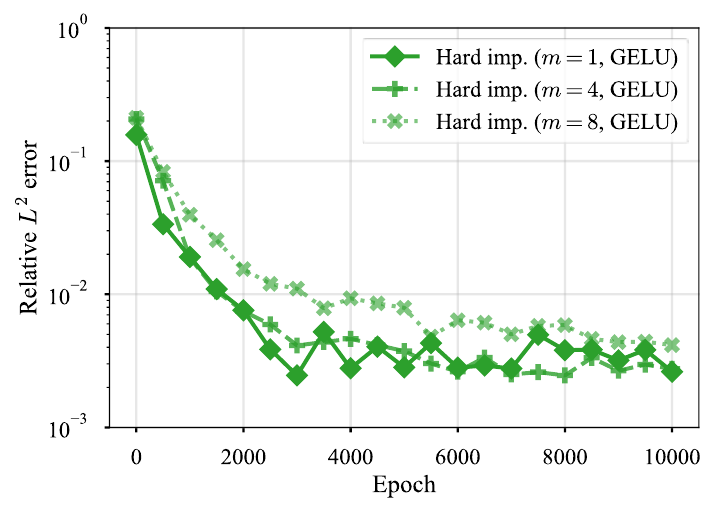}
        \subcaption{GELU}
    \end{minipage}
    \\
    \caption{
        \emph{Poisson equation with mixed boundary conditions: homogeneous Neumann condition case (described in Section~\ref{sec:results_poisson}).}
        Relative $L^2$ error of the approximate solution with different normalization orders $m$.
        Only the mean over 5 i.i.d. runs is shown,
        and results with $m = 2$ are omitted for clarity,
        as they show similar behavior to $m = 1$.
    }
    \label{fig:poisson_error_different_m}
\end{figure}

\begin{table}[tbp]
    \centering
    \caption{
        \emph{Poisson equation with mixed boundary conditions: homogeneous Neumann condition case (described in Section~\ref{sec:results_poisson}).}
        Relative $L^2$ error
        (mean and standard error over 5 independent runs with i.i.d. initialization, scaled by $10^{-3}$)
        of approximate solution with different normalization orders $m$.
    }
    \label{tab:poisson_error_different_m}
    \begin{tabular}{l|cccc}
        \toprule
        Activation function & $m = 1$         & $m = 2$          & $m = 4$          & $m = 8$ \\
        \midrule
        tanh                & $6.65 \pm 2.08$ & $14.27 \pm 1.68$ & $10.82 \pm 1.52$ & $22.92 \pm 2.37$ \\
        SiLU                & $4.54 \pm 1.05$ & $ 5.69 \pm 1.24$ & $ 4.92 \pm 1.49$ & $ 7.29 \pm 2.08$ \\
        GELU                & $2.62 \pm 0.21$ & $ 3.25 \pm 0.70$ & $ 2.75 \pm 0.23$ & $ 4.15 \pm 0.94$ \\
        \bottomrule
    \end{tabular}
\end{table}

In~\cite{Sukumar2022MixedBC,Berrone2023BCEnforcement},
the behavior of the Laplacian of the ADFs is discussed analytically.
Here, we numerically investigate the influence of the normalization order $m$
of the distance functions on the convergence of PINN solutions.
Specifically, we consider the Poisson equation problem
(described in Section~\ref{sec:results_poisson}) as an example.
Figure~\ref{fig:poisson_error_different_m} illustrates the error
with different activation functions and normalization orders.
The results, summarized in Table~\ref{tab:poisson_error_different_m},
indicate a clear trend:
as the normalization order $m$ increases,
the convergence of PINN solutions deteriorates for all tested activation functions,
although the extent of the degradation varies.

\begin{figure}[tpb]
    \centering
    \begin{minipage}[b]{.24\linewidth}
        \centering
        \includegraphics[width=.99\linewidth]{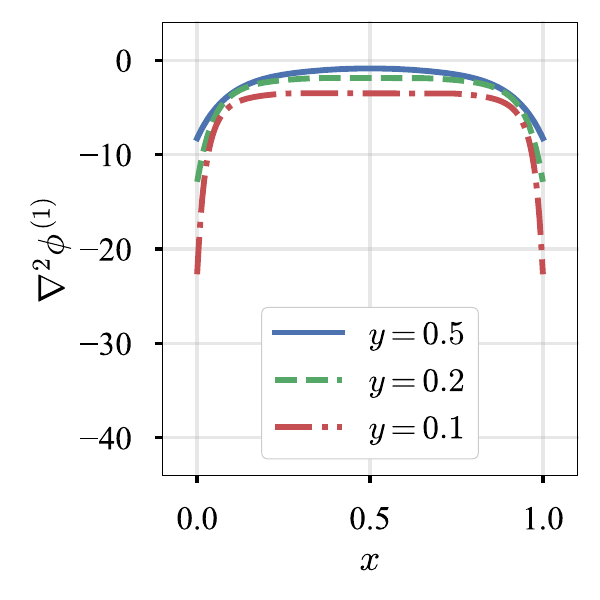}
        \subcaption{$m = 1$}
    \end{minipage}
    \begin{minipage}[b]{.24\linewidth}
        \centering
        \includegraphics[width=.99\linewidth]{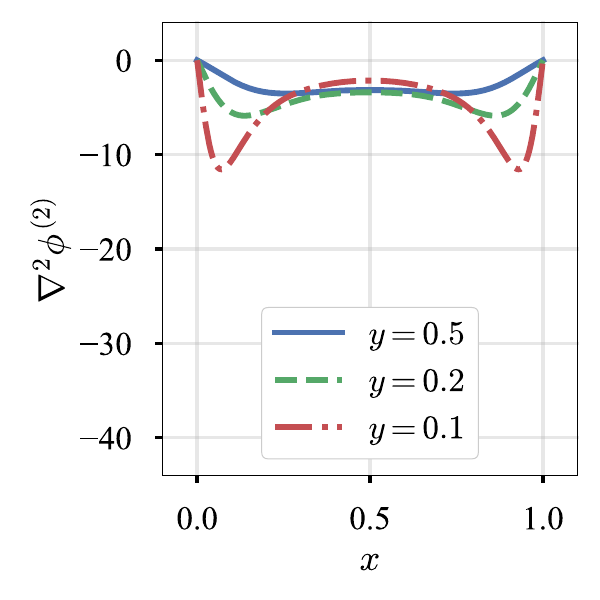}
        \subcaption{$m = 2$}
    \end{minipage}
    \begin{minipage}[b]{.24\linewidth}
        \centering
        \includegraphics[width=.99\linewidth]{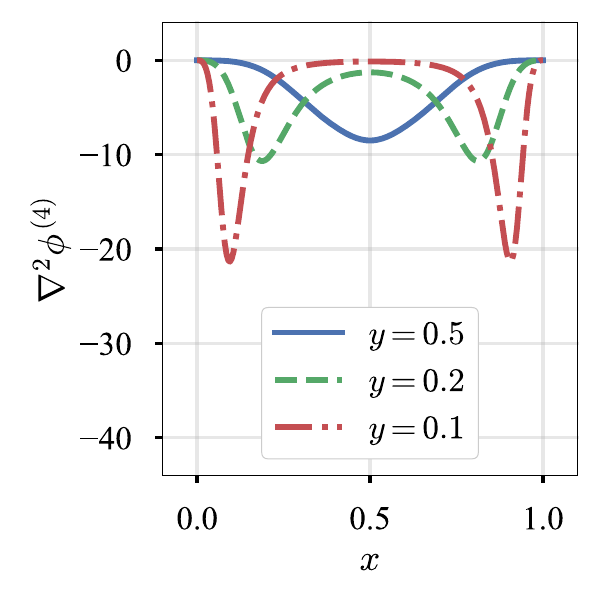}
        \subcaption{$m = 4$}
    \end{minipage}
    \begin{minipage}[b]{.24\linewidth}
        \centering
        \includegraphics[width=.99\linewidth]{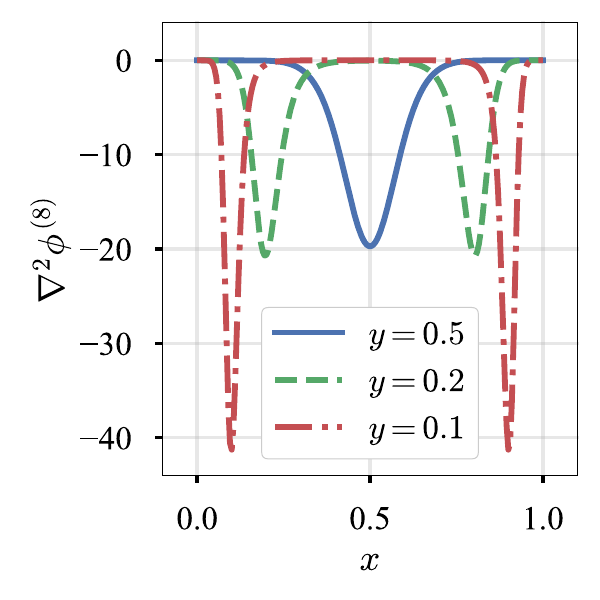}
        \subcaption{$m = 8$}
    \end{minipage}
    \caption{
        \emph{Laplacian of the normalized ADFs to a unit square (slice).}
        Slice along $y = 0.5, 0.2, 0.1$ are shown.
        As $m$ increases, the oscillatory behavior near the boundary becomes more pronounced.
    }
    \label{fig:laplacian_adf_different_m}
\end{figure}

To understand the reason behind this behavior,
we examine the Laplacian of the ADFs over a unit square
with various normalization orders $m$, as depicted in Figure~\ref{fig:laplacian_adf_different_m}.
It is evident that the Laplacian $\nabla^2 \phi^{(m)}$ exhibits an oscillatory behavior 
particularly near the boundary, as the normalization order $m$ is increased.
Simplifying the problem to the Poisson equation with the homogeneous Dirichlet boundary condition,
the approximate solution is defined as $\tilde{u} = \bar{g}_{D} + \phi^{(m)} \hat{u}$
(setting $\bar{g}_{D} = 0$ in Equation~\eqref{eq:hard_imposition}), where its Laplacian is given by:
\begin{align}
    \nabla^2 \tilde{u}
    &= \nabla \cdot \nabla \left( \phi^{(m)} \hat{u} \right) \\
    &= \nabla \cdot \left( \hat{u} \nabla \phi^{(m)} + \phi^{(m)} \nabla \hat{u} \right) \\
    &= \hat{u} \nabla^2 \phi^{(m)} + \phi^{(m)} \nabla^2 \hat{u} + 2 \left( \nabla \hat{u}, \nabla \phi^{(m)} \right),
\end{align}
where $\left( \cdot, \cdot \right)$ denotes the inner product.
The presence of $\nabla^2 \phi^{(m)}$ in $\nabla^2 \tilde{u}$ suggests a high possibility that
$\nabla^2 \tilde{u}$ inherits the oscillatory nature of $\nabla^2 \phi^{(m)}$,
potentially resulting in the degraded convergence observed in Figure~\ref{fig:poisson_error_different_m} and Table~\ref{tab:poisson_error_different_m}.
This implies that the precise representation of the distance (i.e., higher $m$) is not necessarily advantageous or beneficial;
rather, the normalization order $m$ should be kept moderate
so as not to require $\hat{u}$ to compensate for the near-boundary oscillations
$\nabla^2 \phi^{(m)}$ may induce,
especially when the second derivative appears in the problem of interest.

\begin{figure}[tpb]
    \centering
    \begin{minipage}[b]{.99\linewidth}
        \centering
        \includegraphics[width=.99\linewidth]{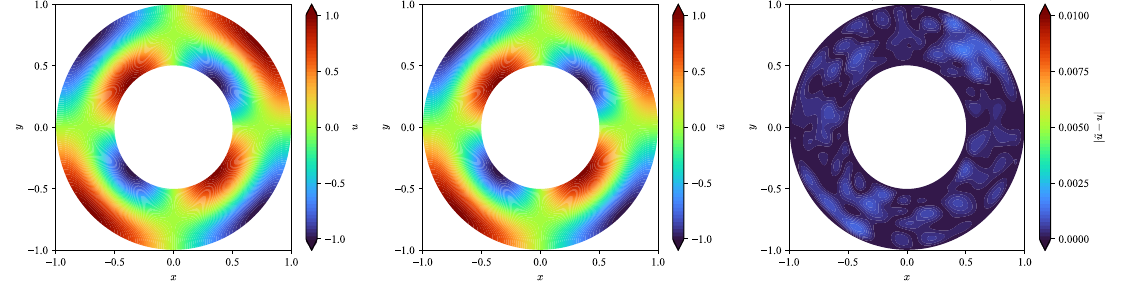}
        \subcaption{
            Hard imposition ($m = 1$, GELU). Relative $L^2$ error: $7.36 \times 10^{-4}$.
        }
    \end{minipage}
    \\
    \vspace{5mm}
    \begin{minipage}[b]{.99\linewidth}
        \centering
        \includegraphics[width=.99\linewidth]{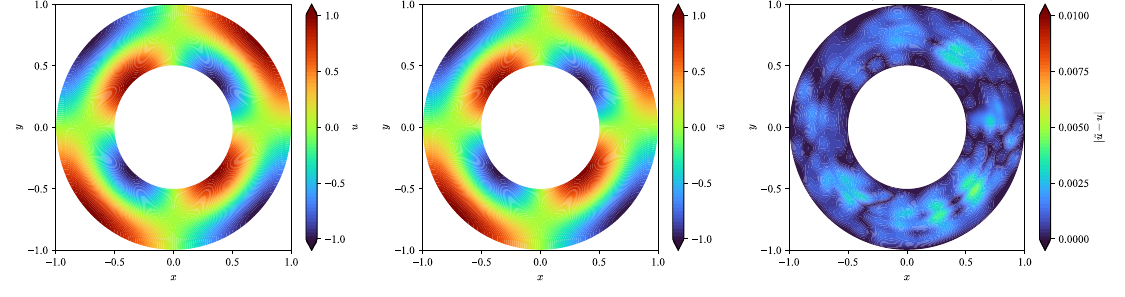}
        \subcaption{
            Hard imposition ($m = 8$, GELU). Relative $L^2$ error: $2.48 \times 10^{-3}$.
        }
    \end{minipage}
    \caption{
        \emph{Poisson equation on an annulus.}
        (From left to right)
        reference solution obtained by the method of manufactured solution,
        approximate solution,
        and absolute error against the reference solution.
    }
    \label{fig:poisson_annulus_results}
\end{figure}

In order to verify that this is not a special case for a square domain,
we also conducted the same analysis for an annulus,
$\Omega = \{ (r, \varphi) \mid 0.5 \le r \le 1, 0 \le \varphi \le 2 \pi \}$.
We applied the method of manufactured solutions to obtain the reference (analytical) solution and source:
\begin{align}
    u
    &= \cos \left( 2 \pi c_1 r \right) \sin \left( c_2 \varphi \right),
    \\
    f
    &= \frac{2 \pi c_1}{r} \sin \left( 2 \pi c_1 r \right) \sin \left( c_2 \varphi \right)
        + \left( \left( 2 \pi c_1 \right)^2 + \left( \frac{c_2}{r} \right)^2 \right) 
            \cos \left( 2 \pi c_1 r \right) \sin \left( c_2 \varphi \right),
\end{align}
where we choose $\left( c_1, c_2 \right) = \left( 1, 2 \right)$ and $N_{\text{PDE}} = 4,096$. 
Figure~\ref{fig:poisson_annulus_results} shows the illustrative results
and Table~\ref{tab:poisson_annulus_error_different_m} summarizes the relative $L^2$ error
with different normalization orders $m$ for the hard imposition.
Although SiLU achieves the lowest error with $m = 2$,
the overall trend is consistent with the square domain case,
and the convergence deteriorates as $m$ increases.

\begin{table}[tpb]
    \centering
    \caption{
        \emph{Poisson equation on an annulus.}
        Relative $L^2$ error of the approximate solution with different normalization orders $m$.
        Mean and standard error over 5 independent runs with i.i.d. initialization.
        All values are scaled by $10^{-3}$.
    }
    \label{tab:poisson_annulus_error_different_m}
    \begin{tabular}{l|cccc}
        \toprule
        Activation function & $m = 1$         & $m = 2$         & $m = 4$         & $m = 8$ \\
        \midrule
        tanh                & $0.26 \pm 0.04$ & $0.49 \pm 0.12$ & $0.45 \pm 0.05$ & $0.81 \pm 0.06$ \\
        SiLU                & $1.64 \pm 0.45$ & $1.34 \pm 0.14$ & $2.69 \pm 0.50$ & $5.90 \pm 1.08$ \\
        GELU                & $0.70 \pm 0.10$ & $1.06 \pm 0.18$ & $1.13 \pm 0.06$ & $1.61 \pm 0.29$ \\
        \bottomrule
    \end{tabular}
\end{table}


\end{appendices}

\bibliography{references}
\bibliographystyle{unsrt}

\end{document}